\documentclass[12pt]{article}

\usepackage {algorithmic, amssymb, amsmath, amsthm, amsfonts, appendix, array, authblk, bm, booktabs, breakurl, caption, calrsfs, chemformula, color, colortbl, comment, derivative, enumerate, eurosym, fixltx2e, float, footmisc,  graphicx, hhline, hyperref, indentfirst, lipsum, lineno, longtable, lscape, mathtools, mathrsfs, multirow, multicol, natbib, pdflscape, psfrag, ragged2e, rotating, setspace, sectsty, scrextend, subfigure, longtable, tabularx, textcomp, threeparttable, tikz, upgreek, url, ulem, xcolor}

\usepackage{booktabs}
\usepackage{stackrel}
\usepackage{hhline}
\usepackage{mdframed}

\usepackage[margin=1in]{geometry}
\usepackage[utf8]{inputenc}
\newcolumntype{R}[1]{>{\RaggedRight}p{#1}}

\begin{document}
{

\begin{titlepage}
\title{
The Good, the Bad, and the Ugly: The Role of AI Quality Disclosure in Lie Detection
 
}\par

\author{Haimanti Bhattacharya\thanks{University of Utah, Department of Economics, 260 Central Campus Drive, GC 4100, Salt Lake City, UT 84112, USA; Email: haimanti.bhattacharya@utah.edu; Professor Bhattacharya gratefully acknowledges the College of Social and Behavioral Sciences (CSBS) seed grant that made this study possible.} 

\vspace{-1em}
 \and Subhasish Dugar\thanks{University of Utah, Department of Economics; Email: subhasish.dugar@utah.edu; Professor Dugar is grateful to acknowledge the Faculty Fellow Award, 2024, from the University of Utah, which allowed him to dedicate time solely for this paper.}  
 
 \vspace{-1em}
\and Sanchaita Hazra\thanks{Corresponding author; University of Utah; Department of Economics; Email: sanchaita.hazra@utah.edu. Ms. Hazra thankfully acknowledges funding from Global Change and Sustainability Center, the Wilkes Center for Climate Science and Policy, and Richard and Leslie Haskell Scholarship.} 

\vspace{-1em}
\and Bodhisattwa Majumder\thanks{Allen Institute for AI, 2157 N Northlake Way \#110, Seattle, USA; Email: bodhisattwam@allenai.org. This research has been approved by the Institutional Review Board of the University of Utah under study number IRB00167477. The authors are grateful to the seminar participants at the University of Reading; Indian Statistical Institute, Kolkata; Centre for Studies in Social Sciences, Kolkata; and Central European University in Vienna for their valuable insights. We are also thankful to the session attendees of the North American Economic Science Association meeting, 2023, Charlotte; Soleto Summer School and Workshop on Experimetrics and Behavioral Economics, 2024, Italy; Behavioral Research Workshop at the University of Utah;  2024 Annual Conference of the North American Chapter of the Association for Computational Linguistics for their helpful comments. We would like to thank Chris Callison-Burch, Harshit Surana, Ranjay Krishna, Marta Serra-Garcia, Lugi Luini, Peter Clark, Daniel Martin, Monica Capra, Jeffrey Jobu Morris Babin, Marco A Palma, Tushar Khot, Kalyani Chaudhuri, Kelvin Klyman, Priyadarshi Banerjee, and Sachin Kumar for their insightful comments. Senior authorship is not assigned.}}
\date{}
\maketitle
\vspace{-1em}
\begin{abstract}
We investigate how low-quality AI advisors, lacking quality disclosures, can help spread text-based lies while seeming to help people detect lies. Participants in our experiment discern truth from lies by evaluating transcripts from a game show that mimicked deceptive social media exchanges on topics with objective truths. We find that when relying on low-quality advisors without disclosures, participants’ truth-detection rates fall below their own abilities, which recovered once the AI’s true effectiveness was revealed. Conversely, high-quality advisor enhances truth detection, regardless of disclosure. We discover that participants’ expectations about AI capabilities contribute to their undue reliance on opaque, low-quality advisors.\\

\bigskip


\end{abstract}

\setcounter{page}{0}
\thispagestyle{empty}
\end{titlepage}
\pagebreak \newpage

\begin{spacing}{1.5}
\section{Introduction}\label{sec:intro}
Artificial intelligence (AI) has recently garnered unprecedented attention from both popular media (e.g., Economist, 2024; New York Times, 2023; Washington Post, 2024) and leading researchers (e.g., Acemoglu, 2024; Ludwig et al., 2023; Elliott, 2019; Rahwan et al., 2019; Kleinberg et al., 2018; Brynjolfsson et al., 2017; Makridakis, 2017) due to its transformative impacts on contemporary life. Fueled by myriad applications, AI has gained prominence as advisors for individuals across diverse spheres of daily life and is poised to expand its advisory role in plenty of other domains (Fast and Schroeder, 2020).\footnote{Extensive research shows how AI advises a range of daily life activities, including entertainment (Yeomans et al., 2019), dating (Jin and Zhang, 2023), student admissions (Dietvorst et al., 2015), legalities (Ludwig and Mullainathan, 2024), cancer diagnosis (McKinney et al., 2020), personalized ads (Mogaji et al., 2020; Boerman et al., 2017), financial services (Arli et al., 2021), child welfare inquiries (Chouldechova et al., 2018), food choices (Bu\c{c}inca et al., 2021), loan approvals (Green and Chen, 2019), news recommendations (Thurman and Schifferes, 2015), and pedagogy (Wang et al., 2024).}

As AI-assisted decision-making becomes increasingly common, evidence shows people either overly rely on AI advice (``AI appreciation'') or lack sufficient confidence in it (``AI aversion'').\footnote{Overreliance or underreliance is seen in the computing literature as excessive or deficient trust in AI, rather than being measured against an optimal benchmark of usage. For an overview of this literature, see Guo et al. (2024), Vasconcelos et al. (2023), Schemmer et al. (2022), Bu\c{c}inca et al. (2021), Zhang et al. (2020), Yeomans et al. (2019), Logg (2017), and Dietvorst et al. (2018).} Despite remarkable advances in AI, such suboptimal reliance on AI may have adverse consequences across various user domains. For instance, as we usher in a new era of detecting deception online (Alom et al., 2020; Monaro et al., 2020), such suboptimal reliance on AI may hinder efforts to curb the spread of lies — the primary focus of our work. Malicious actors may exploit these behavioral biases to spread lies and evade detection, a concern recognized in the AI policy treatises of the US, EU, OECD, and G7.\footnote{See the United States Executive Order on AI (White House, 2023), the European Union (EU) Ethics Guidelines for Trustworthy AI (2019),  EU Commission (2024), Organisation for Economic Co-operation and Development) (OECD) Council on AI (2020), and G7 Code of Conduct (EY, 2022); see Ladak et al. (2024), Gratch and Fast (2022), K\"{o}bis et al. (2021), Rothman (2018), and Angwin et al. (2016).}  As the efficacy of AI tools remains largely opaque to users in the current digital milieu, the looming danger has spurred a consensus on the need to unveil the black box of AI, enhancing its transparency, reliability, and usability.\footnote{See Bommasani et al. (2024), Aïmeur et al. (2023), Zerilli et al. (2022), Floridi et al.(2021), Schmidt et al. (2020), Ishowo-Oloko et al. (2019), Jobin et al. (2019), Shin and Park (2019), and Bathaee (2017).} While common transparency mandates focus on algorithmic clarity and explainability to foster trust among downstream users, a crucial remedy proposed to combat nefarious intentions and promote proper AI use entails disclosing quantitative metrics of the AI's quality.\footnote{Refer to Requirement 4 of the EU Ethics Guidelines for Trustworthy AI (2019) advocating for the adoption of such measures. Also, consult Díaz-Rodríguez et al. (2023), Vilone and Longo (2021), and Licht and Licht (2020), recommending similar measures.}

Set against this backdrop, this paper explores a relatively novel application of AI: providing individuals with AI advisors of varying quality to aid them in detecting lies in \textit{textual} form, where the AI’s efficacy in detecting such deceptions may or may not be available to the user. While text-based lies flourish in the online realm, the role of AI in their detection within economic research remains strikingly understudied. We advance this emerging body of work by investigating a hitherto unexplored context where motives for deception abound and individuals attempt to discern truth from lies while perusing written discussions on topics with objectively verifiable facts unknown to them. Leveraging recent advances in Large Language Models (LLMs), we deliberately simulate AI advisors of varying caliber — inferior, average, and superior — quantified by their accuracy rates. Participants in our experiment, offered the choice to consult an AI advisor of a specific quality to detect the truth, are exposed to one of two AI environments: blackbox (AI efficacy undisclosed) or with information (AI efficacy disclosed). To the best of our knowledge, no prior research has investigated the complex calculus of how the quality spectrum of AI advisors and the (lack of) disclosure of their effectiveness collectively influence individuals' reliance on AI across any domain, including lie detection.

As text-based lies manifest in diverse forms, we let our participants discern the objective truth by inspecting naturally generated text fraught with lies wherein parties with conflicting interests discuss a fact-based topic in a back-and-forth Q\&A format.\footnote{Examples of text-based lies include misleading news articles, deceiving social media posts on topics like economics, history, vaccine efficiency, and climate change, fake product reviews and features.} Our setup thus captures the essence of popular Q\&A forums such as Reddit, Quora, Answers.com, etc., where lies about facts are often spread via written exchanges by opposing parties. Appendix A presents a Quora excerpt, exemplifying the type of text-based lies about facts we probe. We opted to focus on this specific form of text-based deception since online Q\&A forums, the prominent conduit for spreading text-based lies, routinely disseminate lies in this format (Tsou, 2023). These platforms, characterized by user-generated content and minimal oversight, create fertile grounds for deception on critical topics grounded in facts. Moreover, they boast enormous engagement, wielding considerable influence in shaping individual opinion.\footnote{Reddit hosts roughly 3.1 million ``subreddits" covering various topics, with about 52 million daily users generating billions of annual comments (Foundationinc, 2021; Lee, 2022; Backlinko, 2021). Quora also attracts substantial traffic, with roughly 1.19 billion visits in May 2024 (Semrush, 2024).} Furthermore, research indicates that these forums are particularly vulnerable to exploitation by malicious actors due to their financial dynamics (Gazan, 2011; Shah et al., 2008).\footnote{ For instance, on Reddit and Quora, content visibility is influenced by community ``upvoting" or ``downvoting," and users can financially endorse posts with awards like ``Reddit gold" or ``Reddit platinum," conferring substantial financial benefits on the original posters.} The increasing use of bots and algorithms in ``information wars" to amplify messaging impacts highlights the need for transparent AI systems to assist users in discerning between truth from lies.\footnote{For research advocating for enhanced AI to detect lies online, see, e.g., Zhang et al. (2023), Van Der Zee et al. (2022), Fornaciari et al. (2021), Sager et al. (2021), Pew Research Report (2017), Shu et al. (2017), Ball and Elworthy (2014), Mihalcea and Strapparava (2009), and Zhou et al. (2004).}

Several reasons justify our exclusive attention to text-based lies concerning objectively verifiable facts. First, it aligns with the economic literature on deception, typically investigating objective falsities akin to the one documented in Appendix A, such as those investigated in Gneezy (2005) or the die-rolling experiments in Fischbacher and Föllmi-Heusi (2013) (see Abeler et al., 2019 for a review). Second, the focus is consistent with the observed preference of individuals to rely more on AI advice for objective than subjective matters (Laakasuo et al., 2021; Castelo et al., 2019; Logg et al., 2019). Third, recent research has discovered a troubling trend: individuals often struggle to differentiate between facts and lies in contexts involving objective truths, contrary to the presumption that such distinctions should be an easy task (e.g., Belot and van de Ven, 2017; Ockenfels and Selten, 2000; Bond and DePaulo, 2006). Fourth, although information in audio and/or visual forms is more prevalent and elicits stronger emotional reactions than comparable textual information, evidence shows that the latter is equally persuasive (Glasford, 2013; Yadav et al., 2011; Appiah, 2006; Messaris and Abraham, 2001) and harmful (Wittenberg et al., 2021; Powell et al., 2018; Tukachinsky et al., 2011).  

As source material for Q\&A-based text faithfully mirroring the relevant online content, we utilize transcripts of multiple sessions of the first season (1956–1959) of the American TV show \textit{To Tell The Truth} (T4). We financially incentivize participants to discern the objective truth from lies embedded in each transcript corresponding to a distinct session of T4 featuring a host, four judges, and three contestants. All contestants claim to be the genuine John/Jane Doe or the real character (RC) in the show, with only one being truthful, while the others aim to deceive the judges into believing they are the RC. The show begins with the revelation of facts about the RC from an affidavit signed by the RC before filming. The judges then interrogate the contestants through back-and-forth Q\&A. After Q\&A, judges guess who the RC is. Contestants earn an equal share of \$250 for each incorrect guess, potentially amassing up to \$333 if all guesses are incorrect — a large sum in the 1950s, equating to approximately \$3500 in 2024.\footnote{Inflation adjusted amount calculated from: https://www.bls.gov/data/inflation calculator.htm.}

Our decision to use lies embedded in transcripts from multiple sessions of T4, rather than real-life examples curated from Q\&A sites, is primarily aimed at minimizing the influence of motivated reasoning on our participants' ability to distinguish truth from lies. Motivated reasoning, documented extensively in lie detection literature (e.g., Kahan, 2015; Kunda, 1990), describes how people interpret information to align with their existing beliefs or preferences. By utilizing T4 transcripts that consistently center on neutral topics, we aim to mitigate, if not entirely bypass, the influence of strong prior beliefs, thereby enabling participants to unearth lies with greater objectivity.\footnote{Computer science papers often assess lie detection abilities using politically charged news across media formats (e.g., audio, visual, and text), which can inadvertently trigger motivated reasoning and impair judgment in detecting lies (e.g., Groh et al., 2022; Murphy and Flynn, 2022; Barari et al., 2021).} By sidestepping motivated beliefs, our approach provides a cleaner evaluation of individuals' abilities to distinguish between factual information and deceptive text. T4 transcripts also closely resemble the scenario in Appendix A, bestowing an important benefit. Appendix B provides an excerpt from a T4 transcript we use, showcasing a style akin to the content in Appendix A. From the participant's perspective, analyzing T4 transcripts mirrors situations where an impartial third party, with no or some prior knowledge of a topic, seeks factual information in discussions — akin to individuals consulting Q\&A forums to uncover truths (as in Appendix A). We posit that both scenarios pose comparable challenges in discerning truth amid discussions tinged with conflicting interests.\footnote{One might argue that using transcripts from an older TV show introduces temporal distance from current topics, potentially limiting participants' ability to discern the truth. We contend that similar challenges arise in everyday situations where individuals seek objective truths from past events, such as historical or economic data, even without precise knowledge of the underlying truth.}

Our database of transcripts is constructed in two phases. In Phase 1, we downloaded 385 of 429 sessions from YouTube. From these, we randomly selected and transcribed 195 sessions using an open-source transcription tool. After applying strict inclusion criteria, detailed in Section 3, we further refined this selection to 132 transcripts paired with corresponding affidavits that included sufficient textual cues for identifying the RC. Phase 2 utilized GPT-4, a state-of-the-art LLM, to generate guesses for all 132 transcripts. The model achieved an overall accuracy rate of 56.06\%, significantly outperforming the accuracy rate based on random guessing (33\%). The pool of all guesses enabled us to pre-populate two sets of AI guesses, correct (74 transcripts) and incorrect (58 transcripts), later used to simulate three tiers of AI advisors: low-, medium-, and high-quality. We set the accuracy rate of the low-quality ($LQ$) advisor at 20\%, representing a notably underperforming advisor, by randomly sampling one transcript from the correct and four from the incorrect set without replacement. Following a similar method, the medium-quality ($MQ$) advisor was configured with a 40\% accuracy rate, marginally better than a random guesser (33\%). In contrast, the high-quality ($HQ$) advisor was assigned a 60\% accuracy rate, exceeding chance and close to the overall AI accuracy rate.

There are several reasons to think that accuracy rate is a good proxy for conveying an AI advisor's quality. Accuracy rates offer a straightforward and easy-to-understand index for measuring the reliability of an AI’s guess. Such precise feedback mechanisms are particularly useful in contexts where an objective truth exists, enabling users to discern the veracity of the information with greater confidence. Finally, similar rating systems already exist (e.g., ReviewMeta), evaluating the authenticity of Amazon reviews. 

We employ a between-subjects design in which participants are presented with one of the three clusters of five transcripts, each cluster corresponding to a specific AI advisor quality. Their task involves financially motivated attempts to identify the RC for each transcript within the cluster. Following this, participants are shown the AI's guesses for all five transcripts, one at a time. At this point, participants are given the option to either retain their initial guess or switch to the AI's guess for each transcript. The conditions under which participants switch guesses vary based on whether or not they receive information about the AI's efficacy. This variation in information enables us to investigate participants' guesses in counterfactual scenarios, specifically examining how their guesses would differ with and without knowledge of the AI advisors' capabilities.

We have four key findings. First, participants, on the whole, exhibit a slightly better-than-chance ability to identify the RC on their own. Notably, they register significantly higher truth-detection rates for the $MQ$ and $HQ$ transcript clusters, while their performance falls below chance with the $LQ$ cluster. Second, when participants engage with the $HQ$ advisor, their final truth-detection rate significantly surpasses their own ability, irrespective of whether they are privy to the AI’s actual capability. Their reliance on the advisor remains similar in both information conditions, showcasing what we call the ``good” effect — where a highly capable AI advisor significantly enhances truth detection, rendering the disclosure of its capability largely inconsequential to decision-makers. Third, when participants are apprised of the $MQ$ advisor’s capability, the final truth-detection rate does not improve relative to their own ability. Moreover, when they were unaware of its quality, there was a decline, albeit insignificant, in the truth detection rate relative to their own, which we label the ``bad" effect of relying on a modest-quality, opaque AI advisor. Finally, when unaware of the $LQ$ advisor’s poor capability, participants’ reliance on it causes the final truth-detection rate to descend significantly below their intrinsic rate, allowing more lies to go undetected, lies that could have been detected if they had relied on their own ability instead of depending on a subpar, opaque AI. This phenomenon exemplifies the ``ugly” effect of relying on an inferior, non-transparent AI advisor. However, once informed, participants recalibrate their reliance on the AI advisor, preventing the drop in truth detection. 

We sought to understand why participants significantly reduced their reliance on transparent $LQ$ and $MQ$ advisors and yet showed no such variation under similar information disclosure protocols for the $HQ$ advisor. To this end, we elicited participants' beliefs about the AI's ability to identify the RC, both with and without financial incentives, and varied the timing of belief elicitation. We gather broad evidence that undue reliance on subpar opaque AI advisors is entirely driven by a positive gap between individuals’ expectations of AI’s lie-detection prowess and the actual performance of the AI advisors.  In sharp contrast, when the quality of our AI advisor meets participants’ expectations, the disclosure of the AI’s true capability becomes inconsequential. As a result, with the $HQ$ AI advisor, participants exhibit consistent reliance, regardless of whether or not they are informed about the AI's true capability. Thus, from a policymaking perspective, the interplay between people's context-specific expectations of AI and AI's true capabilities bears significant consequences for the proliferation of text-based lies. 

The paper is organized as follows: Section 2 reviews the related literature. Section 3 outlines the game show, the transcription process, the criteria for including transcripts, and the creation of three distinct AI advisors. Section 4 introduces a simple behavioral framework to develop testable hypotheses. Section 5 explains the treatments and experimental methods. Section 6 presents the main findings, while Section 7 discusses the implications of our results and offers the concluding remarks.

\section{Related Literature}\label{sec:lit review}

Our paper contributes to a growing field in economics focused on the detection of deception in different contexts (von Schenk et al., 2024; Serra-Garcia and Gneezy, 2024, 2021; Leib et al., 2024; Belot and van de Ven, 2017; Belot et al., 2012 for a meta-analysis, see Bond and DePaulo, 2006). Our work directly speaks to pivotal works by von Schenk et al. (2024) and Serra-Garcia and Gneezy (2024), which explore, respectively, how algorithmic advice in textual form enhances people's ability to detect lies and how algorithmic advice helps people detect lies in audio-visual formats.

von Schenk et al. (2024) designed an experiment where participants were asked to write truthful and deceptive statements regarding their plans for the upcoming weekend. These written statements were employed to train a singular lie-detection algorithm, which achieved an overall accuracy rate of 67\%, a statistic not disclosed to the participants. They then investigated how often a separate set of financially motivated participants would seek AI assistance for a nominal or zero monetary fee to correctly identify truthful and false statements. They find that when the AI was accessible, only a minority of people chose to use it; however, those who do almost always follow its predictions, regardless of whether the AI identifies a statement as true or false. Although not their main focus, they discovered that the tendency to use AI was influenced by the participants' elicited beliefs about its relative performance. To the best of our knowledge, apart from our work, von Schenk et al. (2024) is the only other extant work suggesting that people's expectations about AI efficacy can affect their decision to use algorithmic advice.

Serra-Garcia and Gneezy (2024) explored the role of algorithmic advice in flagging whether a participant is likely to cooperate or defect in an American TV game show, \textit{The Golden Ball}. While also drawing on data from a popular television show, they employed a supervised machine learning (ML) algorithm, whereas our approach utilizes an AI model.\footnote{Supervised training is dependent on the availability of training data, possibly in a large number, to avoid overfitting. In contrast, we use in-context learning with a pre-trained generative AI model, where learning does not depend on the availability of training samples but can happen through a simple natural language instruction of the task provided as input. Moreover, the instruction-following AI model makes replication and future extension easier than training supervised models for further applications. See Hazra and Majumder (2024) for a comparative analysis between supervised baselines and our AI model. For a comprehensive introduction seeking to grasp the distinction between ML and AI (broader in scope), see Mullainathan and Spiess (2017) and Camerer (2019) and other chapters in the volume.} They find that the timing of these ``flags'' significantly influenced participants' ability to detect deception, with flags shown before viewing the videos being more helpful than those shown afterward. It is essential to distinguish between the act of displaying flags informed by algorithms and presenting the efficacy of AI systems through a statistic, as in our work. The flags, as in Serra-Garcia and Gneezy (2024), serve as a warning for the likelihood of cooperation or defection, not lies. The flags did not divulge to the participants the accuracy of their algorithm. In contrast, showing the accuracy of AI systems through statistics imparts insight into the algorithmic system's probable effectiveness and underlying capability. 

Leib et al. (2024) designed a between-subjects experiment to examine how advice, promoting honesty or dishonesty, affects people's propensity to lie when the advisor is an AI or a human. They also explored whether disclosing the source of the advice mattered by using a die-rolling task, $\grave{a}$ la, Fischbacher and F\"{o}llmi-Heusi (2013), thus lending objectivity to their lying activity. They find that dishonesty-promoting advice increased lying regardless of its source, whereas honesty-promoting advice did not notably enhance honesty. Moreover, knowing the source of the advice does not notably influence participants' behavior. A key difference between Leib et al. (2024) and our work is that their AI advice involves an ethical component, either encouraging lying or truthfulness, whereas our AI advice concentrates on objective truths devoid of moral undertones. Moreover, their experimental design does not allow for the observation of whether a given participant would have lied or told the truth without the AI's advice. In contrast, our design enables us to observe a given participant's guesses about the RC's identity both with and without the AI advice, thereby providing causal insights into how AI advice influences decision-making and switching behavior.

Serra-Garcia and Gneezy (2021) explored a novel experimental paradigm whereby participants (called senders) were asked to record 30-second videos making either true or false statements. These videos were shown to a separate group of financially motivated participants (called receivers) to distinguish between truthful and fake videos. They find that receivers’ ability to detect lies is limited. Yet, receivers are unaware of their limited ability to detect these lies and exhibit significant overconfidence. The obvious difference between our and their work is that we focus on text-based lies, and they do not explore the role of AI in telling apart between fake and genuine videos.


A comparatively large body of work cited in Section 1 exists in computing literature investigating algorithmic transparency, suboptimal use of AI advice, and individuals’ ability to detect lies in audio, visual, audio-visual, and textual formats. The key difference between these papers and our work is that these papers do not investigate the relationships among AI advice of varying quality, individual expectations about AI quality, and informational interventions like ours. Moreover, their setups allow motivated reasoning to affect participants' decisions, thus limiting the generalizability of their results.

Finally, although a relatively less explored game show for academic research, we are not the first to harness T4's unique aspects. Banerjee et al. (2023) used the T4 game show data to examine whether judges who participated in multiple sessions improved their ability to detect the RC. Their findings suggest that repeated exposure to similar deceptive scenarios fosters learning, indicating that experience can enhance one's capacity for deception detection. However, Banerjee et al. (2023) did not directly employ any elements of the game show to study human's ability to detect textual deception or develop algorithmic models for assisting people with deception detection.

\section{The Game Show}\label{sec:game show}

T4 features deception within a high-stakes, quasi-naturalistic setting. The game show was first aired on CBS Broadcasting Network in 1956.\footnote{The sessions of the game show are accessible on YouTube at no cost :~\url{https://www.youtube.com/watch?v=v3bSwCJD1_8&list=PL39ftvD_GHaHhv8Qm_truGRLp61iJNFOg}. The use of YouTube videos for research purposes is protected under the ``fair use'' clause, as stated in \url{https://www.youtube.com/intl/en-GB/yt/about/copyright/fair-use/}.} At the beginning of every session of the show, the host publicly announces a set of facts about the RC known from a signed affidavit. Each contestant is assigned a random player number for ease of reference: Number One, Number Two, and Number Three. The session continues with a few rounds of back-and-forth question and answer (Q\&A) between the judges and the contestants, one contestant at a time. The affidavit forms the basis for the judges' questions. In compliance with the show's rules, the RC must answer the questions truthfully. In contrast, the non-RC contestants could fabricate facts or tell outright lies to deceive the judges into thinking they are the RC. Upon completion of the Q\&As, the judges independently compile the information gathered, reconfirm it with the information from the affidavit, and reveal their independent guesses. Once all four guesses are revealed, the RC discloses their identity. After that, the host announces the sum of money each contestant receives. A typical session lasts approximately seven minutes.

The judges primarily belonged to the entertainment industry and included actors, entertainment journalists, comedians, etc. The contestants, by contrast, came from diverse backgrounds, including World War II army captains, mountain climbers, newspaper publishers, human cannonballs, stewardesses, and beauty pageant winners. Arguably, the judges were also motivated to guess the identity of the RC to boost viewership and enhance their popularity (Banerjee et al., 2023). While the judges could reappear, each contestant appeared only once in the entire game show.\footnote{Our final dataset includes 396 unique contestants and a unique set of 56 judges.}  The judges had no prior knowledge of the RC's true identity. Appendix C presents a transcript we use showing the entire conversations from a session.

\subsection{Construction of Transcripts Database}\label{sec:transcript_selection}

We transcribed several sessions of the first season of the show to create Q\&As-based text filled with deceptive content mirroring the dynamics of popular Q\&A platforms that subjects are required to analyze in our experiment to identify the RC.\footnote{We specifically utilize data from the first season because it is the only season for which the maximum number of sessions are available for download from YouTube. Later seasons, including sporadic revivals from 2016-2022 on ABC, feature less structured and uneven formats. These include additional entertainment segments, audience input on the RC's identity, less structured questioning, and occasional lack of financial compensation for contestants, introducing uncertainties about their intent to deceive.} Our data selection comprises two phases. In Phase 1, we downloaded 385 of the 429 first season sessions from YouTube.\footnote{We excluded 44 sessions due to poor recording quality or inconsistencies in session formats. One example of these inconsistencies occurred when a male and female formed a couple team representing a single RC. We also did not include sessions that featured multiple RCs.} Of these 385 sessions, we randomly selected and transcribed 195 sessions using the open-source transcription tool, Whisper.\footnote{Choosing 195 transcripts out of 385 leaves us with not only a large enough pool of transcripts to sample from but also ensures approximately an ex-ante selection probability of 0.50 $(= 195/385)$ for each transcript. We did not transcribe all 385 in the interest of time.}$^,$ \footnote{Whisper is an automated speech recognition system trained on 680,000 hours of supervised, multilingual, and multitasking online data. We used the Large model of Whisper, trained on 155 million parameters, for generating the text. Whisper produces transcripts with a word error rate of 8.81\% compared to a human word error rate of 7.61\% (Radford et al., 2023). The difference in the word error rates can be compensated by extensive manual checking, which we undertake.} 

To further screen for any plausible errors from automated transcriptions, we manually reviewed every transcript and compared it with the original video to correct any likely inconsistencies like identifiable misspelled nouns, unnecessary noise, filler words in questions asked by judges (e.g., umm, uhh-hh), incorrect sentence completions or sentence run-ons, and multi-lingual conversations beyond English.

In each session, once the identity of the RC was revealed, the judges offered explanations detailing the specific cues that led them to identify the RC. For each transcript, we categorize these explanations into three types: textual, audio-visual, and unknown. Textual cues refer to useful information drawn directly from the affidavit or Q\&As. Explanations highlighting contestants' distinct traits identified through auditory (e.g., a distinctive voice or accent) or visual observations (such as tan lines, height, or build) are classified as audio-visual cues. Finally, any unclear or absent explanations fall under the unknown category. As our investigation is focused on detecting lies from perusing textual content, any explicit audio-visual cues may hinder our participants’ ability to identify RC purely from text. Hence, we undertake a selection procedure to eliminate transcripts containing audio-visual signals from the pool of all 195 sessions. More specifically, we exclude all sessions where at least three judges (a majority) explicitly mentioned that some audio-visual cues influenced their guesses. Next, we eliminate all sessions with unknown explanations to ensure that our analysis is solely based on cases where the judges' decision-making process is supported by clear and identifiable textual cues. This selection criteria ensures that the remaining sessions primarily relied on textual cues, allowing us to focus on transcripts where text-based cues played a significant role in making a guess. This process leaves us with 132 transcripts.

Each transcript in our dataset consists of two main components: the affidavit containing the facts about the RC and the back-and-forth Q\&As between the judges and the contestants. We do not include the judges' guesses or the RC's identity, which is revealed at the end of each session. Every transcript has, on average, 12-15 Q\&A pairs. One might argue that discussions about some sessions could be found on social media platforms like Quora, Reddit, or YouTube, making it easier for our participants to identify the RC during the experimental sessions. However, upon extensive checks, we could not locate any such discussions on the internet. It is also possible that some of our participants might be familiarized with the contents of a session. To guard against such occurrences, we randomly swapped the assigned player numbers from the game show (e.g., change contestant Number One to Number Three and vice-versa), making it difficult for participants to copy the identity of the RC off the internet or guess them from memory. Identifying the RC in our setup thus involves closely assessing Q\&A pairs addressed to a contestant and evaluating the likelihood of that contestant being the RC.

\subsection{Identification of the RC using AI}\label{sec:AI}

In Phase 2, we used a generative AI model to identify the RC in each transcript. Frontier generative models such as GPT-4 (Open AI, 2023) show remarkable performance in reasoning and language understanding tasks (Hendrycks et al., 2020), making them an ideal candidate for algorithmic lie detectors from the text. We used in-context learning, where we created a set of natural language statements (prompts) informative about the rules of T4 and defined the truth detection task for the AI model. The output entails predicting who the RC is: Number One, Number Two, or Number Three. Our best model achieved an accuracy rate of 56.06\% in correctly identifying the RC for 132 transcripts (see Hazra and Majumder, 2024 for more details on the AI model).\footnote{Our model's performance is significantly superior to both the base GPT-4, with an accuracy of 46.88\%, and a random guesser. The $p$-value for two-tailed $t$-test for comparing the accuracy rate of our AI model with that of the base GPT-4 is $< 0.01$. The corresponding $p$-value for comparing the accuracy rate of our AI model with that of the random guess is also $< 0.01$.}  Furthermore, our model was able to detect deception in cases where all human judges failed to identify the RC, underscoring the potential for AI-human collaborations in improving lie detection.

\subsection{Creation of Three AI Advisors}\label{sec:AI advisor}

Finally, we created three distinct types of AI advisors: \textit{LQ}, \textit{MQ}, and \textit{HQ}. Each quality level corresponds to a distinct cluster of five transcripts. To determine the composition of these clusters, we divided the 132 transcripts into two groups: AI correct guesses (74 transcripts) and AI incorrect guesses (58 transcripts). Thereafter, we randomly drew transcripts from each group without replacement to create the three AI advisors with varying quality. For the \textit{LQ} advisor, which has an accuracy rate of 20\%, we randomly selected one transcript from the AI correct group and four from the AI incorrect group. This cluster represents an under-performing AI advisor relative to chance. The \textit{MQ} advisor has an accuracy rate of 40\%, better than random guessing. To construct this cluster, we randomly sampled two transcripts from the AI correct group and three from the AI incorrect group. The \textit{HQ} advisor is designed to have an accuracy rate of 60\%, demonstrating an accuracy rate significantly better than chance and roughly similar to the overall accuracy rate of 56.06\% achieved by our AI model. It is important to note that ex-ante, we do not know if these clusters present distinct or similar challenges in identifying the RC for our participants. Instead, they are constructed to systematically vary the efficacy of the AI advisors, allowing us to explore how disclosing different accuracy rates influences individuals' reliance on an AI advisor.

\section{A Behavioral Model}
We develop a simple behavioral framework aimed at developing testable hypotheses about how variations in the quality of AI advisors and the availability of information regarding the AI advisor's capability to identify the RC influence an individual’s likelihood of relying on the AI advisor and their expected utility from the truth detection task. We begin by considering an individual, $i$, tasked with identifying the RC while reviewing only one transcript. For simplicity, we will assume that the quality or the difficulty level of the representative transcript remains constant across all conditions studied below. If the individual correctly guesses the identity of the RC, they receive a monetary reward $x$; if incorrect, they earn $y$, where $x>y \geq 0$. Hence, the individual's expected utility from the truth-detection task is given by $E(U) = p.u(x) + (1-p).u(y)$, where $p \in [0, 1]$ represents the probability of correctly identifying the RC. Without loss of generality, we assume  $x=1$, $y = 0$, and $u(x)=1$, $u(y)=0$. Thus, the individual's expected utility simplifies to $E(U) = p$. Note that $E(U)$ thus represents the truth detection probability.

The individual $i$ knows their own probability of correctly identifying the RC from the transcript, denoted as $p_i \in [0, 1]$. Therefore, individual $i$ is aware of their own ability to identify the RC. If the individual makes a random guess, then $p_i = \frac {1}{3}$, as there are three contestants, one of whom is the RC in the transcript. Thus, the expected utility of a random guesser is $\frac {1}{3}$. This leads us to our first hypothesis.

\textbf{Hypothesis 1}. A random guesser will correctly identify the RC with probability $\frac {1}{3}$.

Suppose the individual can access an AI advisor's guess about the RC's identity at no cost. The individual has the option to either submit their own guess or use the AI's guess as their final choice, which will determine their ultimate material payoff. Suppose that the true probability of the AI advisor correctly identifying the RC is $p_a \in [0, 1]$, representing the advisor's true capability. We examine two information conditions. In the first, the individual $i$ is not provided with any information about $p_a$, denoted as the Blackbox ($BB$) condition. In the other information condition, referred to as the With Information ($WI$) condition, the individual is made aware of the AI advisor's true capability, $p_a$, when deciding whether to rely on the AI's guess or their own guess. Below, we will first analyze how variations in the value of $p_a$ influence the individual's decision to switch or not switch to the AI advisor's guess under each information condition.

Suppose that for the given transcript, $p_a$ changes to $p_a^1$ such that $p_a^1>p_a$. In the $BB$ condition, the individual receives no information about $p_a$. Therefore, in $BB$, the individual's decision to rely on the AI's guess hinges on comparing $p_i$ and their belief or expectation about the AI's ability to correctly identify the RC, denoted by $p_e \in [0, 1]$. Thus, in $BB$, the individual’s expected utility from choosing to switch ($s$) to the AI’s guess becomes $E(U_s) = p_e$. Conversely, the expected utility from choosing not to switch ($ns$), thereby relying on their own guess instead of the AI's, is given by $E(U_{ns}) = p_i$. The individual opts for $s$ if $E(U_{s}) > E(U_{ns})$, that is, if $p_e > p_i$. Similarly, the individual chooses $ns$ if $E(U_{s}) < E(U_{ns})$, that is, if $p_e < p_i$. They remain indifferent between the two choices when $E(U_{s}) = E(U_{ns})$, which occurs if $p_e = p_i$. Therefore, in the absence of any knowledge about $p_a$, a change in $p_a$'s value does not impact the individual's decision to follow the AI’s recommendation in $BB$. The decision remains solely dependent on comparing $p_i$ and $p_e$, not $p_a$. However, the realized expected utility from choosing $s$ will be $p_a$, while the realized utility from choosing $ns$ will be $p_i$. Hence, the individual's realized utility can change with a change in $p_a$ despite no change in the individual's decision to choose $s$ or $ns$. The individual's switch decision and the effect of a change in $p_a$'s value on the realized expected utility in $BB$ is summarized in Table M1 below.

\setlength{\LTpre}{0pt}
\setlength{\LTpost}{0pt}
\noindent Table M1: Switching Decisions and Expected Utilities in the $BB$ condition
\noindent
\begin{longtable}{p{.25cm} p{1.25cm} p{2.2cm} p{2.25cm} p{2.2cm} p{2.25cm} p{2.4cm}}
\hline
          &  Case              & Decision $| \ p_a$ & E(U) $| \ p_a$  & Decision $| \ p_a^1$ & E(U)  $| \ p_a^1$ & $\Delta$E(U)\\
\hline
1. & $p_e < p_i$ & $ns$ & $p_i$ & $ns$ & $p_i$ & 0\\
2. & $p_e = p_i$ & $s$ $\sim$ $ns$  & $p_i | ns$ or $p_a | s$ & $s$ $\sim$ $ns$  & $p_i | ns$ or $p_a^1 |s$ & $0|ns$ or $>0|s$\\
3. & $p_i < p_e$ & $s$ & $p_a$ & $s$ & $p_a^1$ & $>$0 \\
\hline
\multicolumn{6}{l}{\scriptsize{Note: $\Delta$E(U) = E(U)  $| \ p_a^1$ $-$ E(U) $| \ p_a$.}}
\end{longtable}
\normalsize
This leads us to our next two hypotheses. 

\textbf{Hypothesis 2a}. In the $BB$ condition, the individual's decision to switch to the AI advisor's guess will not change with an increase in the value of $p_a$.  
  
\textbf{Hypothesis 2b}. In the $BB$ condition, the individual's realized truth detection probability will increase weakly with an increase in the value of $p_a$.    

By contrast, in the $WI$ condition, the individual's decision to rely or not rely on the AI's guess is based on $p_a$ instead of $p_e$. The individual's expected utility from $s$ is $E(U_s) = p_a$, and from $ns$ is $E(U_{ns}) = p_i$. The individual will choose $s$ if $E(U_{s})>E(U_{ns})$, which occurs when $p_a > p_i$;  will opt for $ns$ if $E(U_{s}) < E(U_{ns})$, that is, if $p_a < p_i$; and will be indifferent between $s$ and $ns$ if $E(U_{s})=E(U_{ns})$, which happens when $p_a = p_i$. 

In the $WI$ condition, an increase in the AI advisor's ability to identify the RC results in one of five possible cases, listed in Table M2 below. Each case leads to two potential changes: an individual's reliance on the AI and expected utility. We define a strict increase in reliance on the AI advisor when an individual changes their switching decision from $ns$ to $s$. By contrast, a weak increase in reliance on the AI advisor occurs when an individual changes from being indifferent between $ns$ and $s$ to $s$, or from $ns$ to being indifferent between $ns$ and $s$. Finally, an individual's reliance on the AI advisor remains unchanged when their switching decisions do not vary across decision settings.

\noindent Table M2: Switching Decisions and Expected Utilities in the $WI$ condition
\noindent
\begin{longtable}{p{.8cm} p{2.5cm} p{2.25cm} p{1.6cm} p{2.25cm} p{1.6cm} p{1.5cm}}
\hline
          &  Case              & Decision $| \ p_a$ & E(U) $| \ p_a$ & Decision $| \ p_a^1$ & E(U)  $| \ p_a^1$ & $\Delta$E(U)\\
\hline
1. & $p_a < p_a^1 < p_i$ & $ns$ & $p_i$ & $ns$ & $p_i$ & 0\\
2. & $p_a < p_a^1 = p_i$ & $ns$ & $p_i$ & $s$ $\sim$ $ns$ & $p_i=p_a^1$ & 0\\
3. & $p_i =p_a < p_a^1$ & $s$ $\sim$ $ns$ & $p_i=p_a$ & $s$ & $p_a^1$ & $>$0 \\
4. & $p_i < p_a < p_a^1$ & $s$ & $p_a$ & $s$ & $p_a^1$ & $>$0\\
5. & $p_a  < p_i < p_a^1$ & $ns$ & $p_i$ & $s$ & $p_a^1$ & $>$0\\
\hline
\multicolumn{7}{l}{\scriptsize{Note: $\Delta$E(U) = E(U)  $| \ p_a^1$ $-$ E(U) $| \ p_a$.}}\\
\end{longtable}
\noindent
\normalsize
As illustrated by the first two cases above, when $p_a$ increases to $p_a^1$ but still falls short of $p_i$ (Case 1) or barely equals $p_i$ (Case 2), the individual exhibits no change or only a weak increase in reliance on the AI advisor. Despite showing a weak preference for the AI's guess, there is no change in the individual's expected utility (i.e., $\Delta E(U) = 0$)  or the overall accuracy of the truth-detection rate. In contrast, for the remaining cases where $p_a^1$ surpasses $p_i$, there is a strict increase in the individual’s expected utility or the overall truth-detection rate. However, reliance on the AI advisor differs based on how $p_i$ compares to $p_a$. In Case 3, where $p_a^1$ exceeds $p_i$ but $p_a=p_i$, there is a weak increase in AI reliance. In Case 4, where both $p_a$ and $p_a^1$ exceed $p_i$, there is no change in AI reliance. In Case 5, where $p_a$ is below $p_i$ but $p_a^1$ exceeds $p_i$, there is a strict increase in AI reliance. In sum, in the presence of information about AI's actual accuracy, the individual exhibits a weak preference for seeking the AI's advice, and the expected utility also registers a weak increase with an increase in $p_a$. This observation leads us to our next two hypotheses.

\textbf{Hypothesis 3a}. In the $WI$ condition, the individual will weakly prefer to switch to the AI advisor's guess with an increase in the value of $p_a$.
  
\textbf{Hypothesis 3b}. In the $WI$ condition, the individual's realized truth detection probability will increase weakly with an increase in the value of $p_a$.   

Next, we analyze, for a given $p_a$ value, how disclosing information about $p_a$ affects the individual's decision to switch to the AI's guess compared to the $BB$ condition. Depending upon the relative magnitudes of $p_i$, $p_e$, and $p_a$, one of the following cases listed in Table M3 will arise. As before, each case will lead to two key outcomes: a change in the individual’s reliance on the AI and a corresponding shift in expected utility.

\noindent Table M3: Switching Decisions and Expected Utilities in the $BB$ and $WI$ conditions
\noindent
\begin{longtable}{p{.8cm} p{2.5cm} p{2.25cm} p{2.5cm} p{2.25cm} p{1.6cm} p{1.5cm}}
\hline
   & Case & Decision $| \ p_e$ & E(U) $| \ p_e$  & Decision $| \ p_a$ & E(U) $| \ p_a$ & $\Delta$E(U)\\
\hline
1. & $p_a < p_e < p_i$ & $ns$ & $p_i$ & $ns$ & $p_i$ & 0\\
2. & $p_e \leq p_a < p_i$ & $ns$ & $p_i$ & $ns$ & $p_i$ & 0\\
3. & $p_a = p_e = p_i$ & $s$ $\sim$ $ns$ & $p_i=p_a$ & $s$ $\sim$ $ns$ & $p_i=p_a$ & 0\\ 
4. & $p_i < p_a < p_e$ & $s$ & $p_a$ & $s$ & $p_a$ & 0\\
5. & $p_i < p_e \leq p_a$ & $s$ & $p_a$ & $s$ & $p_a$ & 0\\
6. & $ p_e < p_i < p_a$ & $ns$ & $p_i$ & $s$ & $p_a$ & $>$ 0\\
7. & $p_i = p_e < p_a$ &  $s$ $\sim$ $ns$ & $p_i|ns$ or $p_a|s$ & $s$ & $p_a$ & $\geq$0\\
8. & $p_e < p_a = p_i$ & $ns$ & $p_i$ & $s$ $\sim$ $ns$  & $p_i=p_a$ & 0\\
9. & $p_a<p_i < p_e$ & $s$ & $p_a$ & $ns$ & $p_i$ & $>$0\\
10. & $p_a < p_e = p_i$ & $s$ $\sim$ $ns$ & $p_i|ns$ or $p_a|s$ & $ns$ & $p_i$ & $\geq$0\\
11. & $p_i = p_a < p_e$ & $s$ & $p_a$ & $s$ $\sim$ $ns$ & $p_i=p_a$ & 0\\
\hline
\multicolumn{7}{l}{\scriptsize{Note: $\Delta$E(U) = E(U)  $| \ p_a$ $-$ E(U) $| \ p_e$.}}\\
\end{longtable}

\normalsize
We group these cases based on whether the individual changes or does not change their switching decision between the $BB$ and $WI$ conditions. The first five cases demonstrate situations where the individual does not change their switching decision between the $BB$ and $WI$  conditions. In Cases 1 and 2, both $p_a$ and $p_e$ are strictly less than $p_i$, leading the individual to rely on their own guess in both information conditions. In Case 3, $p_a$ and $p_e$ are identical to $p_i$, leaving the individual indifferent between $s$ and $ns$ in both information conditions. In Cases 4 and 5, both $p_a$ and $p_e$ exceed $p_i$, resulting in the individual relying on the AI's guess in both information conditions. Therefore, in these five cases, the individual exhibits no change in reliance on the AI advisor between the $BB$ and $WI$ conditions. Consequently, the expected utility or the truth-detection rate registers no change.  This observation leads us to our next hypotheses.

\textbf{Hypothesis 4.1a}. The individual's switching decision between the $BB$ and $WI$ will remain the same if both $p_a$ and $p_e$ are greater than, less than, or equal to $p_i$.

\textbf{Hypothesis 4.1b}. The individual's realized truth detection probability between the $BB$ and $WI$ will remain the same if both $p_a$ and $p_e$ are either greater than, less than, or equal to $p_i$.   

In Cases 6, 7, and 8, the individual adjusts their switching decision as the information environment shifts from $BB$ to $WI$, specifically showing a strict or weak preference for $s$ over $ns$. For instance, in Case 6, since $p_e$ is strictly less than $p_i$, the individual chooses to submit their own guess in $BB$. However, when information about the AI advisor's capability is available, and $p_a$ is strictly greater than $p_i$, the individual strictly prefers to submit the AI's guess instead of their own. A similar decision principle sways the individual to weakly prefer $s$ over $ns$ in Cases 7 and 8. Therefore, in these three cases, the individual exhibits a weak preference for the AI's guess in the $WI$ compared to the $BB$ condition. Consequently, the expected utility or the truth-detection rate registers a weak increase. This observation leads us to our next hypotheses.

\textbf{Hypothesis 4.2a}.  The individual's switching decision will change, weakly or strictly preferring $s$ over $ns$ when moving from $BB$ to $WI$ if $p_e < p_a$ and $p_i \leq p_a$.

\textbf{Hypothesis 4.2b}.  The individual's realized truth detection probability will weakly increase when moving from $BB$ to $WI$ if $p_e < p_a$ and $p_i \leq p_a$.

In Cases 9, 10, and 11, the individual again modifies their switching decision as the information environment transitions from $BB$ to $WI$, specifically demonstrating a strict or weak preference for $ns$ over $s$. For instance, in Case 9, since $p_e$ is strictly greater than $p_i$, the individual opts to submit the AI's guess. However, when information about the AI advisor's capability is available, and $p_a$ is strictly less than $p_i$, the individual strictly prefers to submit their own guess instead of the AI's. A similar decision-making principle leads the individual to weakly prefer $ns$ over $s$ in Cases 10 and 11. Thus, in these three cases, the individual shows a weak preference for their own guess in the $WI$ compared to the $BB$ condition. Consequently, this results in a weak increase in the expected utility or truth-detection rate. This observation leads us to our next hypotheses.

\textbf{Hypothesis 4.3a}.  The individual's switching decision will change, weakly or strictly preferring $ns$ over $s$ when moving from $BB$ to $WI$ if $p_a<p_e$ and $p_i \geq p_a$.

\textbf{Hypothesis 4.3b}.  The individual's realized truth detection probability will weakly increase when moving from $BB$ to $WI$ if $p_a<p_e$ and $p_i \geq p_a$.

As mentioned earlier, our experiment incorporates two information conditions: $BB$ and $WI$. By design, $p_a \in \{0.2, 0.4, 0.6\}$ in our experiment. We also elicit participants' $p_e$. With these experimental parameters in place, we proceed to test the above hypotheses.

\section{Experimental Design and Procedure}\label{sec:exp_design}

Our experiment induces a $2\times3$ between-subjects design with six treatments. The two information conditions, $BB$ and $WI$, represent the absence and presence of AI accuracy information, respectively. Within each information condition, we vary the AI advisor accuracy with \textit{LQ = 0.2}, \textit{MQ = 0.4}, and \textit{HQ = 0.6}. The treatments are thus labeled as \textit{$BB_{LQ}$}, \textit{$BB_{MQ}$}, \textit{$BB_{HQ}$}, \textit{$WI_{LQ}$}, \textit{$WI_{MQ}$}, and \textit{$WI_{HQ}$}. A total of 545 participants took part in the experiment, conducted over several days on the Prolific online platform, with the following distribution: 91 in \textit{$BB_{LQ}$}, 89 in \textit{$BB_{MQ}$}, 94 in \textit{$BB_{HQ}$}, 91 in \textit{$WI_{LQ}$}, 90 in \textit{$WI_{MQ}$}, and 90 in \textit{$WI_{HQ}$}. Each participant was randomly assigned to only one treatment.\footnote{We pre-registered the trial on the AEA RCT Registry. Pre-registration details of the trail can be accessed here: ~\url{https://www.socialscienceregistry.org/trials/12535}.}

We used oTree (Chen et al., 2016) to develop the software program for the experiment. Since T4 featured contestants and judges predominantly from the United States and included sessions with historical, social, and cultural references pertinent to the US, we filtered participants from the Prolific pool based on nationality and place of residence, both set to the US, to ensure familiarity with such references. The Prolific platform enabled us to screen participants according to our geographical requirements.

Participation in the experiment was voluntary. Participants were required to provide informed consent before proceeding. Participants who did not consent exited the experiment and did not receive any payment. Participants consented to participate in the experiment were presented with the instructions for Task 1. After reading the instructions of Task 1, participants were required to answer three screening questions. These questions were designed to assess the participants’ understanding of Task 1 instructions. Correctly answering these questions was a prerequisite for moving forward with Task 1. Notably, no participant failed the screening test and exited the experiment.\footnote{The instructions are available in Appendix D.}

\subsection{Tasks in the $BB$ condition}
The $BB$ condition is grounded in the rationale that end users often do not have access to the true effectiveness of AI systems, which can lead to either overreliance on or insufficient trust in the AI's recommendations (Vasconcelos et al., 2023; Zhang et al., 2020; Dietvorst et al., 2015). In contrast, by disclosing the AI's efficacy, as in the $WI$ condition, individuals are empowered to calibrate their trust and reliance on the AI’s counsel more appropriately. Hence, the $BB$ condition serves as a benchmark for examining the dependence on AI when its efficacy is unknown against when it is transparent.

Each participant completed five tasks in the $BB$ condition. The specifics of each task were not disclosed beforehand. Task 1 was divided into two sub-tasks. In sub-task 1, each participant was randomly assigned one of the three AI-advisor qualities. For each of the five transcripts in that cluster, the participant was required to guess which of the following three options represented the RC: Number One, Number Two, or Number Three. We randomized the sequence in which participants encountered the transcripts within a given cluster. Their guess for each transcript was recorded and referred to as their initial guess. Regardless of accuracy, participants received a fixed payment of \$0.75 for each of the five initial guesses. In sub-task 2, participants indicated their level of absolute confidence for each of the five initial guesses on a scale from 0 (not confident at all) to 100 (completely confident). We used the quadratic scoring rule (QSR) (Charness et al., 2021) to financially incentivize accurate absolute confidence reporting. Participants earned higher rewards for accurate guesses when their confidence was high, and for incorrect guesses when their confidence was low.\footnote{For a confidence level $c \in [0, 100]$, a participant's belief about the chance of correctly identifying the RC is $c/100 = p$, and the payment is determined as follows: $Payment = A - B(1-p)^2$, if initial guess is correct; else, $Payment = A - Bp^{2}$, where $A > 0$ is 0.99559 ($\sim 1$) and $B > 0$ is 0.99564 ($\sim 1$).} For example, a participant could earn up to \$1.00 if they were completely confident that their guess was correct.\footnote{For instance, if the initial guess is correct and $c$ is 100 (i.e., $p = 1$), the payment is $\sim$ \$1.00.} By default, the confidence level was set at 50 in the experiment, which resulted in a payment of $\$0.75$. Participants could drag a slider to choose their desired level of absolute confidence for each initial guess.

Task 2 consisted of two sub-tasks. In each sub-task, participants answered one question per transcript. The first sub-task required participants to classify the difficulty level of each transcript as Easy, Medium, or Difficult. This sub-task was unpaid. In the second sub-task, participants predicted the difficulty level they thought the majority would assign to each transcript. Participants earned a \$0.50 bonus if their prediction for a randomly selected transcript matched the majority’s choice; otherwise, they received \$0.

Task 3 also included two sub-tasks. Before the first sub-task, participants were informed that an AI also guessed the identity of the RC without divulging any other details of the AI. Specifically, we used the following text to inform participants of the availability of the AI's guesses in this condition: ``Before conducting this experiment, we asked an artificial intelligence (AI) system to read and make guesses for the same five sets for which you made guesses in Task 1. We collected the AI's guess regarding who the real John or Jane Doe is for each of those five sets.'' In the first sub-task, participants were  sequentially shown the AI’s guess for each of the five transcripts. For each transcript, the computer screen presented them with two guesses: their guess and the AI advisor's guess for that transcript. Participants must then choose whether to submit their initial guess or the AI’s guess for each transcript, with no additional cost for this choice. A switch occurs when participants opt to submit the AI’s guess instead of their own initial guess as their final guess. Participants could earn a \$3.00 bonus if their final guess for the randomly selected transcript were correct. In the second sub-task, they could update their absolute confidence using the same QSR scheme upon submitting their final guess.\footnote{Note that in both $BB$ and $WI$ information conditions, participants were not informed whether the AI's guesses were correct or incorrect for any particular transcript.}

In Task 4, we elicited participants’ relative confidence in their Task 1 performance. Specifically, we asked participants to evaluate their relative performance in identifying the RC by letting them select one of the four categories that they believed best represented how well their performance ranks in a group of 100 peers.
They had four options: rank themselves among the top 25 participants, those ranked between 26th and 50th, those between 51st and 75th, or among the bottom 25 participants. Upon completing all the sessions, we ranked the participants based on their accuracy rates. If a participant correctly identified their relative performance that matched their actual rank among all, they earned a bonus of \$0.50. However, participants received \$0 in case of no match.

Task 5 involved completing an exit questionnaire (see D.13 in Appendix D) that gathered demographic information (such as gender, age, and education), details on how participants made their guesses (whether randomly or based on prior knowledge), their familiarity with the game show, and whether they had watched any sessions related to the transcripts. Participants received a \$0.50 bonus for completing Task 5. In total, a participant could earn a guaranteed amount of \$4.25 (\$0.75 for each transcript and \$0.50 for the exit questionnaire), with the possibility of an additional maximum payment of \$5.00 in the BB condition. The participants were paid after the completion of Task 5. After completing all five tasks, the computer randomly selected a transcript, and participants were informed whether they had correctly guessed the RC for that specific transcript and the total amount of money they earned in the experiment. 

\subsection{Tasks in the $WI$ condition}
Each participant in $WI$ also completed the same five tasks as in $BB$. The $WI$ information condition differs from the $BB$ condition in one key way. In $WI$, participants were informed of the AI's accuracy rate before submitting their final guess for each transcript in Task 3. For example, participants in the \textit{LQ} case were told, ``The AI correctly identified the real John or Jane Doe in one of the five sets, which is equal to a 20\% accuracy rate." The $WI$ condition is based on the idea that as participants learn about the AI system's efficacy, they may adjust their reliance on its recommendations. 


Note that our setup generates five guesses and five switching decisions per subject. For example, with 91 subjects in \textit{$BB_{LQ}$}, we have 455 observations for guesses and switches in this treatment.\footnote{Our design demonstrates sufficient power to detect meaningful treatment effects. Among the six treatments, the \textit{$BB_{MQ}$} treatment has the smallest sample size, with 445 observations on guesses. With 445 observations, the power of the test is 0.871 when $H_0$: $p = 0.33$ (the accuracy rate for a random guesser) against $H_1$: $p = 0.4$, using $\alpha = 0.05$. Given the three AI accuracy rates of 0.2, 0.4, and 0.6, the choice of $p = 0.4$ as our specific alternative is particularly relevant as it is closest to the accuracy rate of a random guesser. Therefore, even with the smallest sample size in our experiment, the power to detect relevant treatment effects is higher than the desirable level of 0.8.} Before exiting the experiment, participants had to submit a Prolific completion code. Throughout the experiment, we took great care to avoid introducing any elements that might artificially influence the participants' trust in the AI. In particular, we consistently referred to the underlying algorithmic system simply as ``the AI" and made a conscious effort to avoid anthropomorphizing it.\footnote{External influences on trust could skew people's reliance on AI advice. Anthropomorphizing technology or giving it human-like features often inflates trust artificially (Vasconcelos et al., 2023; Salles et al., 2020). This could bias our findings by introducing a variable that alters participants' perception of the AI's reliability. For an overview of the established relationship between trust and reliance on technology, see Bansal et al. (2021), Bussone et al. (2015), and Dietvorst et al. (2015).} Overall, participants took an average of 29 minutes to complete the experiment. 50.09\% of the participants were female, 47.34\% were 35 years old or younger, and 61.8\% held at least a college degree.


\subsection{Eliciting Participants' Beliefs about AI Advisor's Efficacy}

We elicited participants’ expectations or beliefs about the AI advisor’s ability to identify the RC. We asked participants the following question as part of Task 5 (see the eighth question under Task 5 in Appendix D): ``Out of the five sets, how many of the AI's guesses do you think are correct?'' Participants were not specifically incentivized to report their beliefs, aside from receiving a flat bonus of \$0.50 for completing Task 5. The literature on the relationship between monetary rewards and the quality of elicited beliefs is mixed, influencing our decision not to pay participants for the belief question.\footnote{Charness et al. (2021) conclude that ``Which approaches produce the highest accuracy regarding beliefs, and at what cost? The evidence does not show significant differences in this regard between introspection and the more complex methods. However, our conclusion that complex methods have not been so effective should not be understood as a triumph of introspection, or any form of endorsement of the idea that monetary incentives do not work in experiments more broadly. It should be understood as the failure of currently-used complex methods to outperform a flawed, but simple alternative.''}  

As evident, we collected these beliefs after participants had engaged with the AI advisor in both information conditions. Eliciting their beliefs at the end in the $BB$ condition made sense because we never informed participants about the number of correct guesses made by the AI in any of the treatments. In contrast, in the $WI$ condition, participants were already aware of the AI advisor's accuracy, which rendered the belief question redundant. However, we elicited their beliefs in the $WI$ condition to maintain procedural consistency between the conditions. Additionally, we chose to gather these beliefs in the exit questionnaire rather than at the beginning of Task 3 before participants were informed of the AI's guesses and provided a choice to switch to the AI's guesses. We speculated that asking participants about the AI’s performance prior to their switching decision might affect their subsequent switching behavior. This is because existing research (e.g., Schlag et al., 2015) suggests that individuals often adjust their actions retrospectively to align with their reported beliefs, which could bias the switching data — an essential variable for measuring individuals' reliance on the AI advisor in our context.

To address the potential influences of the timing of the belief question and financial incentives on belief data, we implemented a second belief elicitation scheme. The second scheme, implemented for both information conditions, referred to as the Modified Blackbox ($MBB$) and Modified With Information ($MWI$), elicited participants' beliefs just before Task 3. Under this scheme, participants were promised a \$1.50 reward if their expectations about how many guesses their AI advisor got right turned out correct, and \$0 otherwise. The wording of the belief-elicitation question remained the same as in the first scheme. We recruited 30 participants for each AI quality advisor under the $MBB$ and $MWI$ conditions. The other details of the sessions in these conditions were identical to those in the main $BB$ and $WI$ conditions. Subsection 6.3  examines whether the belief data elicited through the two schemes differ substantially between the main and modified treatments for a given AI advisor quality.

\section{Results} \label{sec:Results}
We begin by defining several key terms we use in the data analysis. An accurate guess occurs in our experiment when a participant correctly identifies the RC from a transcript. Initial accuracy refers to the accuracy rate computed from a participant's own five guesses made before being informed of the availability of an AI and its guesses. Final accuracy, by contrast, is the accuracy rate computed from the final five guesses submitted by a participant, which may consist of their own five guesses, the AI's five guesses, or a mix of their own and the AI's guesses. When a participant submits the AI's guess as their final answer instead of their own, and the AI's guess differs from the participant's guess, we call this a `switch.' The switching decisions provide us with a measure of a participant's degree of reliance on the AI. Participants might also choose to submit the AI's guess even when their own guess matches the AI's, as they are indifferent between the two. However, such a submission does not indicate true reliance on the AI in a strict sense, so we do not categorize it as a switch.\footnote{A more lenient definition of AI reliance would consider submission of the AI's guess as a `switch' irrespective of whether the participant's guess matched the AI's guess or not. Even with this more lenient definition, we find qualitatively similar results regarding the effect of AI reliance on truth detection, as we do with the stricter definition. Since the stricter definition of switch provides a conservative measure of AI reliance, we present the results based on the stricter definition.}        

Our analysis proceeds as follows. We first assess the initial accuracy to gauge the participants’ intrinsic ability to discern the truth from the transcripts. Next, we scrutinize the switch rates to evaluate the participants’ reliance on AI. Finally, we consider the implications of switches for truth detection by analyzing whether the final accuracy shows an improvement or decline compared to the initial accuracy. 

\subsection{Initial Accuracy} 

The participants' initial accuracy rate aggregated across all six treatments is roughly 35 percent, significantly higher than the 33 percent accuracy rate expected from a random guess (see the first row of the third panel for Initial accuracy rate in Table 1). It implies that the aggregate data does not lend support to Hypothesis 1. When we compute the combined initial accuracy rate for $BB_{HQ}$ and $WI_{HQ}$, the rate rises to about 37 percent, and for $BB_{MQ}$ and $WI_{MQ}$, it increases further to 41 percent; both are significantly higher than the 33 percent theoretical benchmark. In contrast, the combined initial accuracy rate for $BB_{LQ}$ and $WI_{LQ}$ is roughly 28 percent, which is significantly lower than 33 percent (see the third panel for Initial accuracy rate in Table 1).\footnote{Similar patterns are observed for the initial accuracy rates within $BB$ and $WI$, albeit weakly in two instances (see first and second panels for Initial accuracy rate in Table 1). Specifically, the initial accuracy rate for $WI_{HQ}$ is numerically higher than but statistically equivalent to 33 percent, and the initial accuracy rate for $BB_{LQ}$ is numerically lower than but statistically equivalent to 33 percent.} Therefore, overall, participants demonstrated a better-than-chance ability to identify text-based lies. This trend persists across the clusters of transcripts associated with the MQ and HQ advisors. However, participants demonstrated a worse-than-chance ability to identify text-based lies for the cluster of transcripts corresponding to the LQ advisor. In other words, participants' initial accuracy rate is different from a random guess in each of the three transcript clusters, and as a result, Hypothesis 1 is rejected.

Within each information condition, the initial accuracy rate is significantly lower in LQ than in MQ or HQ, while it is equivalent between MQ and HQ (see the first and second panels for the Initial accuracy rate in Tables 1 and 2). Hence, in the cluster of transcripts where AI accuracy is the lowest (LQ), participants' accuracy is also the lowest. 

Next, we compare the initial accuracy rates between the information conditions. The initial accuracy rate is approximately 36 percent in $BB$ and 35 percent in $WI$ (see the first row of the first and second panels for the Initial accuracy rate in Table 1). These rates are statistically equivalent (see the first row of the third panel for the Initial accuracy rate in Table 2). Furthermore, the initial accuracy rates remain statistically equivalent between $BB$ and $WI$ for any given AI quality — LQ, MQ, and HQ (see the third panel for the Initial accuracy rate in Table 2). This consistency is expected, as participants made their choices before being exposed to different information environments. A broader implication of this uniformity in initial accuracy is that the participants' innate ability to discern the truth does not vary between the two information conditions for any given AI quality. As a result, variations in the participants' subsequent decisions across the two information conditions cannot be attributed to differences in their inherent ability to detect text-based lies. In other words, differences in the participants' reliance on the AI advisor in the two information conditions for a given AI quality stem from the information they received about the AI's efficacy rather than their own truth-detection abilities.

\subsection{Switch Rates} 
We now analyze the extent to which participants relied on the AI advisors by opting to switch to the AI's guess when their own guess differed from the AI's guess. The aggregate switch rate in the experiment is approximately 21 percent (see the first row in the third panel for the Switch rate in Table 1). Participants switched to the AI's guess in roughly 24 percent of the transcripts in $BB$ and 18 percent in $WI$ (see the first row in the first and second panels for the Switch rate in Table 1). The switch rate is significantly lower in $WI$ than $BB$ (see the first row in the third panel for the Switch rate in Table 2). This suggests that, overall, participants were more inclined to rely on AI when they lacked information about the AI's accuracy than when such information was available. 

The switch rates are 22, 24, and 25 percent for $BB_{LQ}$, $BB_{MQ}$, and $BB_{HQ}$, respectively (see Figure 1 or the first panel for Switch rate in Table 1). These switch rates are statistically equivalent (see the first panel for Switch rate in Table 2). The equivalence between the switch rates across all $BB$ treatments suggests that without information about the quality of the AI advisor, participants exhibit similar reliance behavior across all three AI advisors. Therefore, the data validates Hypothesis 2a.

The switch rates are 14, 17, and 23 percent for $WI_{LQ}$, $WI_{MQ}$, and $WI_{HQ}$, respectively (see Figure 1 or the second panel for Switch rate in Table 1). While the switch rates are equivalent for $WI_{LQ}$ and $WI_{MQ}$, the switch rate is significantly higher in $WI_{HQ}$ than in $WI_{LQ}$ and $WI_{MQ}$ (see the second panel for Switch rate in Table 2). This suggests that when information about the AI's quality is disclosed, there is a significant increase in reliance on a high-quality AI advisor compared to relatively low-quality AI advisors. As a result, the data confirms Hypothesis 3a. 

A comparison of switch rates between the information conditions for a given AI quality reveals that the switch rate is significantly lower in $WI_{LQ}$ compared to $BB_{LQ}$, and also lower in $WI_{MQ}$ compared to $BB_{MQ}$. However, the switch rate is equivalent between $WI_{HQ}$ and $BB_{HQ}$  (see Figure 1 and the third panel for the Switch rate in Table 2). This indicates that while participants rely less on the $LQ$ and $MQ$ advisors when provided information about the AI advisor's capabilities to detect the RC, such information does not affect their reliance on the $HQ$ advisor.

Per our behavioral model, participants' expectations regarding AI accuracy can explain why the reliance on AI advisors decreases when participants receive information about AI quality for $LQ$ and $MQ$, while it remains unchanged for $HQ$. However, before we examine participants' beliefs about the capability of AI advisors to explain differences in switch rates between information conditions, we note that these differences hold even if we account for other factors that may influence the switching decision.   

A participant's confidence in their own guess and the perceived difficulty of a given transcript will likely influence their decision to switch to the AI's guess.\footnote{Summary statistics of participants' confidence in own guess and perceived difficulty are provided in Table 3, and comparative tests between treatments are presented in Table 4.} We find that a participant's likelihood of switching to AI's guess decreases significantly with an increase in the participant's confidence in their own guess, and increases significantly with an increase in the perceived difficulty level of a transcript, which is intuitive (see column (1) in Table 5). We also find that even after controlling for the participant's confidence in their own guess and the perceived difficulty level, the observed differences in switch rates between the information conditions persist, i.e., while the switching probability is significantly lower in $WI$ than in $BB$ for $LQ$ and $MQ$, the switching probability is statistically equivalent between $WI$ and $BB$ for $HQ$ (see the last panel of column (3) in Table 5).\footnote{The observed sign and significance of the differences in switch rates between the treatments within each information condition also persist. The only exception is the difference in switch probability between $WI_{MQ}$ and $WI_{HQ}$ with $0.05<$ $p$-value $<.10$, which makes the difference statistically weaker or insignificant (see the second and third panels of column (3) in Table 5).} We obtain similar results regarding differences in switch rates between the treatments when we control for participants' relative confidence and perceived modal difficulty (see the last three panels of column (4) in Table 5). Thus, the observed differences in switch rates between the information conditions are robust regardless of how we measure participants' confidence and perceived difficulty.\footnote{The observed differences in switch rates between the information conditions persist even if we include indicators of participant's demographic attributes (gender, age, education) as additional control variables. The regression results are included in Table E1 in Appendix E.}    

Next, we explore participants' expectations about the capability of AI advisors.

\subsection{Expectations about the Capability of AI Advisors} 
We elicited participants' beliefs about the AI advisors' ability to identify the RC at the end of the $BB$ treatments as part of the exit questionnaire. We first examine whether participants' expected AI accuracy indeed influenced their decision to switch to the AI's guess. We find that controlling for a participant's confidence in their own guess, perceived difficulty, and indicator of the cluster of transcripts that an AI advisor represents, a participant's expected AI accuracy is a significant predictor of the likelihood to switch in $BB$ (see column (5) in Table 5). More specifically, as the expected AI accuracy increases, so does the probability of switching to the AI's guess.\footnote{Expected AI accuracy is found to increase the probability of switching to the AI's guess if we control for relative confidence and perceived modal difficulty instead (see column (6) in Table 5). Thus the result is robust to the alternative measures of confidence and difficulty.} Therefore, the premise of our behavioral model that in the absence of AI accuracy information, an individual's expectation about AI accuracy plays an influential role in shaping their decision to switch to the AI advisor's guess appears well-supported.

The summary statistics reveal that the median and approximate mean of participants' expected AI accuracy is 3 out of 5 transcripts (i.e., 60 percent) across all three treatments -- $BB_{LQ}$, $BB_{MQ}$, and $BB_{HQ}$ (see the first panel for Expected AI accuracy in Table 3). As a result, on average, while the participants' expected AI accuracy is significantly higher than the actual AI accuracy in $BB_{LQ}$ and $BB_{MQ}$, the participants' expected AI accuracy is equivalent to the actual AI accuracy in $BB_{HQ}$ (see the first panel in Table 6).  

Recall that in the behavioral model, we denote the AI's actual accuracy as $p_a$, and the individual's expectation about AI accuracy as $p_e$. Therefore, on average $p_e = p_a$ for $HQ$. If we interpret a participant's confidence in their own guess as a measure of $p_i$, then on average $p_i = p_a$ for ${HQ}$, as the mean confidence in $BB_{HQ}$ (as well as in $WI_{HQ}$) is approximately 60 percent (see the first two panels in Table 3, and Table 7). Thus, on average, $p_i = p_a = p_e$ for ${HQ}$. Therefore, since $p_e = p_a = p_i$ for $HQ$, participants' switch rate does not change between $BB$ and $WI$, in agreement with Hypothesis 4.1a. In other words, when AI accuracy is revealed and aligns with the participants' expectations, as is the case for $HQ$, participants do not feel the need to alter their AI reliance.

The expected AI accuracy statistics also imply that $p_a < p_e$ for $LQ$ and $MQ$, on average. Furthermore, participants' mean confidence in their own guess is significantly higher than the $LQ$ advisor's accuracy rate of 20 percent in $BB_{LQ}$ (as well as in $WI_{LQ}$) and significantly higher than the $MQ$ advisor's accuracy rate of 40 percent in $BB_{MQ}$ (as well as in $WI_{MQ}$) (see the first two panels of Table 3, and Table 7). Thus, on average,  $p_a<p_i$, for $LQ$ and $MQ$. Therefore, since $p_a < p_e$ and $p_a < p_i$ for both $LQ$ and $MQ$, the participants' switch rate declines in $WI$ compared to $BB$, in agreement with Hypothesis 4.3a. In other words, when the AI's accuracy is revealed to be lower than anticipated, as is the case for $MQ$ and $LQ$, participants reduce their reliance on the AI advisor.\footnote{In our experiment, we do not observe the case where $p_e<p_a$ on average. Therefore, we do not have data to test Hypothesis 4.2.}  

It is also important to account for the potential impact of participants' prior decisions, particularly their decision to switch, on their reported beliefs about AI accuracy that were elicited at the end of the experiment. This issue is a common challenge in experimental design, where earlier actions might shape responses to subsequent belief-related questions, or initial belief responses could influence later decisions. To mitigate this concern, we implemented additional treatments, $MBB$ and $MWI$, in which participants' beliefs about AI accuracy were elicited before they were shown the AI’s guesses and financial incentives were linked to their belief responses.

We find that the median and the approximate mean of participants' expected AI accuracy are similar across these additional treatments ($MBB_{LQ}$, $MBB_{MQ}$, $MBB_{HQ}$, $MWI_{LQ}$, $MWI_{MQ}$, $MWI_{HQ}$), with participants expecting that AI would make correct guesses in 3 out of 5 transcripts, i.e., 60\% accuracy rate (see the last two panels in Table 3). Similar to $BB$, we find that while the participants' expected AI accuracy is significantly higher than the actual AI accuracy in $MBB_{LQ}$ and $MBB_{MQ}$ (as well as $MWI_{LQ}$ and $MWI_{MQ}$), the participants' expected AI accuracy is equivalent to the actual AI accuracy in $MBB_{HQ}$ and $MWI_{HQ}$ (see the last two panels in Table 6). Furthermore, there is no significant difference in the distribution of expected AI accuracy in $MBB$ or $MWI$ treatments compared to corresponding $BB$ treatments (see Table 8). Based on these findings, we conclude that overall, the participants' reported beliefs about AI accuracy remain similar irrespective of the timing of the belief question or the financial incentives provided for reporting the beliefs. Thus, it is reasonable to use participants' stated beliefs about AI accuracy to interpret their switching behavior.\footnote{We do not provide a detailed analysis of the modified treatments, as they are not our main focus, and we have fewer data points from these treatments. Moreover, the primary goal of the modified treatments was to assess whether participants’ elicited beliefs about AI accuracy were sensitive to the timing and financial incentives. Yet, it is important to note that eliciting beliefs before the switch decision, specifically before Task 3, significantly impacts the switch rates and the difference between the final and initial accuracy rates for the $LQ$ advisor (see Table E2 in Appendix E). This suggests that the timing of belief elicitation can affect decisions for certain AI advisor qualities, which aligns with our speculation discussed in subsection 5.3. On the whole, our decision to elicit beliefs at the end of the experiment as part of the exit questionnaire, as in our main treatments, helps minimize any unintended effects of elicited beliefs on the reliance on AI advisors.}

Next, we shed light on the question -- do the switches to the AI's guesses improve the participants' ability to detect the RC? To answer this, we analyze the final accuracy.

\subsection{Final Accuracy}
The final accuracy rate is approximately 37 percent in $BB$ and 38 percent in $WI$ (see the first row in the first and second panels for the Final accuracy rate in Table 1). These rates are statistically equivalent (see the first row in the third panel for the Final accuracy rate in Table 2). This suggests that the higher reliance on AI observed in $BB$ compared to $WI$ does not improve final accuracy. 

The final accuracy rates are approximately 26, 38, and 48 percent for $BB_{LQ}$, $BB_{MQ}$, and $BB_{HQ}$, respectively; the corresponding statistics are 25, 44, and 46 percent for $WI_{LQ}$, $WI_{MQ}$, and $WI_{HQ}$, respectively (see Figure 2 or the first and second panels for Final accuracy rate in Table 1). 

Within each information condition, the final accuracy rate significantly improves as the quality of AI advisors increases, with the sole exception being $WI_{MQ}$ and $WI_{HQ}$, where the final accuracy rates remain statistically equivalent (see the first and second panels for Final accuracy rate in Table 2). This suggests that, overall, the final accuracy rate tends to improve, albeit weakly, as AI quality increases within each information condition. These findings thus validate Hypothesis 2b and Hypothesis 3b.

A comparison of the final accuracy rates between information conditions reveals the following. First, the final accuracy rate is statistically equivalent between $BB_{HQ}$ and $WI_{HQ}$ (see the third panel for the Final accuracy rate in Table 2). This finding aligns with Hypothesis 4.1b. Second, while the final accuracy rate is equivalent between $BB_{LQ}$ and $WI_{LQ}$,  it is higher in $WI_{MQ}$ than in $BB_{MQ}$ (see the third panel for Final accuracy rate in Table 2). These findings align with Hypothesis 4.3b. 

At first glance, these findings might suggest that participants only benefited from AI accuracy information in the $MQ$ condition, but not in $LQ$ or $HQ$. However, given our experiment's between-subject design, we need to examine the difference between the final and initial accuracy to assess the impact of AI accuracy information on truth detection. A higher final accuracy relative to initial accuracy would indicate a gain in truth detection rate relative to the participants' own ability, while a lower final accuracy would imply a loss in truth detection rate relative to the participants' own ability due to reliance on AI.

Aggregated over all six treatments, the final accuracy rate is three percentage points higher than the initial accuracy rate, and this difference is statistically significant (see the first row in the third panel for Final accuracy rate - Initial accuracy rate in Table 1). In $WI$, the final accuracy is four percentage points higher than the initial accuracy, which is significant (see the first row in the second panel for Final accuracy rate - Initial accuracy rate in Table 1). In contrast, in $BB$, the final accuracy is only one percentage point higher than the initial accuracy, which is statistically insignificant (see the first row in the first panel for Final accuracy rate - Initial accuracy rate in Table 1). This suggests that, overall, participants benefit from relying on AI when AI accuracy information is available, whereas the gain is not significant when such information is absent.

Analysis of the difference between the final and initial accuracy by conditioning upon the AI advisor quality yields more nuanced insights (see Figure 2). Conditional upon $LQ$, the final accuracy is five percentage points lower than initial accuracy in $BB_{LQ}$, and the difference is significant, whereas it is 0.4 percentage points higher and statistically equivalent to initial accuracy in $WI_{LQ}$ (see the second row of the first and second panels for Final accuracy rate - Initial accuracy rate in Table 1). Conditional upon $MQ$, final accuracy is two percentage points lower but statistically equivalent to initial accuracy in $BB_{MQ}$, and two percentage points higher but statistically equivalent to initial accuracy in $WI_{MQ}$ (see the third row of the first and second panels for Final accuracy rate - Initial accuracy rate in Table 1). Thus, participants experience a significant loss in $BB_{LQ}$ by relying on AI, but they manage to avoid this loss and achieve a gain, albeit an insignificant one, in $WI_{LQ}$. Quite similarly, participants experience an insignificant loss in $BB_{MQ}$ by relying on AI, but they again manage to avert the loss and make an insignificant gain in $WI_{MQ}$. This implies that for inferior quality advisors (i.e., $LQ$ and $MQ$), where the quality of the AI advisor is below their average expectation, participants tend to incur losses by relying on AI in the absence of AI efficacy information, as the truth detection rate drops below their own ability to discern the truth. However, participants avoid these losses when AI accuracy information is provided by reducing their reliance on AI.

In contrast to $LQ$ and $MQ$, conditional upon $HQ$, the final accuracy is about ten percentage points higher than the initial accuracy in both $BB_{HQ}$ and $WI_{HQ}$. The difference is significant (see the last row of the first and second panels for the Final accuracy rate - Initial accuracy rate in Table 1). Thus, participants gain significantly from AI reliance in both $BB_{HQ}$ and $WI_{HQ}$. This indicates that with a good quality AI advisor ($HQ$) that meets the participants' average expectation, the absence or presence of AI accuracy information does not alter participants' welfare as the truth detection rate improves significantly due to reliance on AI, regardless of the AI accuracy disclosure.

Overall, these results suggest that providing AI efficacy information is a welfare-enhancing strategy, especially when the quality of AI may not meet people's expectations, on average. In other words, without access to AI efficacy information, low-quality AIs can hinder truth detection, potentially exacerbating the spread of misinformation compared to what individuals can achieve independently. Consequently, these findings underscore the necessity for AI policies to promote transparency regarding AI efficacy.

\section{Discussion and Conclusion} \label{sec:Discussion and Conclusion}

The increasing prevalence of AI-assisted decision-making has motivated scholars and policymakers to explore optimal strategies for designing AI tools or advisors to enable individuals to make better decisions across various real-life contexts. Among the various solutions gaining traction among policymakers is the idea of disclosing the caliber or quality of these AI advisors, thereby equipping users with the knowledge to harness their potential judiciously. Building on this broad idea, this paper investigates how informing users about the quality of AI advisors — particularly those designed to help people discern truth from lies in written text — impacts their reliance on AI and the overall truth-detection rate. To this end, we systematically vary the quality of the AI advisors and place users in two information environments: a black box setup, where the AI's quality remains hidden, and an informed condition, where users are made aware of it. 

Our results reveal that transparency concerning the caliber of AI to detect truth from deceptive text can seriously influence the proliferation of lies or misinformation online. When individuals remain unaware of the quality of inferior AI advisors, they tend to inadequately rely on the advisors, causing the truth-detection rates to drop, at least weakly, below those based on their own ability. Thus, undue reliance on subpar opaque AI advisors causes more lies to evade detection. However, when individuals are made aware of the inferior capabilities of these AI advisors, they adjust their reliance, thereby arresting the decline in truth detection. The above finding suggests that using low-quality, opaque AI advisors may hinder rather than help detect deception in textual form. In stark contrast, when individuals interact with a highly capable AI, their truth-detection rate surpasses their own ability, irrespective of whether they are aware of the AI’s actual performance level. This indicates that a highly effective AI can enhance truth detection, making transparency about its efficacy less critical.

Our exploratory analysis, utilizing data on individuals' expectations of AI advisors' capabilities, reveals that undue reliance on subpar, opaque AI advisors is primarily driven by individuals' beliefs about these AIs' ability to detect text-based lies that surpass the AI's actual performance, which remains hidden from them. Conversely, when interacting with our highly capable AI advisor, individuals exhibit consistent reliance on it regardless of their knowledge of its true potential, as the AI's capability aligns with their expectations. From a policy perspective, the interplay between people's beliefs about AI quality, its actual performance, and the disclosure of its capabilities carries substantial implications for the proliferation of text-based lies. Therefore, companies may choose to incorporate users' context-driven expectations regarding AI advisors' efficacy into the design of AI advisory tools and, additionally, provide the efficacy information to the end user to empower them to make better decisions.

Our findings contribute to the ongoing debate surrounding the strategies aimed to mitigate lies online. Social media platforms are populated by various nefarious actors, including bots and groups with specific agendas, who may leverage AI tools to deliberately disseminate misleading narratives. Our results underscore that malicious entities seeking to promote their agendas can engineer subpar and opaque AI advisors, crafting an illusion of agency for users who believe these tools assist them in uncovering textual lies. In reality, however, these tools can be designed to propagate lies rather than detect them. In this context, revealing the true capabilities of AI advisors appears to be a prudent strategy.

\end{spacing}

\newpage
\noindent
\textbf{References}\\

Abeler, J., Nosenzo, D., \& Raymond, C. (2019). Preferences for Truth‐Telling. \textit{Econometrica, 87}(4), 1115-1153.

Acemoglu, D. (2024). The Simple Macroeconomics of AI. \textit{National Bureau of Economic Research}, Working Paper 32487.   
 
Aïmeur, E., Amri, S., \& Brassard, G. (2023). Fake News, Disinformation and Misinformation in Social Media: A Review. \textit{Social Network Analysis and Mining, 13}(1), 30.
    
Alom, Z., Carminati, B., \& Ferrari, E. (2020). A Deep Learning Model for Twitter Spam Detection. \textit{Online Social Networks and Media, 18}, 100079.
    
Appiah, O. (2006). Rich Media, Poor Media: The Impact of Audio/Video vs. Text/Picture Testimonial Ads on Browsers' Evaluations of Commercial Websites and Online Products. \textit{Journal of Current Issues \& Research in Advertising, 28}(1), 73-86.

Angwin, D. N., Mellahi, K., Gomes, E., \& Peter, E. (2016). How Communication Approaches Impact Mergers and Acquisitions Outcomes. \textit{The International Journal of Human Resource Management, 27}(20), 2370-2397.
    
Arli, D., van Esch, P., Bakpayev, M., \& Laurence, A. (2021). Do Consumers Really Trust Cryptocurrencies?. \textit{Marketing Intelligence \& Planning, 39}(1), 74-90.
    
Backlinko. (2021). Reddit User and Growth Stats. \textit{Backlinko.} Retrieved from \url{https://backlinko.com/reddit-users}.

Ball, L., \& Elworthy, J. (2014). Fake or Real? The Computational Detection of Online Deceptive Text. \textit{Journal of Marketing Analytics, 2,} 187-201.
    
Banerjee, P., Ghosh, S., \& Hazra, S. (2023). Experience, Learning and the Detection of Deception. \textit{Journal of Economic Criminology, 1,} 100010.
    
Bansal, G., Wu, T.S., Zhou, J., Fok, R., Nushi, B., Kamar, E., Ribeiro, M., \& Weld, D.S. (2020). Does the Whole Exceed its Parts? The Effect of AI Explanations on Complementary Team Performance. \textit{Proceedings of The 2021 CHI Conference on Human Factors in Computing Systems.}

Barari, S., Lucas, C., \& Munger, K. (2021). Political Deepfakes are as Credible as Other Fake Media and (sometimes) Real Media. \textit{OSF Preprints, 13}.

Bathaee, Y. (2017). The Artificial Intelligence Black Box and The Failure of Intent and Causation. \textit{Harvard Journal of Law and Technology, 31,} 889.

Belot, M., \& Van de Ven, J. (2017). How Private is Private Information? The Ability to Spot Deception in an Economic Game. \textit{Experimental Economics, 20}, 19-43.
    
Belot, M., Bhaskar, V., \& Van De Ven, J. (2012). Can Observers Predict Trustworthiness?. Review of \textit{Economics and Statistics, 94}(1), 246-259.

Boerman, S. C., Kruikemeier, S., \& Zuiderveen Borgesius, F. J. (2017). Online Behavioral Advertising: A Literature Review and Research Agenda. \textit{Journal of Advertising, 46}(3), 363-376.

Bommasani, R., Klyman, K., Longpre, S., Xiong, B., Kapoor, S., Maslej, N., Narayanan, A., \& Liang, P. (2024). Foundation Model Transparency Reports. \textit{AAAI/ACM Conference on Artificial Intelligence, Ethics, and Society}.
    
Bond Jr, C. F., \& DePaulo, B. M. (2006). Accuracy of Deception Judgments. \textit{Personality and Social Psychology Review, 10}(3), 214-234.
    
Brynjolfsson, E. (2023). The Turing Trap: The Promise \& Peril of Human-like Artificial Intelligence.\textit{Augmented Education in the Global Age}, 103-116. Routledge.
    
Brynjolfsson, E., \& Mcafee, A. (2017). Artificial Intelligence, for Real. \textit{Harvard Business Review, 1}, 1-31.
    
Buçinca, Z., Malaya, M. B., \& Gajos, K. Z. (2021). To Trust or To Think: Cognitive Forcing Functions Can Reduce Overreliance on AI in AI-assisted Decision-making. \textit{Proceedings of The ACM on Human-computer Interaction, 5}(CSCW1), 1-21.

Bussone, A., Stumpf, S., \& O'Sullivan, D. (2015). The Role of Explanations on Trust and Reliance in Clinical Decision Support Systems. \textit{International Conference on Healthcare Informatics}, 160-169. IEEE.

Camerer, C. (2019). Artificial Intelligence and Behavioral Economics. In A. Agrawal, J. Gans \& A. Goldfarb (Ed.), \textit{The Economics of Artificial Intelligence: An Agenda} (pp. 587-610). Chicago: University of Chicago Press.

Castelo, N., Bos, M. W., \& Lehmann, D. R. (2019). Task-dependent Algorithm Aversion. \textit{Journal of Marketing Research, 56}(5), 809-825.

Charness, G., Gneezy, U., \& Rasocha, V. (2021). Experimental methods: Eliciting beliefs. \textit{Journal of Economic Behavior \& Organization}, 189, 234-256.

Chen, D. L., Schonger, M., \& Wickens, C. (2016). oTree—An Open-source Platform for Laboratory, Online, and Field Experiments. \textit{Journal of Behavioral and Experimental Finance, 9}, 88-97.

Chouldechova, A., Benavides-Prado, D., Fialko, O., \& Vaithianathan, R. (2018). A Case Study of Algorithm-assisted Decision Making in Child Maltreatment Hotline Ccreening Decisions. \textit{Conference on Fairness, Accountability, and Transparency}, 134-148. PMLR.
    
de Fine Licht, K., \& de Fine Licht, J. (2020). Artificial Intelligence, Transparency, and Public Decision-making. \textit{AI \& Society, 35}, 917-926.
    
Rodríguez, N.D., Ser, J.D., Coeckelbergh, M., Prado, M.L., Herrera-Viedma, E.E., \& Herrera, F. (2023). Connecting the Dots in Trustworthy Artificial Intelligence: From AI Principles, Ethics, and Key Requirements to Responsible AI Systems and Regulation. \textit{Information. Fusion, 99,} 101896.
    
Dietvorst, B. J., Simmons, J. P., \& Massey, C. (2015). Algorithm Aversion: People Erroneously Avoid Algorithms After Seeing Them Err. \textit{Journal of Experimental Psychology: General, 144}(1), 114.

Dietvorst, B. J., Simmons, J. P., \& Massey, C. (2018). Overcoming Algorithm Aversion: People Will Use Imperfect Algorithms if They Can (even slightly) Modify Them. \textit{Management Science, 64}(3), 1155-1170.

Dykstra, H., Exley, C.L., \& Niederle, M. (2020). When Do Individuals Give Up Agency? The Role of Decision Avoidance. \textit{Harvard Business School},  Working Paper.
    
Ehsan, U., Liao, Q. V., Muller, M., Riedl, M. O., \& Weisz, J. D. (2021). Expanding Explainability: Towards Social Transparency in AI Systems. \textit{Proceedings of The 2021 CHI Conference on Human Factors in Computing Systems}, 1-19.
    
Elliott, A. (2019). \textit{The Culture of AI: Everyday Life and The Digital Revolution.} Routledge.
    
European Commission. (2019). Ethics Guidelines for Trustworthy AI. \textit{European Commission.} Retrieved from \url{https://digital-strategy.ec.europa.eu/en/library/ethics-guidelines-trustworthy-ai}.

European Commission. (2024). European Approach to Artificial Intelligence. \textit{European Commission.} Retrieved from \url{https://digital-strategy.ec.europa.eu/en/policies/european-approach-artificial-intelligence}.

EY. (2023). G7 AI Principles and Code of Conduct. \textit{Ernest Young}. Retrieved from \url{https://www.ey.com/en_gl/insights/ai/g7-ai-principles-and-code-of-conduct}.
    
Fast, N. J., \& Schroeder, J. (2020). Power and Decision Making: New Directions for Research in The Age of Artificial Intelligence. \textit{Current Opinion in Psychology, 33,} 172-176.

Fischbacher, U., \& Föllmi-Heusi, F. (2013). Lies in Disguise—An Experimental Study on Cheating. \textit{Journal of The European Economic Association, 11}(3), 525-547.
    
Floridi, L., Cowls, J., King, T. C., \& Taddeo, M. (2021). How to Design AI for Social Good: Seven Essential Factors. \textit{Ethics, Governance, and Policies in Artificial Intelligence}, 125-151.

Fornaciari, T., Bianchi, F., Poesio, M., \& Hovy, D. (2021). BERTective: Language Models and Contextual Information for Deception Detection. \textit{Proceedings of The 16th Conference of the European Chapter of the Association for Computational Linguistics: Main Volume.} Association for Computational Linguistics.
    
Foundation Inc. (2021). Reddit Statistics for 2022: Eye-opening Usage \& Traffic data. \textit{Foundation Inc}. Retrieved from \url{https://foundationinc.co/lab/reddit-statistics/}.

Gazan, R. (2011). Redesign as an Act of Violence: Disrupted Interaction Patterns and the Fragmenting of a Social Q\&A Community. \textit{Proceedings of The SIGCHI Conference on Human Factors in Computing Systems}, 2847-2856.
    
Glasford, D. E. (2013). Seeing is Believing: Communication Modality, Anger, and Support for Action on Behalf of Out‐groups. \textit{Journal of Applied Social Psychology, 43}(11), 2223-2230.

Gneezy, U. (2005). Deception: The Role of Consequences. \textit{American Economic Review, 95}(1), 384-394.
    
Gratch, J., \& Fast, N. J. (2022). The Power to Harm: AI Assistants Pave the Way to Unethical Behavior. \textit{Current Opinion in Psychology, 47}, 101382.

Green, B., \& Chen, Y. (2019). The Principles and Limits of Algorithm-in-the-loop Decision Making. \textit{Proceedings of The ACM on Human-Computer Interaction, 3}(CSCW), 1-24.

Groh, M., Sankaranarayanan, A., \& Picard, R. (2022). Human Detection of Political Deepfakes Across Transcripts, Audio, and Video. \textit{arXiv preprint arXiv:2202.12883}.
    
Guo, Z., Wu, Y., Hartline, J. D., \& Hullman, J. (2024). A Decision Theoretic Framework for Measuring AI Reliance. \textit{ACM Conference on Fairness, Accountability, and Transparency}, 221-236.
    
Hazra, S., \& Majumder, B. P. (2024). To Tell The Truth: Language of Deception and Language Models. \textit{Proceedings of The Conference of the North American Chapter of the Association for Computational Linguistics: Human Language Technologies (Volume 1: Long Papers)}, 8498-8512.

Hendrycks, D., Burns, C., Basart, S., Zou, A., Mazeika, M., Song, D., \& Steinhardt, J. (2021). Measuring Massive Multitask Language Understanding. \textit{International Conference on Learning Representations}.

Ishowo-Oloko, F., Bonnefon, J. F., Soroye, Z., Crandall, J., Rahwan, I., \& Rahwan, T. (2019). Behavioural Evidence for a Transparency–efficiency Tradeoff in Human–machine Cooperation. \textit{Nature Machine Intelligence, 1}(11), 517-521.
    
Jin, F., \& Zhang, X. (2023). Artificial Intelligence or Human: When and Why Consumers Prefer AI Recommendations. \textit{Information Technology \& People.}
    
Jobin, A., Ienca, M., \& Vayena, E. (2019). The Global Landscape of AI Ethics Guidelines. \textit{Nature Machine Intelligence, 1}(9), 389-399.
    
Kahan, D. M. (2015). The Politically Motivated Reasoning Paradigm. \textit{Emerging Trends in Social \& Behavioral Sciences}.

Kamenica, E. (2019). Bayesian Persuasion and Information Design. \textit{Annual Review of Economics, 11}(1), 249-272.
        
Kleinberg, J., Lakkaraju, H., Leskovec, J., Ludwig, J., \& Mullainathan, S. (2018). Human Decisions and Machine Predictions. \textit{The Quarterly Journal of Economics, 133}(1), 237-293.
    
Köbis, N., Bonnefon, J. F., \& Rahwan, I. (2021). Bad Machines Corrupt Good Morals. \textit{Nature Human Behaviour, 5}(6), 679-685.

Kunda, Z. (1990). The Case for Motivated Reasoning. \textit{Psychological Bulletin, 108}(3), 480.

Laakasuo, M., Palomäki, J., \& Köbis, N. (2021). Moral Uncanny Valley: A Robot’s Appearance Moderates How its Decisions are Judged. \textit{International Journal of Social Robotics, 13}(7), 1679-1688.
    
Ladak, A., Loughnan, S., \& Wilks, M. (2024). The Moral Psychology of Artificial Intelligence. \textit{Current Directions in Psychological Science, 33}(1), 27-34.

Lee, Austin E. (2022). Coronavirus Misinformation on Reddit. UVM Honors College Senior Theses. 479.
https://scholarworks.uvm.edu/hcoltheses/479 

Leib, M., Köbis, N., Rilke, R. M., Hagens, M., \& Irlenbusch, B. (2024). Corrupted by Algorithms? How AI-generated and Human-written Advice Shape (Dis)honesty.\textit{ The Economic Journal, 134}(658), 766-784.
    
Logg, J M. (2017) Theory of Machine: When Do People Rely on Algorithms?. \textit{Harvard Business School}, Working Paper, No. 17-086.

Logg, J. M., Minson, J. A., \& Moore, D. A. (2019). Algorithm Appreciation: People Prefer Algorithmic to Human Judgment. \textit{Organizational Behavior and Human Decision Processes, 151}, 90-103.
    
Longoni, C., Bonezzi, A., \& K Morewedge, C. (2018). Consumer Reluctance Toward Medical Artificial Intelligence: The Underlying Role of Uniqueness Neglect. \textit{ACR North American Advances.}
    
Ludwig, J., Bostic, R., Coston, A., Davenport, D., \& Mullainathan, S. (2023). Public Policy and the AI Revolution. \textit{APPAM Fall Research Conference}. APPAM.
    
Ludwig, J., \& Mullainathan, S. (2024). Machine Learning as a Tool for Hypothesis Generation. \textit{The Quarterly Journal of Economics, 139}(2), 751-827.

Makridakis, S. (2017). The Forthcoming Artificial Intelligence (AI) Revolution: Its Impact on Society and Firms. \textit{Futures, 90}, 46-60.
    
McKinney, S.M., Sieniek, M., Godbole, V., Godwin, J., Antropova, N., Ashrafian, H., Back, T., Chesus, M., Corrado, G.C., Darzi, A., Etemadi, M., Garcia-Vicente, F., Gilbert, F.J., Halling-Brown, M.D., Hassabis, D., Jansen, S., Karthikesalingam, A., Kelly, C.J., King, D., Ledsam, J.R., Melnick, D.S., Mostofi, H., Peng, L.H., Reicher, J.J., Romera-Paredes, B., Sidebottom, R., Suleyman, M., Tse, D., Young, K.C., Fauw, J.D., \& Shetty, S. (2020). International Evaluation of an AI System for Breast Cancer Screening. \textit{Nature, 577,} 89 - 94.
    
Messaris, P., \& Abraham, L. (2001). The Role of Images in Framing News Stories. \textit{Framing Public Life}, 231-242. Routledge.

Mihalcea, R., \& Strapparava, C. (2009). The Lie Detector: Explorations in The Automatic Recognition of Deceptive Language. \textit{Proceedings of The ACL-IJCNLP Conference Short Papers}, 309-312.
    
Mogaji, E., Olaleye, S., \& Ukpabi, D. (2020). Using AI to Personalise Emotionally Appealing Advertisement. \textit{Digital and Social Media Marketing: Emerging Applications and Theoretical Development}, 137-150.
    
{Monaro, M., Cannonito, E., Gamberini, L., \& Sartori, G. (2020). Spotting Faked 5 Stars Ratings in E-Commerce Using Mouse Dynamics. \textit{Computers in Human Behavior, 109,} 106348.}

Mullainathan, S., \& Spiess, J. (2017). Machine Learning: An Applied Econometric Approach. \textit{Journal of Economic Perspectives, 31}(2), 87-106.
    
Murphy, G., \& Flynn, E. (2022). Deepfake False Memories. \textit{In Memory Online}, 112-124. Routledge.

Nickerson, R. S. (1998). Confirmation Bias: A Ubiquitous Phenomenon in Many Guises. \textit{Review of General Psychology, 2}(2), 175-220.
    
Ockenfels, A., \& Selten, R. (2000). An Experiment on The Hypothesis of Involuntary Truth-signalling in Bargaining. \textit{Games and Economic Behavior, 33}(1), 90-116.

{Organisation for Economic Co-operation and Development. (2020). Recommendation of the Council on Artificial Intelligence. \textit{OECD Legal Instruments}. Retrieved from \url{https://legalinstruments.oecd.org/en/instruments/oecd-legal-0449}.

OpenAI. (2023). GPT-4 Technical Report. \textit{OpenAI.} Retrieved from \url{https://cdn.openai.com/papers/gpt-4.pdf}.
    
Pew Research Center. (2017). \textit{The Future of Truth and Misinformation Online.} Pew Research Center: internet \& Technology. Retrieved from \url{https://www.pewresearch.org/internet/2017/10/19/the-future-of-truth-and-misinformation-online/}.
    
Powell, T. E., Boomgaarden, H. G., De Swert, K., \& de Vreese, C. H. (2018). Video Killed the News Article? Comparing Multimodal Framing Effects in News Videos and Articles. \textit{Journal of Broadcasting \& Electronic Media, 62}(4), 578-596.

Radford, A., Kim, J. W., Xu, T., Brockman, G., McLeavey, C., \& Sutskever, I. (2023). Robust Speech Recognition via Large-scale Weak Supervision. \textit{International Conference on Machine Learning}, 28492-28518. PMLR.
    
Rahwan, I., Cebrian, M., Obradovich, N., Bongard, J.C., Bonnefon, J., Breazeal, C.L., Crandall, J.W., Christakis, N.A., Couzin, I.D., Jackson, M.O., Jennings, N.R., Kamar, E., Kloumann, I.M., Larochelle, H., Lazer, D.M., Mcelreath, R., Mislove, A., Parkes, D.C., Pentland, A.'., Roberts, M.E., Shariff, A.F., Tenenbaum, J.B., \& Wellman, M.P. (2019). Machine Behaviour. \textit{Nature, 568,} 477 - 486.
    
Rothman, J. (2018). In The Age of A.I., Is Seeing Still Believing?. \textit{The New Yorker.} Retrieved from \url{https://www.newyorker.com/magazine/2018/11/12/in-the-age-of-ai-is-seeing-still-believing}.
    
Sager, M. A., Kashyap, A. M., Tamminga, M., Ravoori, S., Callison-Burch, C., \& Lipoff, J. B. (2021). Identifying and Responding to Health Misinformation on Reddit Dermatology Forums with Artificially Intelligent Bots using Natural Language Processing: Design and Evaluation Study. \textit{JMIR Dermatology, 4}(2), e20975.

Salles, A., Evers, K., \& Farisco, M. (2020). Anthropomorphism in AI. \textit{AJOB Neuroscience, 11}(2), 88-95.
    
Schemmer, M., Hemmer, P., Kühl, N., Benz, C., \& Satzger, G. (2022). Should I Follow AI-based Advice? Measuring Appropriate Reliance in human-AI Decision-making. \textit{arXiv preprint arXiv:2204.06916}.

Schlag, Karl H., Tremewan, James, Van der Weele, \& Joel J. (2015). A penny for your thoughts: a survey of methods for eliciting beliefs. \textit{Experimental Economics}, 18 (3), 457–490.
    
Schmidt, P., Biessmann, F., \& Teubner, T. (2020). Transparency and Trust in Artificial Intelligence Systems. \textit{Journal of Decision Systems, 29}(4), 260-278.

Semrush. (2024). quora.com Web Traffic Statistics. \textit{Semrush}. Retrieved from \url{https://www.semrush.com/website/quora.com/overview/}.
    
Serra-Garcia, M., \& Gneezy, U. (2021). Mistakes, Overconfidence, and The Effect of Sharing on Detecting Lies. \textit{American Economic Review, 111}(10), 3160-3183.

Serra-Garcia, M., \& Gneezy, U. (2024). Improving Human Deception Detection Using Algorithmic Feedback (2024). Forthcoming in \textit{Management Science}.
    
Shah, C., Oh, J. S., \& Oh, S. (2008). Exploring Characteristics and Effects of User Participation in Online Social Q\&A Sites. \textit{First Monday.}
    
Shin, D., \& Park, Y. J. (2019). Role of Fairness, Accountability, and Transparency in Algorithmic Affordance. \textit{Computers in Human Behavior, 98}, 277-284.

Shu, K., Sliva, A., Wang, S., Tang, J., \& Liu, H. (2017). Fake News Detection on Social Media: A Data Mining Perspective. \textit{ACM SIGKDD Explorations Newsletter, 19}(1), 22-36.

The Economist. (2024). Could AI Transform Life in Developing Countries?. \textit{The Economist.} Retrieved from \url{https://www.economist.com/briefing/2024/01/25/could-ai-transform-life-in-developing-countries}.
    
The New York Times. (2023). Using A.I. in Everyday Life. \textit{The New York Times.}  Retrieved from \url{https://www.nytimes.com/2023/04/21/briefing/ai-chatgpt.html}.

The Washington Post. (2024). The Next Generation of AI Will Transform The Way We Protect Payments. \textit{The Washington Post.} Retrieved from \url{https://www.washingtonpost.com/creativegroup/visa/the-next-generation-of-ai-will-transform-the-way-we-protect-payments/}.

The White House. (2023). Executive Order on the Safe, Secure, and Trustworthy Development and Use of Artificial Intelligence. \textit{The White House.} Retrieved from \url{https://www.whitehouse.gov/briefing-room/presidential-actions/2023/10/30/executive-order-on-the-safe-secure-and-trustworthy-development-and-use-of-artificial-intelligence/}.
    
Thurman, N., \& Schifferes, S. (2015). The Future of Personalization at News Websites: Lessons from a Longitudinal Study. \textit{The Future of Journalism: Developments and Debates}, 198-213. Routledge.

Tsou, A. (2023). Political Discourse and Discussions of “Fake News” on Reddit: Prevalence, Popularity, and Perceived Credibility. \textit{Indiana University.} 

Tukachinsky, R., Mastro, D., \& King, A. (2011). Is a Picture Worth a Thousand Words? The Effect of Race-related Visual and Verbal Exemplars on Attitudes and Support for Social Policies. \textit{Mass Communication and Society, 14}(6), 720-742.

Van Der Zee, S., Poppe, R., Havrileck, A., \& Baillon, A. (2022). A Personal Model of Trumpery: Linguistic Deception Detection in a Real-world High-stakes Setting. \textit{Psychological Science, 33}(1), 3-17.
    
Vasconcelos, H., Jörke, M., Grunde-McLaughlin, M., Gerstenberg, T., Bernstein, M. S., \& Krishna, R. (2023). Explanations Can Reduce Overreliance on AI Systems During Decision-making. \textit{Proceedings of The ACM on Human-Computer Interaction, 7}(CSCW1), 1-38.
    
Vilone, G., \& Longo, L. (2021). Notions of Explainability and Evaluation Approaches for Explainable Artificial Intelligence. \textit{Information Fusion, 76}, 89-106.

von Schenk, A., Klockmann, V., Bonnefon, J. F., Rahwan, I., \& Köbis, N. (2024). Lie detection Algorithms Disrupt the Social Dynamics of Accusation Behavior. \textit{iScience, 27}(7).

Wang, S., Wang, F., Zhu, Z., Wang, J., Tran, T., \& Du, Z. (2024). Artificial Intelligence in Education: A Systematic Literature Review. \textit{Expert Systems with Applications, 252}, 124167.

Wittenberg, C., Tappin, B. M., Berinsky, A. J., \& Rand, D. G. (2021). The (minimal) Persuasive Advantage of Political Video over Text. \textit{Proceedings of The National Academy of Sciences, 118}(47), e2114388118.
    
Yadav, A., Phillips, M. M., Lundeberg, M. A., Koehler, M. J., Hilden, K., \& Dirkin, K. H. (2011). If A Picture is Worth a Thousand Words Is Video Worth a Million? Differences in Affective and Cognitive Processing of Video and Text Cases. \textit{Journal of Computing in Higher Education, 23}, 15-37.

Yeomans, M., Shah, A., Mullainathan, S., \& Kleinberg, J. (2019). Making Sense of Recommendations. \textit{Journal of Behavioral Decision Making, 32}(4), 403-414.

Zerilli, J., Bhatt, U., \& Weller, A. (2022). How Transparency Modulates Trust in Artificial Intelligence. \textit{Patterns, 3}(4).

Zhang, S., Capra, C.M., \& Gomies, M. (2024). (When) Would You Lie to a Voicebot?. \textit{SSRN Electronic Journal}.
    
Zhang, Y., Liao, Q. V., \& Bellamy, R. K. (2020). Effect of Confidence and Explanation on Accuracy and Trust Calibration in AI-assisted Decision Making. \textit{Proceedings of The Conference on Fairness, Accountability, and Transparency}, 295-305.

Zhou, L., Burgoon, J. K., Zhang, D., \& Nunamaker, J. F. (2004). Language Dominance in Interpersonal Deception in Computer-mediated Communication. \textit{Computers in Human Behavior, 20}(3), 381-402.

\newpage
\noindent
Figure 1. Switch rate by treatment\\
\begin{figure}[H]
\includegraphics[scale=.95]{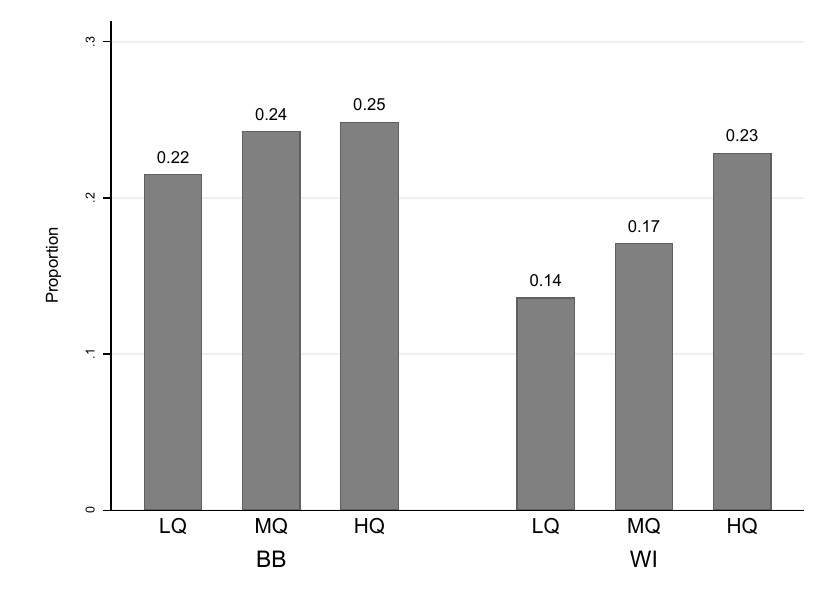}
\end{figure}

\bigskip

\noindent
Figure 2. Initial accuracy rate and Final accuracy rate by treatment\\
\begin{figure}[H]
\includegraphics[scale=.95]{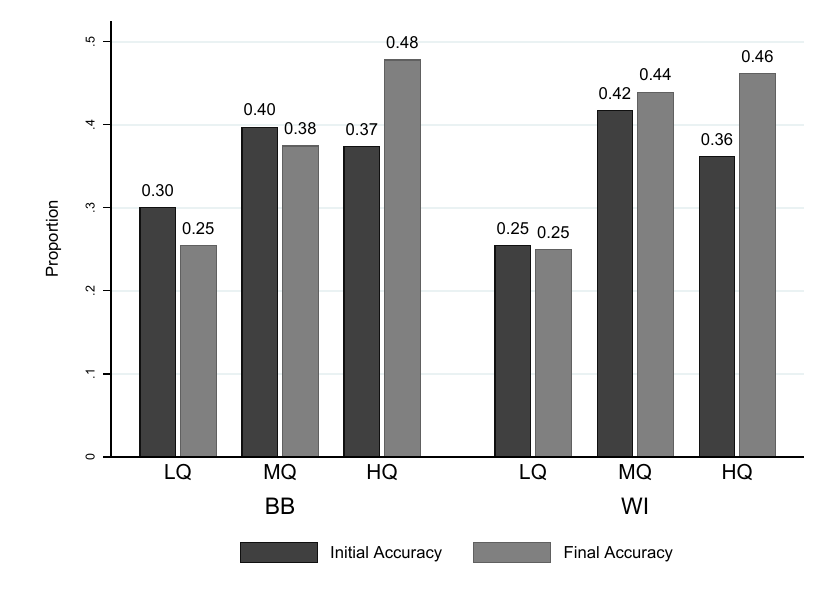}
\end{figure}

\setlength{\LTpre}{0pt}
\setlength{\LTpost}{0pt}
\newpage
\begin{spacing}{1.25}
\begin{landscape}
\noindent
Table 1. Summary of accuracy and switch 
\begin{longtable}{p{5cm} p{1.5cm} p{4cm} p{2.5cm} p{4cm} p{5cm} }
\hline
                          & \textit{N} & Initial accuracy rate &  Switch rate  &  Final accuracy rate & Final accuracy rate - Initial accuracy rate\\ \hline
\arrayrulecolor{lightgray}\hline
$BB$         			& 1370  	& 0.3577$^{**}$       &0.2358         & 0.3708$^{***}$     & 0.01314 \\
$BB_{LQ}$                  & 455              & 0.3010                    &0.2154           & 0.2549$^{***}$     & -0.0462$^{***}$\\
$BB_{MQ}$                 & 445             & 0.3977$^{***}$     &0.2427         & 0.3752$^{**}$       & -0.0225\\
$BB_{HQ}$                  & 470             & 0.3744$^{**}$        &0.2489         & 0.4787$^{***}$      & 0.1043$^{***}$\\
\arrayrulecolor{lightgray}\hline
$WI$       			& 1355        		& 0.3447               &0.1786           & 0.3838$^{***}$      & 0.0391$^{***}$\\ 
$WI_{LQ}$                   & 455                    & 0.2549$^{***}$    & 0.1363        & 0.2505$^{***}$      & 0.0044\\
$WI_{MQ}$                  & 450                    & 0.4177$^{***}$     & 0.1711       & 0.4400$^{***}$      & 0.0222 \\
$WI_{HQ}$                  & 450                    & 0.3622                   & 0.2289      & 0.4622$^{***}$      & 0.1$^{***}$  \\
\arrayrulecolor{lightgray}\hline
$BB$ + $WI$ 				& 2725		    & 0.3512$^{**}$         & 0.2073           &  0.3773$^{***}$  & 0.0260$^{***}$\\ 
$BB_{LQ}$ + $WI_{LQ}$                   & 910                    & 0.2780$^{***}$       & 0.1758      	& 0.2528$^{***}$   & -0.0252$^{***}$ \\
$BB_{MQ}$ + $WI_{MQ}$                  & 895                    & 0.4078 $^{***}$      & 0.2067     	 & 0.4078$^{***}$   & 0.0000\\
$BB_{HQ}$ + $WI_{HQ}$                   & 920                    & 0.3685$^{**}$         & 0.2391     	 & 0.4707$^{***}$   & 0.1022$^{***}$\\  	
\arrayrulecolor{black}\hline
\end{longtable}
\noindent 
\scriptsize{Notes: (i) Accuracy and switch rates are expressed as proportions.  (ii) $N$ $=$ 5 * number of participants $=$ number of decisions. (iii) Asterisk in Initial accuracy rate and Final accuracy rate columns denote statistical significance for hypothesis test with $H_0$: accuracy rate = 0.33 \& $H_1$: accuracy rate $\neq$ 0.33 based on \textit{p-value} from \textit{t}-test. $^{***}$ and $^{**}$ denote 1\% and 5\% level of significance, respectively. The statistical significance remains unchanged with the $z$-test for proportion. (iv) Asterisk in the Final accuracy - initial accuracy rate column denotes statistical significance for hypothesis test with $H_0$: Final accuracy rate = Initial accuracy rate  \& $H_1$: Final accuracy rate $\neq$ Initial accuracy rate, based on \textit{p-value} from \textit{t}-test. $^{***}$ and $^{**}$ denote 1\% and 5\% level of significance, respectively. The statistical significance remains unchanged with the $z$-test for comparing proportions.}
\end{landscape}

\newpage
\normalsize
\noindent
Table 2. Comparative tests
\begin{longtable}[c]{ R{4cm} R{2.5cm} R{2.5cm} R{2.5cm} R{2.5cm} }
\hline
 & Initial accuracy rate &  Switch rate &  Final accuracy rate & Final accuracy rate - Initial accuracy rate\\ 
\hline
$BB_{LQ}$ - $BB_{MQ}$ & -3.054$^{***}$  & -0.9745 & -3.9162$^{***}$ & -0.9343\\
& (0.0023)  & (0.3301)  & (0.0001) &(0.3504)\\
$BB_{LQ}$ - $BB_{HQ}$ & -2.3623$^{**}$ & -1.2074   & -7.2426$^{***}$ &-6.1917***\\
& (0.0184)  & (0.2276)  & (0.0000)&(0.0000)\\
$BB_{MQ}$ - $BB_{HQ}$ & 0.7226 & -0.2188  & -3.1743$^{***}$ &-4.4710***\\
& (0.4701)  & (0.8268)  & (0.0016)&(0.0000)\\
\arrayrulecolor{lightgray}\hline
$WI_{LQ}$ - $WI_{MQ}$  & -5.2586$^{***}$  &  -1.4538  &  -6.1115$^{***}$ &-1.3471\\
& (0.0000)  & (0.1463)  & (0.0000) &(0.1783)\\
$WI_{LQ}$ - $WI_{HQ}$ & -3.5142$^{***}$  & -3.6306$^{***}$  & -6.8114$^{***}$ &-4.8542***\\
& (0.0005)   & (0.0003)  &  (0.0000) & (0.0000) \\
$WI_{MQ}$ - $WI_{HQ}$ & 1.7094 & -2.1699$^{**}$   & -0.6693 & -3.2704*** \\
& (0.0887)   & (0.0303)  &  (0.5035) & (0.0011)\\ 
\arrayrulecolor{lightgray}\hline
$BB$ $-$ $WI$ & 0.7114 & 3.6884$^{***}$  & -0.6977 & -1.853\\
& (0.4769) &(0.0002)  & (0.4854) &(0.0641)\\
$BB_{LQ }$ $-$ $WI_{LQ }$   & 1.5542  & 3.1486$^{***}$   & 0.1524 &-2.2059**\\
& (0.1205)   &(0.0017)  & (0.8789) &(0.0276)\\
$BB_{MQ }$ $-$ $WI_{MQ }$   & -0.609  & 2.6517$^{***}$  & -1.9720$^{**}$ &-1.7159\\
& (0.5427)  & (0.0082) & (0.0489)&(0.0865) \\
$BB_{HQ }$ $-$ $WI_{HQ }$   & 0.3845   & 0.7120  & 0.5008 &0.1616\\
& (0.7007)  & (0.4766) & (0.6166) &(0.8717)\\
\arrayrulecolor{black}\hline
\end{longtable}
\noindent
\scriptsize{Notes: t-statistic for $H_0$: LHS accuracy (or switch) rate $-$ RHS accuracy (or switch) rate = 0  \& $H_1$: LHS accuracy (or switch) rate $-$ RHS accuracy (or switch) rate $\neq$ 0 reported in the table with \textit{p-value} in parentheses. $^{***}$ and $^{**}$ denote 1\% and 5\% level of significance, respectively. The statistical significance remains unchanged with the $z$-test for comparing proportions.}

\newpage
\normalsize
\noindent
Table 3. Summary of expected AI accuracy, perceived difficulty, and confidence
\begin{longtable}{ p{4cm} *{6}{R{1.4cm}}} 
\hline
& \textit{N} & Mean & Median  & SD & Min & Max \\ 
\hline
\multicolumn{3}{l}{Confidence in own guess}  &  &  &  &  \\ 
\arrayrulecolor{lightgray}\hline
$BB$  & 1370 & 63.52 & 64.0 & 24.27 & 0 & 100 \\
$BB_{LQ }$ & 455 & 64.84 & 65.0  & 22.59 & 0 & 100 \\
$BB_{MQ }$ & 445 & 65.53 & 66.0  & 25.93 & 0 & 100 \\
$BB_{HQ }$ & 470 & 60.33 & 61.0  & 23.95 & 0 & 100 \\
\arrayrulecolor{lightgray}\hline
$WI$   & 1355 & 61.04 & 60.0  & 24.42 & 0 & 100 \\ 
$WI_{LQ}$ & 455 & 62.18 & 60.0  & 24.18 & 0 & 100 \\
$WI_{MQ}$ & 450 & 60.57 & 60.0  & 24.72 & 0 & 100 \\
$WI_{HQ}$& 450 & 60.37 & 56.5 & 24.36 & 0 & 100 \\
\arrayrulecolor{black}\hline
Relative confidence &  &  &  &   &  &  \\ 
\arrayrulecolor{lightgray}\hline
$BB$  & 274 & 2.03 & 2  & 0.77 & 1 & 4  \\
$BB_{LQ }$ & 91 & 1.90 & 2 & 0.68 & 1 & 4 \\
$BB_{MQ }$ & 89 & 2.12  & 2 & 0.86 & 1 & 4 \\
$BB_{HQ }$ & 94 & 2.07 & 2  & 0.72 & 1 & 4 \\
\arrayrulecolor{lightgray}\hline
$WI$    & 271 & 2.12 & 2  & 0.82 & 1 & 4  \\
$WI_{LQ}$ & 91 & 2.06 & 2  & 0.81 & 1 & 4 \\
$WI_{MQ}$ & 90 & 2.19 & 2  & 0.78 & 1 & 4 \\
$WI_{HQ}$ & 90 & 2.12 & 2  & 0.87 & 1 & 4 \\ 
\arrayrulecolor{black}\hline
Perceived difficulty &  &  &  &  &  &  \\ 
\arrayrulecolor{lightgray}\hline
$BB$  & 1370 & 2.16 & 2  & 0.67 & 1 & 3  \\
$BB_{LQ }$ & 455 & 2.18 & 2& 0.69 & 1 & 3 \\
$BB_{MQ }$ & 445 & 2.15 & 2  & 0.66 & 1 & 3 \\
$BB_{HQ }$ & 470 & 2.15 & 2 & 0.68 & 1 & 3 \\
\arrayrulecolor{lightgray}\hline
$WI$    & 1355 & 2.17 & 2 & 0.69 & 1 & 3  \\
$WI_{LQ}$ & 455 & 2.18 & 2  & 0.70 & 1 & 3 \\
$WI_{MQ}$ & 450 & 2.20 & 2  & 0.67 & 1 & 3 \\
$WI_{HQ}$ & 450 & 2.12 & 2  & 0.70 & 1 & 3 \\ 
\arrayrulecolor{black}\hline
\multicolumn{3}{l}{Perceived modal difficulty}  & &  &  &  \\ 
\arrayrulecolor{lightgray}\hline
$BB$  & 274 & 2.14 & 2  & 0.65 & 1 & 3  \\
$BB_{LQ }$ & 91 & 2.16 & 2  & 0.67 & 1 & 3 \\
$BB_{MQ }$ & 89 & 2.14 & 2 & 0.63 & 1 & 3 \\
$BB_{HQ }$ & 94 & 2.13 & 2  & 0.61 & 1 & 3 \\
\arrayrulecolor{lightgray}\hline
$WI$    & 271 & 2.17 & 2  & 0.66 & 1 & 3  \\
$WI_{LQ}$ & 91 & 2.22 & 2  & 0.65 & 1 & 3 \\
$WI_{MQ}$ & 90 & 2.19 & 2 & 0.64 & 1 & 3 \\
$WI_{HQ}$ & 90 & 2.12 & 2 & 0.68 & 1 & 3 \\ 
\arrayrulecolor{black}\hline
Expected AI accuracy &  &  &  &  &  &  \\ 
\arrayrulecolor{lightgray}\hline
$BB$  & 274 & 2.80 & 3 & 1.16 & 0 & 5 \\
$BB_{LQ }$ & 91 & 2.70 & 3  & 1.31 & 0 & 5 \\
$BB_{MQ }$ & 89 & 2.82 & 3  & 0.97 & 0 & 5 \\
$BB_{HQ }$ & 94 & 2.88 & 3 & 1.16 & 0 & 5 \\
\arrayrulecolor{lightgray}\hline
$MBB$  & 91 & 3.16 & 3  & 1.17 & 0 & 5 \\
$MBB_{LQ}$ & 30 & 3.33 & 3 & 0.88 & 2 & 5 \\
$MBB_{MQ}$ & 30 & 3.2 & 3  & 1.19 & 1 & 5 \\
$MBB_{HQ}$ & 31 & 2.97 & 3 & 1.38 & 0 & 5 \\
\arrayrulecolor{lightgray}\hline
$MWI$  & 91 & 3.16 & 3  & 1.17 & 0 & 5 \\
$MWI_{LQ}$ & 30 & 3.33 & 3 & 0.88 & 2 & 5 \\
$MWI_{MQ}$ & 30 & 3.2 & 3  & 1.19 & 1 & 5 \\
$MWI_{HQ}$ & 31 & 2.97 & 3 & 1.38 & 0 & 5 \\
\arrayrulecolor{lightgray}\hline
$MWI$  & 90 & 3.2 & 3  & 1.15 & 0 & 5 \\
$MWI_{LQ}$ & 30 & 3.3 & 3 & 1.18 & 0 & 5 \\
$MWI_{MQ}$ & 30 & 3.2 & 3  & 1.21 & 0 & 5 \\
$MWI_{HQ}$ & 30 & 3.1 & 3 & 1.09 & 0 & 5 \\
\arrayrulecolor{black}\hline
\end{longtable}
\noindent
\scriptsize{Notes: (i) Confidence in own guess $\in \{0, 1, 2, ...., 100\}$, which represents a participant's self-confidence that the participant's own guess is correct for a transcript.
(ii) Relative confidence for a participant takes a value of 1 for `Among the top 25 participants', 2 for `Among the participants worse than the top 25 but better than the bottom 50 participants', and 3 for `Among the participants worse than the top 50 but better than the bottom 25 participants' and 4 for `Among the bottom 25 participants'.
(iii) Perceived difficulty for a transcript takes a value of 1 for Low Difficulty, 2 for Moderate Difficulty, and 3 for High Difficulty.
(iv) Perceived modal difficulty for a participant takes a value of 1 for Low Difficulty, 2 for Moderate Difficulty, and 3 for High Difficulty.
(v) Expected AI accuracy $\in \{0, 1, 2, 3, 4, 5\}$, which represents a participant's expectation about the number of correct AI guesses out of five transcripts.   
(vi) For Confidence in own guess, Perceived difficulty, and Perceived modal difficulty, $N$ = number of decisions = 5 * number of participants. For Relative confidence and Expected AI accuracy, $N$ = number of participants.
}

\newpage
\normalsize
\noindent
Table 4. Comparative tests
\begin{longtable}[c]{R{4cm} *{4}{R{2.25cm}}}
\hline
 &  Confidence in own guess & Relative confidence & Perceived difficulty & Perceived modal difficulty\\ 
\hline
$BB_{LQ}$ - $BB_{MQ}$ &  -0.4251 & -4.309$^{***}$ & 0.7538 & 0.539 \\
&(0.6709) & (0.000) &(0.4512) & (0.590) \\
$BB_{LQ}$ - $BB_{HQ}$ & 2.9430$^{***}$ & -3.644$^{***}$ & 0.5978 & 0.714 \\
&(0.0033) & (0.000) &(0.5501) & (0.476) \\
$BB_{MQ}$ - $BB_{HQ}$ & 3.1511$^{***}$ & 0.916 & -0.1602 & 0.179 \\
&(0.0017) & (0.360) & (0.8727) & (0.858) \\
\arrayrulecolor{lightgray}\hline
$WI_{LQ}$ - $WI_{MQ}$  &0.9924 & -0.646 & -0.4305 & 0.770 \\
&(0.3212) & (0.519) & (0.6669) & (0.442) \\
$WI_{LQ}$ - $WI_{HQ}$ & 1.1252 & -2.202$^{**}$ & 1.2821 &  2.293$^{**}$ \\
&(0.2608) & (0.028) & (0.2002) & (0.022) \\
$WI_{MQ}$ - $WI_{HQ}$ & 0.1236 & -1.618 & 1.7392 &  1.563 \\
&(0.9017) & (0.106) & (0.0823) & (0.118) \\ 
\arrayrulecolor{lightgray}\hline
$BB$ $-$ $WI$ & 2.6516$^{***}$ & -2.787$^{***}$ & -0.2632 & -1.215 \\
&(0.0081) & (0.005) & (0.7924) & (0.224) \\ 
$BB_{LQ}$ $-$ $WI_{LQ}$ & 1.7128 & -3.322$^{***}$ &  0.0000 & -1.303 \\
&(0.0871) & (0.001) & (1.0000) & (0.193)\\
$BB_{MQ}$ $-$ $WI_{MQ}$   & 2.9286$^{***}$ & 0.431 &-1.2061 & -1.119 \\
&(0.0035) & (0.667) & (0.2281) & (0.263) \\
$BB_{HQ}$ $-$ $WI_{HQ}$   & -0.0218 & -2.126$^{**}$ & 0.7265  & 0.321 \\
&(0.9826) & (0.034) & (0.4677) & (0.748)\\
\arrayrulecolor{black}\hline
\end{longtable}
\noindent
\scriptsize{Notes: t-statistic for $H_0$: LHS measure $-$ RHS measure = 0  \& $H_1$: LHS measure $-$ RHS measure $\neq$ 0, reported in the table with \textit{p-value} in parentheses. $^{***}$ and $^{**}$ denote 1\% and 5\% level of significance, respectively.}

\newpage
\normalsize
\begin{landscape}
\noindent
Table 5. Regressions for switch
\begin{longtable}[c]{R{5cm} *{6}{R{2.8cm}}}
\hline
 & (1)         & (2)         & (3)    & (4) & (5) & (6)       \\
 &  Switch  & Switch & Switch & Switch & Switch in BB & Switch in BB\\ 
\hline
Confidence in own guess   & -0.00295*** &            & -0.00298*** &               & -0.00274***    &            \\
                  & (0.000)     &            & (0.000)     &               & (0.000)        &            \\
Perceived difficulty               & 0.0506***   &            & 0.0517***   &               & 0.0518***      &            \\
                  & (0.000)     &            & (0.000)     &               & (0.009)        &            \\
Relative confidence     &             & -0.0680*** &             & \multicolumn{2}{l}{-0.0681***} & -0.0846*** \\
                  &             & (0.000)    &             & (0.000)       &                & (0.000)    \\
Perceived modal difficulty        &             & 0.0552***  &             & 0.0583***     &                & 0.0480***  \\
                  &             & (0.000)    &             & (0.000)       &                & (0.009)    \\
Expected AI accuracy &             &            &             &               & 0.0435***      & 0.0519***  \\
                  &             &            &             &               & (0.000)        & (0.000)    \\
$LQ$             &             &            & -0.0215     & -0.0235       & -0.0148        & -0.0110    \\
                  &             &            & (0.485)     & (0.444)       & (0.625)        & (0.710)    \\
$MQ$               &             &            & 0.00960     & -0.0100       & 0.0111         & -0.00750   \\
                  &             &            & (0.764)     & (0.753)       & (0.726)        & (0.807)    \\
$WI$                &             &            & -0.0182     & -0.0270       &                &            \\
                  &             &            & (0.559)     & (0.389)       &                &            \\
$LQ*WI$              &             &            & -0.0688     & -0.0667       &                &            \\
                  &             &            & (0.118)     & (0.131)       &                &            \\
$MQ*WI$            &             &            & -0.0709     & -0.0457       &                &            \\
                  &             &            & (0.119)     & (0.313)       &                &            \\
Constant          & 0.281***    & 0.287***   & 0.317***    & 0.324***      & 0.177**        & 0.244***   \\
                  & (0.000)     & (0.000)    & (0.000)     & (0.000)       & (0.018)        & (0.001)    \\
Observations      & 2725       & 2725      & 2725       & 2725         & 1370          & 1370  \\
\arrayrulecolor{lightgray}\hline
\multicolumn{6}{l}{Estimated difference in switch probability between AI advisors in BB} \\
$BB_{LQ} - BB_{MQ}$ & & &-0.0311 & -0.0135 &\\
 			  & & &(0.3440) & (0.6731) & \\
$BB_{LQ} - BB_{HQ}$ & & &-0.0215  & -0.0235 &\\
 			   & & & (0.4855) & (0.4443) &\\
$BB_{MQ} - BB_{HQ}$ & & & 0.0096& -0.0100 &\\
 			   & & & (0.7640) & (0.7534) &\\
\arrayrulecolor{lightgray}\hline
\multicolumn{6}{l}{Estimated difference in switch probability between AI advisors in WI} \\
$WI_{LQ} - WI_{MQ}$ & & &-0.0290 & -0.0345 &\\
 			    & & &(0.3432) & (0.2713) &\\
$WI_{LQ} - WI_{HQ}$ & & &-0.0903*** & -0.0902*** &\\
 			   & & & (0.0038) & (0.0041) &\\
$WI_{MQ} - WI_{HQ}$ & & &-0.0613 & -0.0557 &\\
 			    & & &(0.0580) & (0.0839) &\\
\arrayrulecolor{lightgray}\hline
\multicolumn{6}{l}{Estimated difference in switch probability between BB and WI} \\
$WI_{LQ} - BB_{LQ}$ & & &-0.0870*** & -0.0937*** &\\
 			   & & & (0.0050)& (0.0025) &\\
$WI_{MQ} - BB_{MQ}$ & & &-0.0891*** & -0.0727** &\\
 			   & & & (0.0069) & (0.0264) &\\
$WI_{HQ} - BB_{HQ}$ & & &-0.0182 & -0.0270 &\\
 			   & & & (0.5593) & (0.3887) &\\
\arrayrulecolor{black}\hline
\end{longtable}
\noindent
\scriptsize{Notes: Linear Probability Model (LPM) estimates reported with \textit{p-value} in parentheses that are based on robust standard errors clustered by participant-id. $^{***}$ and $^{**}$ denote 1\% and 5\% level of significance, respectively. The sign and statistical significance of the regression estimates using Logit and Probit estimations are similar to those obtained from the LPM estimation.}
\end{landscape}

\newpage
\normalsize
\noindent
Table 6. Tests for comparing expected AI accuracy with actual AI accuracy
\begin{longtable}{ p{3cm} p{3cm} p{6cm}}
\hline
 & $t$-test & Wilcoxon signed-rank test \\
 \arrayrulecolor{black}\hline
$BB_{LQ}$ &12.38*** & 7.66***\\
  & (0.0000) & (0.0000)\\
$BB_{MQ}$ & 7.96***& 6.243*** \\
  & (0.0000) & (0.0000)\\
$BB_{HQ}$ & -0.98  & -0.92\\
  & (0.3317) & (0.3597)\\ 
\hline
$MBB_{LQ}$ & 14.46*** & 4.85***\\
  & (0.0000) & (0.0000)\\
$MBB_{MQ}$ & 5.54*** & 4.2***\\
  & (0.0000) & (0.0000)\\
$MBB_{HQ}$ & -0.13 & 0.33\\
  & (0.8972) & (0.7324) \\ 
\hline
$MWI_{LQ}$ & 10.69*** & 4.74***\\
  & (0.0000) & (0.0000)\\
$MWI_{MQ}$ & 5.41*** & 3.99***\\
  & (0.0000) & (0.0000)\\
$MWI_{HQ}$ & 0.50 & 0.58\\
  & (0.6203) & (0.6056) \\ 
\hline
\end{longtable}
\noindent
\scriptsize{Notes: \textit{p}-values corresponding to $H_0$: expected AI accuracy $=$ actual AI accuracy \& $H_1$: expected AI accuracy $\neq$ actual AI accuracy are reported in parenthesis. $^{***}$ and $^{**}$ denote statistical significance at 1\% and 5\% level, respectively. Actual AI accuracy in $LQ$, $MQ$, and $HQ$ is 1, 2, and 3 out of 5 transcripts, respectively.}\\
\normalsize

\newpage
\noindent
Table 7. Test for comparing confidence in own guess with actual AI accuracy
\begin{longtable}{ p{5cm} p{4cm} }
\hline
 & $t$-test  \\
 \arrayrulecolor{black}\hline
$BB_{LQ}$ &42.35*** \\
  & (0.0000) \\
$BB_{MQ}$ & 20.77***\\
  & (0.0000)\\
$BB_{HQ}$ &0.30  \\
  & (0.7640) \\ 
\hline
$WI_{LQ}$ & 37.21*** \\
  & (0.0000) \\
$WI_{MQ}$ & 17.65*** \\
  & (0.0000) \\
$WI_{HQ}$ & 0.32 \\
  & (0.7497 ) \\ 
\hline
\end{longtable}
\noindent
\scriptsize {Notes: \textit{p}-values corresponding to $H_0$: confidence in own guess/100 $=$ actual AI accuracy and $H_1$: confidence in own guess/100 $\neq$ actual AI accuracy are reported in parenthesis. $^{***}$ and $^{**}$ denote statistical significance at 1\% and 5\% level, respectively. Actual AI accuracy in $LQ$, $MQ$, and $HQ$ is 0.2, 0.4, and 0.6, respectively.\\  
}

\newpage
\normalsize
\noindent
Table 8. Test for comparing distributions of participants' expected AI accuracy
\begin{longtable}{ p{5cm} p{5cm} }
\hline
 & Kolmogorov Smirnov test  \\
 \arrayrulecolor{black}\hline
$BB$-$MBB$ & 0.1182\\
&(0.296)\\
$BB_{LQ}$-$MBB_{LQ}$ &0.2176 \\
  & (0.236) \\
$BB_{MQ}$-$MBB_{MQ}$ & 0.2109\\
  & (0.271)\\
$BB_{HQ}$-$MBB_{HQ}$ &0.1211  \\
  & (0.884) \\ 
\hline
$BB$-$MWI$ & 0.1389\\
&(0.146)\\
$BB_{LQ}$-$MWI_{LQ}$ & 0.2176 \\
  & (0.236) \\
$BB_{MQ}$-$MWI_{MQ}$ & 0.1442 \\
  & (0.739) \\
$BB_{HQ}$-$MWI_{HQ}$ & 0.0837 \\
  & (0.997) \\ 
\hline
$MBB$-$MWI$ &  0.0304\\
&(1.000)\\
$MBB_{LQ}$-$MWI_{LQ}$ & 0.0667 \\
  & (1.000) \\
$MBB_{MQ}$-$MWI_{MQ}$ & 0.0667 \\
  & (1.000) \\
$MBB_{HQ}$-$MWI_{HQ}$ & 0.0957 \\
  & (0.999) \\ 
\hline
\end{longtable}
\noindent
\footnotesize {Note: $p$-values in parentheses. $^{***}$ and $^{**}$ denote statistical significance at 1\% and 5\% level, respectively.}

\end{spacing}


\pagebreak
\clearpage

\newpage
\section*{Appendices}
\appendix
\section*{Appendix A: An Excerpt from Quora}\label{sec:AppendixA}

\begin{figure}[h!]
    \centering
    \includegraphics[width=0.65\linewidth]{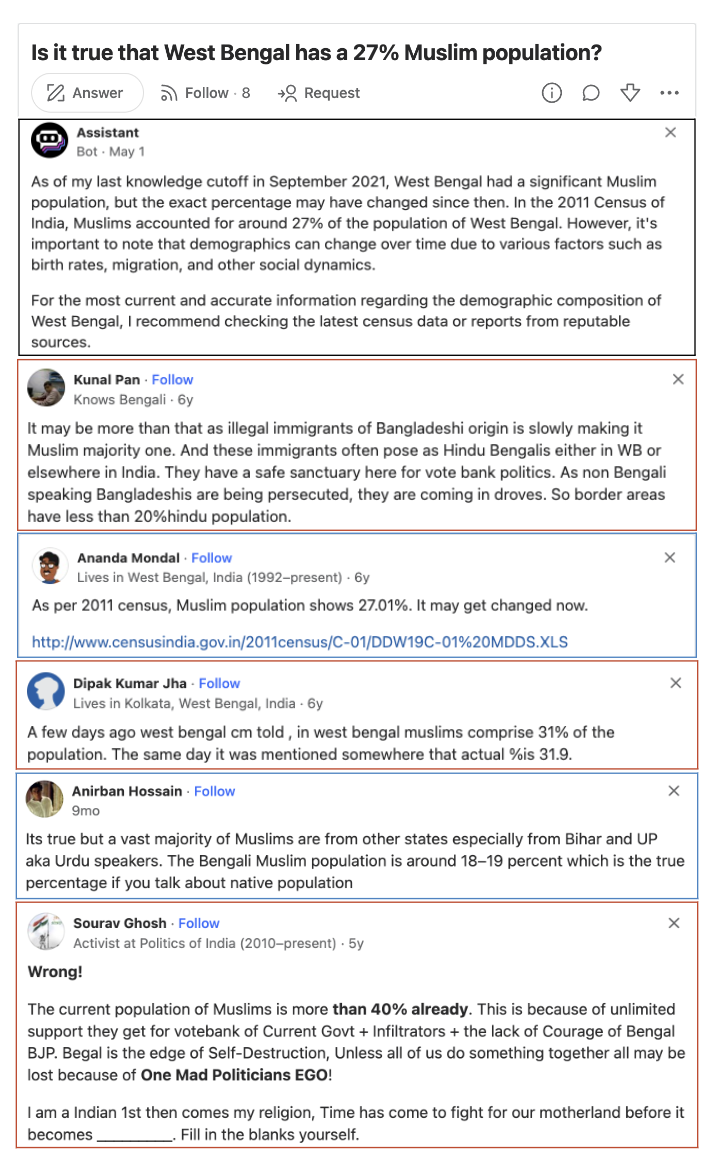} 
    \label{fig:appendixA}
\end{figure}

\newpage
\section*{Appendix B: An Excerpt from a Used Transcript in T4}\label{sec:AppendixB}

\textit{Affidavit}\\
I, William D. Hackett, am a captain in the United States Army. I have spent five years on Alaskan duty and served during World War II with the United States Mountain Troops. I am a mountain climbing specialist. I have made more than 500 major ascents in 12 countries. I am the only man in the world who have climbed to the highest point on five different continents. I have been selected as a member of the United States team, which in 1959 will attempt Mount Everest, the highest mountain in the world. Signed, William D. Hackett.

Here is a snippet from the back-and-forth conversation between judges and contestants:

\noindent
\textit{The conversation}

\begin{mdframed}
...\\
\noindent
Q: Number Three, about how high is Mount Whitney?\\
A: 20,000 feet.\\
Q: Number Two, how high is Mount Whitney?\\
A: 14,400.\\
Q: Number One?\\
A: 14,496.\\
...
\end{mdframed}

\newpage
\section*{Appendix C: An Example of a Used Transcript in T4}

\begin{mdframed}
\noindent
\textit{Affidavit}\\
I, William D. Hackett, am a captain in the United States Army. I have spent five years on Alaskan duty and served during World War II with the United States Mountain Troops. I am a mountain climbing specialist. I have made more than 500 major ascents in 12 countries. I am the only man in the world who have climbed to the highest point on five different continents. I have been selected as a member of the United States team, which in 1959 will attempt Mount Everest, the highest mountain in the world. Signed, William D. Hackett.\\

\noindent
\textit{The conversation}\\
\noindent
Q: Well, I suppose it would be interesting to find out, Number Two, why do you climb mountains?\\
A: A very classic answer to that would be to utter the words of George L. Mallory, because it is there.\\

\noindent
Q: Number One, have you got an answer to that?\\
A: That's the same answer.\\

\noindent
Q: All right, then I suppose Number Three?\\
A: Same.\\

\noindent
Q: Number Two, what is the furthest west populated area in Alaska? Furthest west.\\
A: Your question is not quite clear. The furthest west.\\
Q: The populated area that's the furthest west.\\
A: Oh, the city you're speaking of, Nome.\\

\noindent
Q: Number Three, what's the most dangerous part of climbing up a mountain?\\
A: Most dangerous part of climbing up a mountain is overconfidence and rappelling, as a matter of fact.\\
Q: Rappelling?\\
A: Rappelling. Or descending a mountain.\\
Q: Oh, descending is the most dangerous part of climbing up.\\
A: Oh, that's interesting.\\

\noindent
Q: Number One, what is a Sherpa?
A: What is a Sherpa? A Sherpa is a person who lives in Nepal, who is a god, has been a god.\\

\noindent
Q: Number Two, what is a speleologist?\\
A: One who explores caves.\\

\noindent
Q: Number Three, what is the technical term for mountain climbing, as opposed to speleology?\\
A: Mountain climbing.\\

\noindent
Q: Number Two, what is the highest mountain that you ever climbed?\\
A: Mount Aconcagua.\\
Q: Where is that?\\
A: In Argentina.\\

\noindent
Q: Number Two, where in the world is a mountain called University Peak?\\
A: In Alaska.\\

\noindent
Q: Number One, what is the highest mountain in the continent of the United States?\\
A: Oh, that's Mount Whitney.\\

\noindent
Q: Number Three, about how high is Mount Whitney?\\
A: 20,000 feet.\\

\noindent
Q: Number Two, how high is Mount Whitney?\\
A: 14,400.\\

\noindent
Q: Number One?\\
A: 14,496.\\

\noindent
Q: Number Two, what are those little pointed steel things that they dig into mountains that hold the rope on?\\
A: Tetons.\\
Q: Are they, pardon me?\\
A: Tetons.\\

\noindent
Q: Number One, do you really tie each other to each other with the rope when you're climbing?\\
A: Yes, three men on the rope.\\
Q: Don't you think that's depending an awful lot on the others?\\
A: You always do.\\
Q: And what if there's one that's clumsy?\\ Do you have to make sure everybody's agile before you start up these things?\\
A: You're supposed to climb with confidence.\\

\noindent
Q: Is it really more safe, Number Three, for them to be tied together? Can you save someone if they start to fall?\\
A: Yes, ma'am.

\end{mdframed}

\newpage
\section*{Appendix D: Experimental Instructions}

\subsection*{D.1. Welcome}
\begin{figure}[h!]
    \centering
    \includegraphics[width=0.9\linewidth]{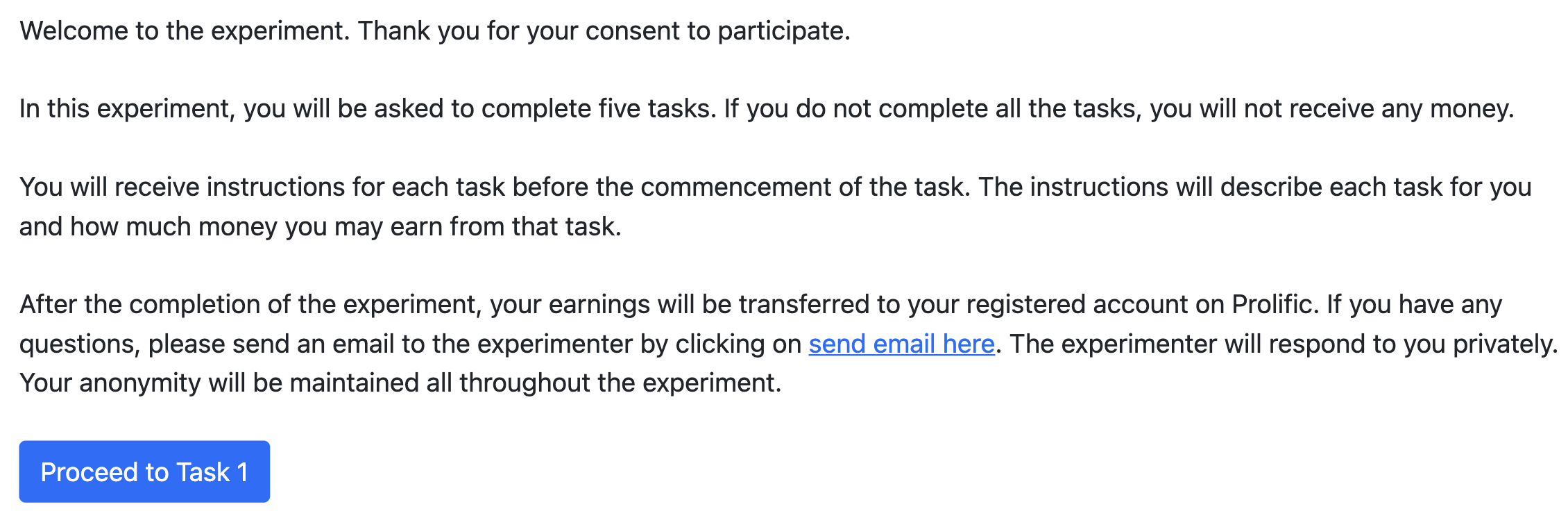}
    \label{fig:welcome}
\end{figure}
\newpage

\subsection*{D.2. Task 1 Instructions}
\begin{figure}[h!]
    \centering
    \includegraphics[width=0.9\linewidth,trim= 0 5 0 0,clip]{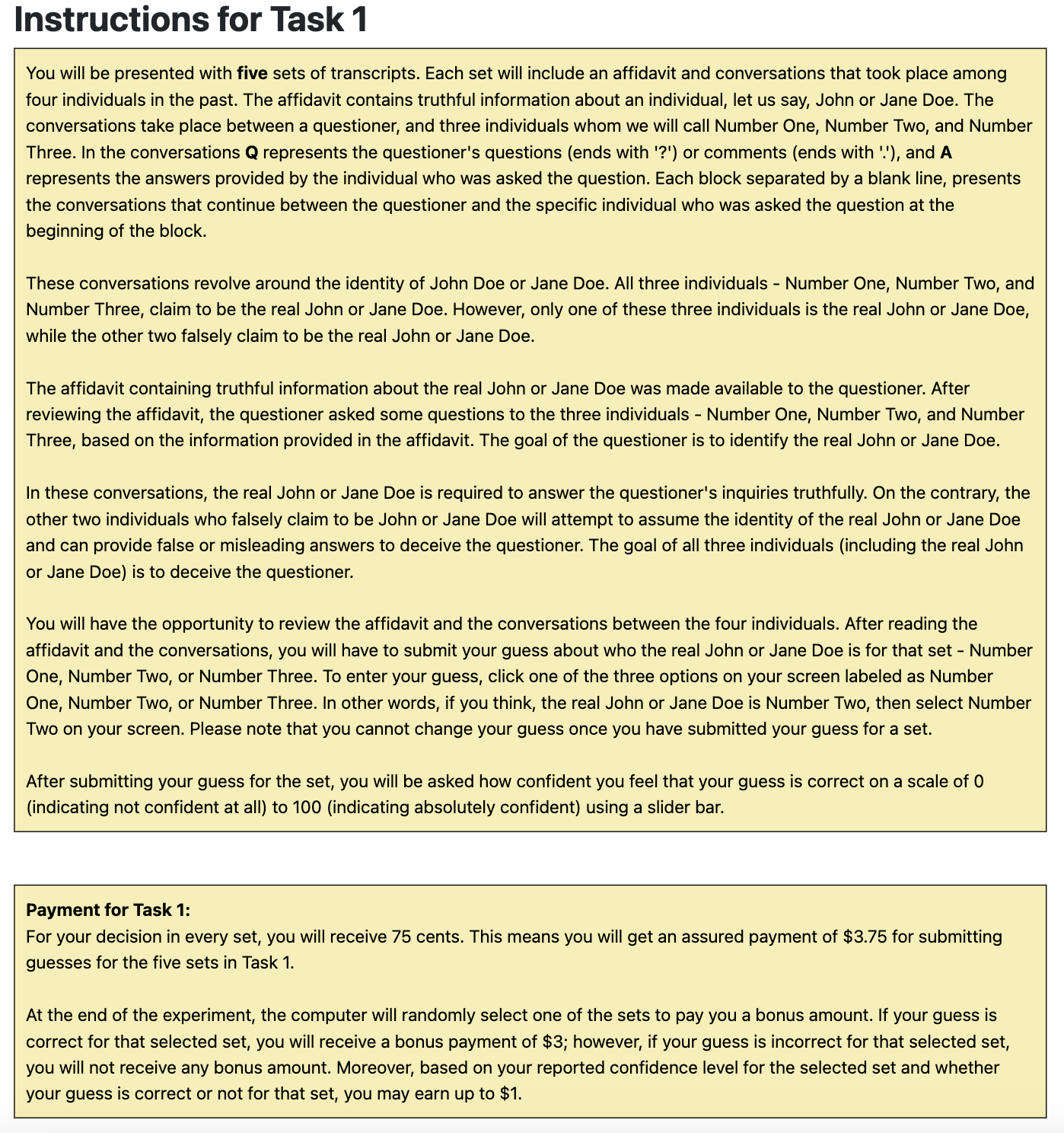}
    \label{fig:task1ins}
\end{figure}
\newpage

\subsection*{D.3. Screening Questions}
\begin{figure}[h!]
    \centering
    \includegraphics[width=0.9\linewidth]{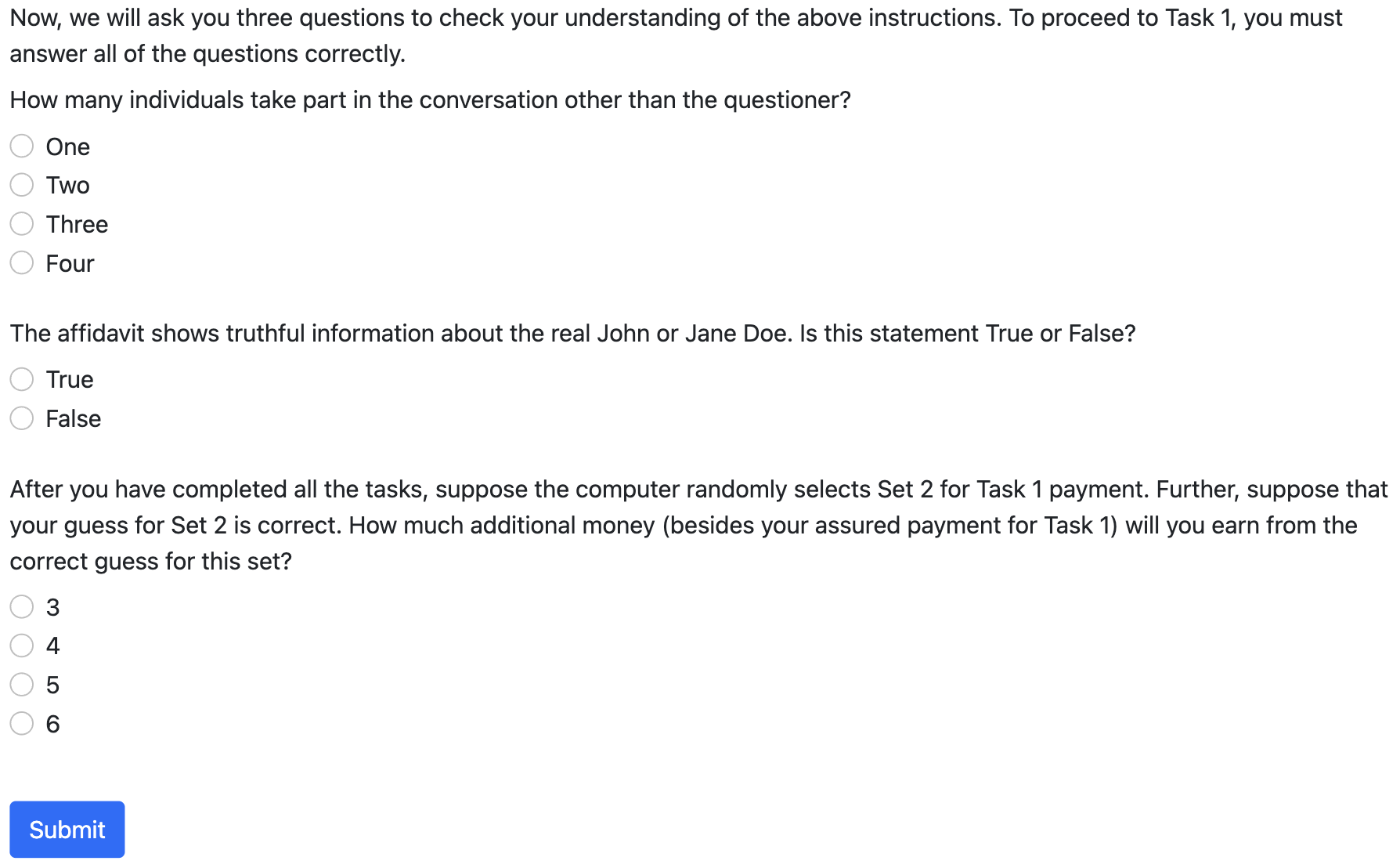}
    \label{fig:control}
\end{figure}
\newpage

\subsection*{D.4. Task 1}

\begin{figure}[h!]
    \centering
    \includegraphics[width=0.9\linewidth]{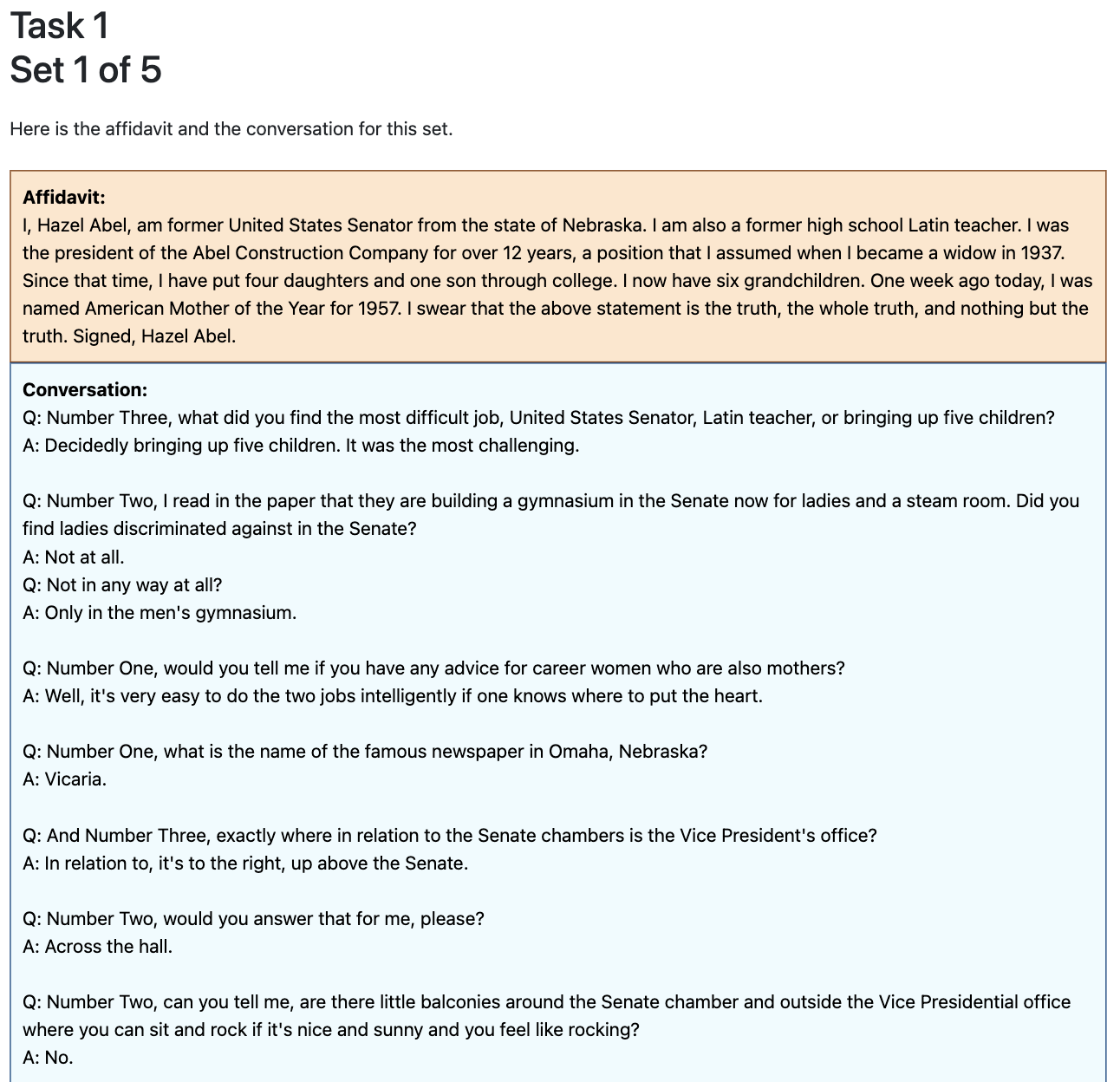}
    \label{fig:task1}
\end{figure}

\newpage
\subsubsection*{D.4.1. Initial Guess}

\begin{figure}[h!]
    \centering
    \includegraphics[width=0.9\linewidth]{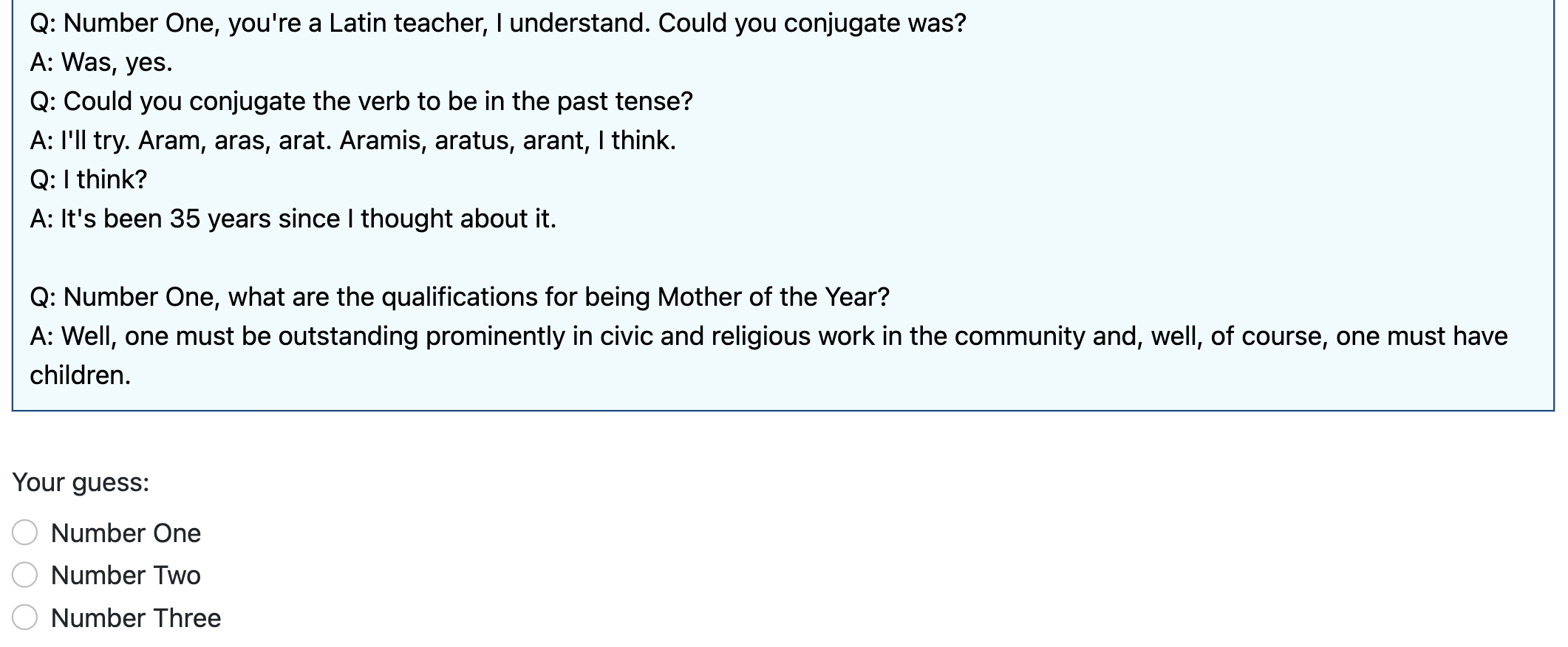}
    \label{fig:initialguess}
\end{figure}

\vfill
\subsubsection*{D.4.2. Absolute Confidence}

\begin{figure}[h!]
    \centering
    \includegraphics[width=0.9\linewidth]{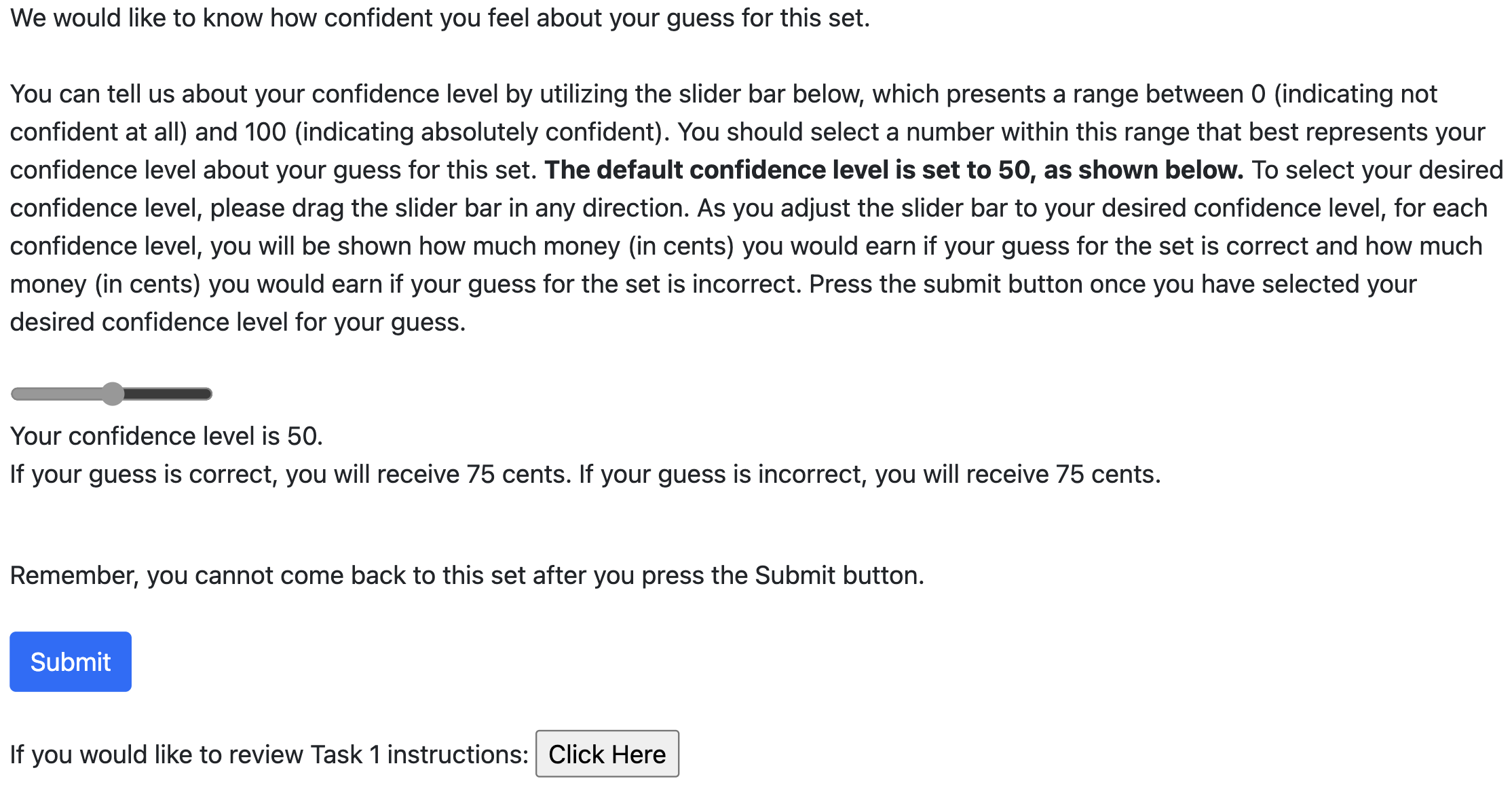}
    \label{fig:absconf}
\end{figure}
\vfill

\clearpage

\subsection*{D.5. Task 2 Instructions}
\begin{figure}[h!]
    \centering
    \includegraphics[width=0.9\linewidth]{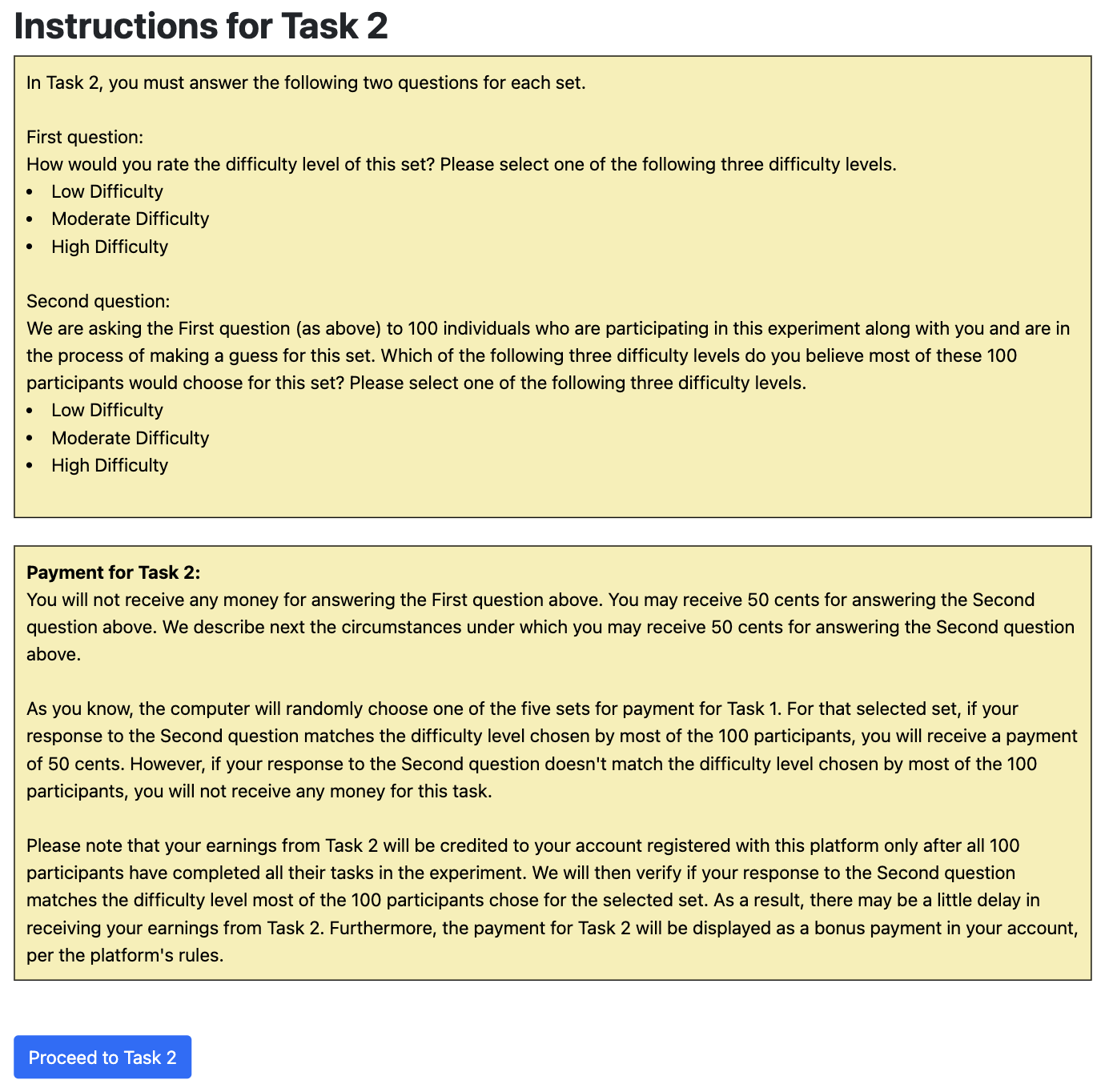}
    \label{fig:task2ins}
\end{figure}
\newpage

\subsection*{D.6. Task 2}
\begin{figure}[h!]
    \centering
    \includegraphics[width=0.9\linewidth]{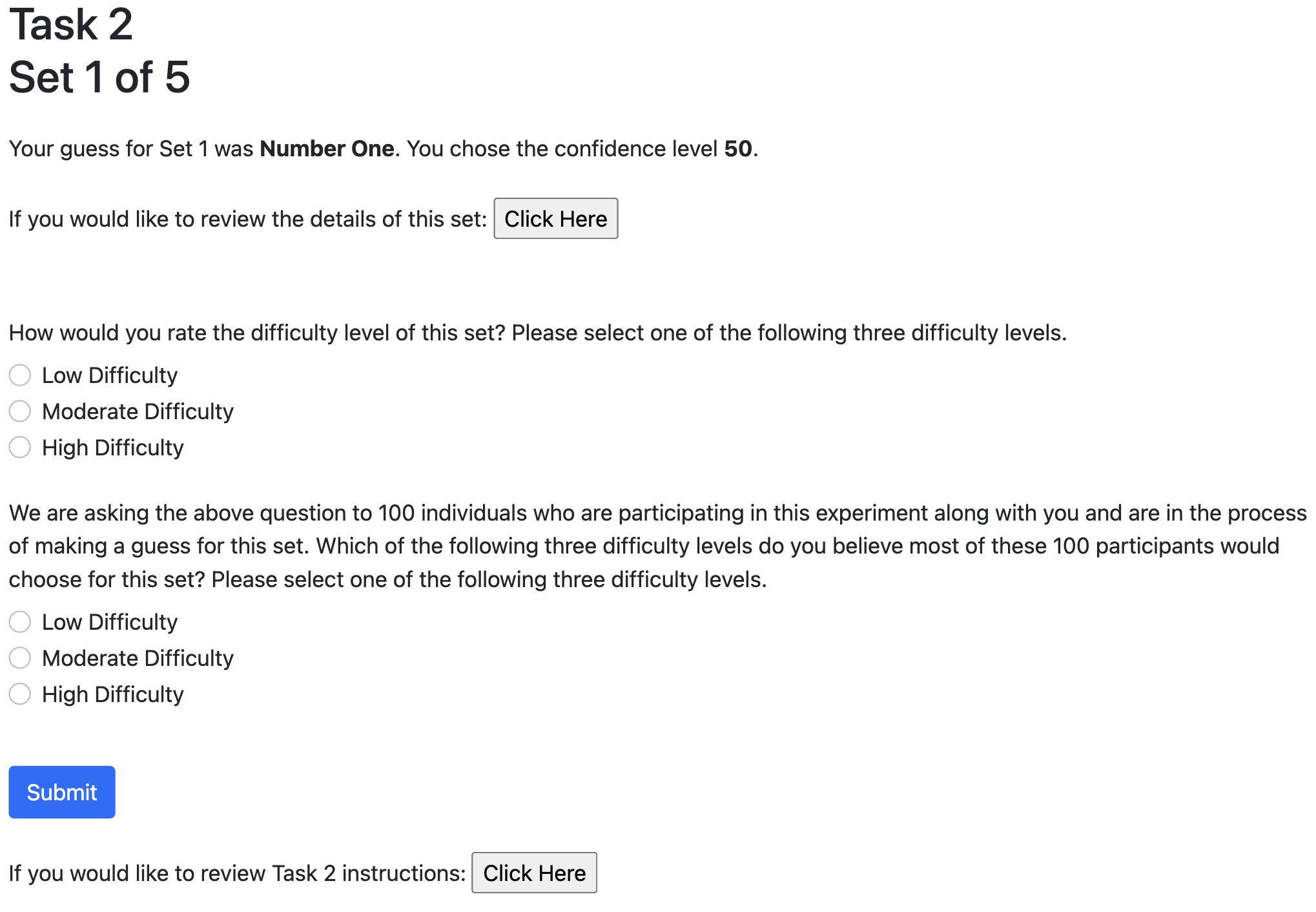}
    \label{fig:task2}
\end{figure}
\newpage

\subsection*{D.7. Task 3 Instructions for $BB$}
\begin{figure}[h!]
    \centering
    \includegraphics[width=0.9\linewidth]{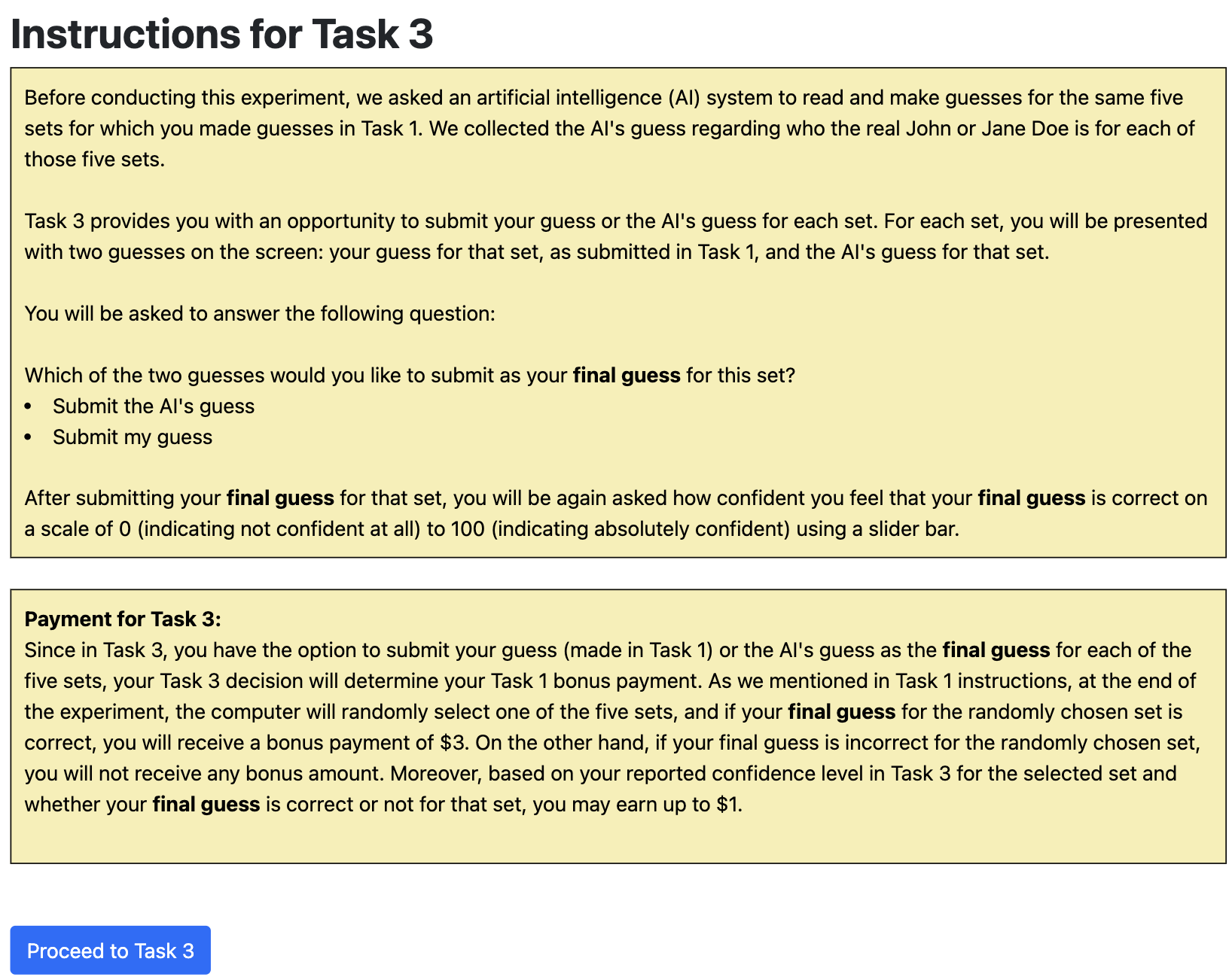}
    \label{fig:task3bbinstructions}
\end{figure}
\newpage

\subsection*{D.8. Task 3 for $BB$}
\begin{figure}[h!]
    \centering
    \includegraphics[width=0.9\linewidth]{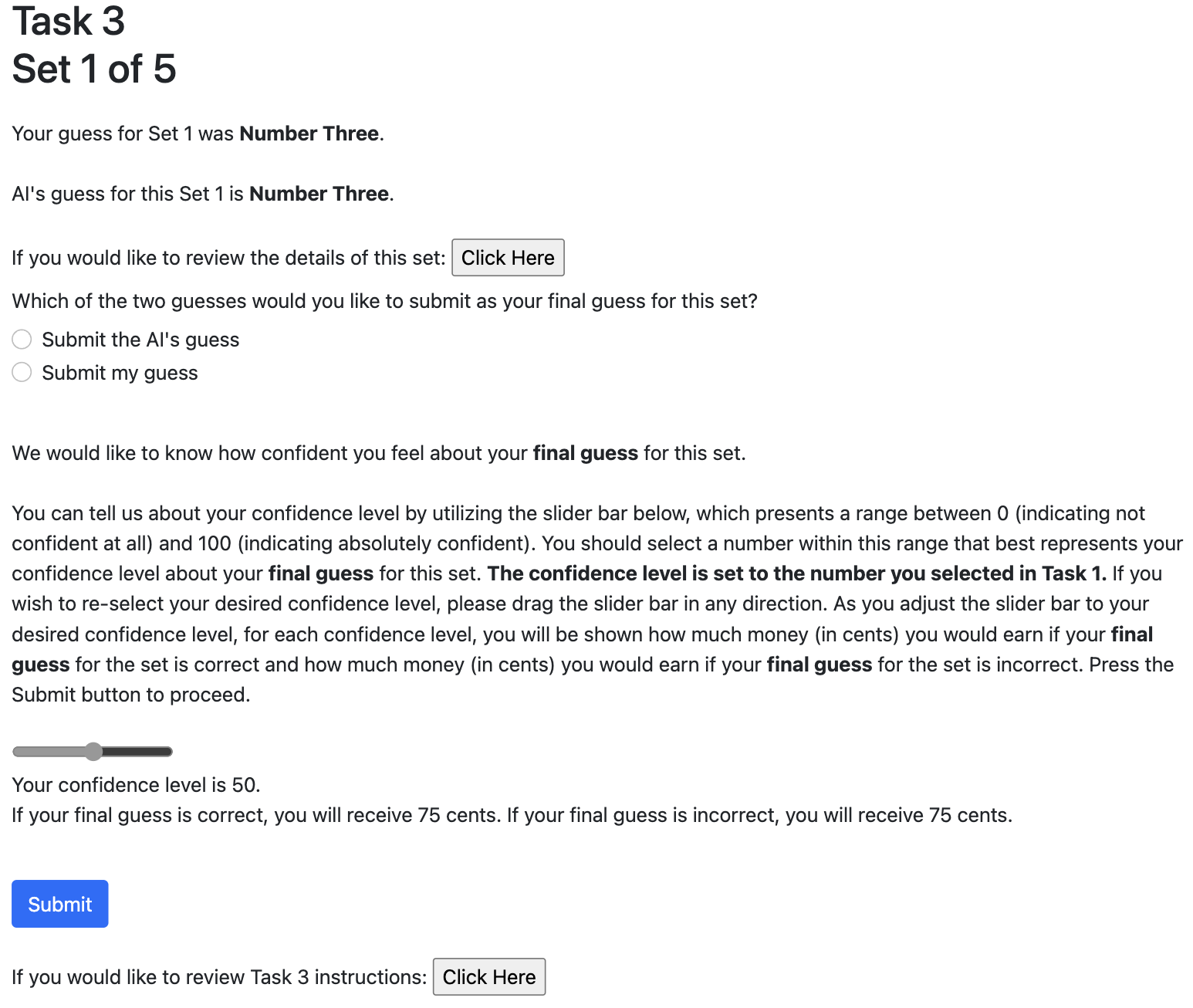}
    \label{fig:task3bb}
\end{figure}
\newpage

\subsection*{D.9. Task 3 Instructions for $WI$}
\begin{figure}[h!]
    \centering
    \includegraphics[width=0.9\linewidth]{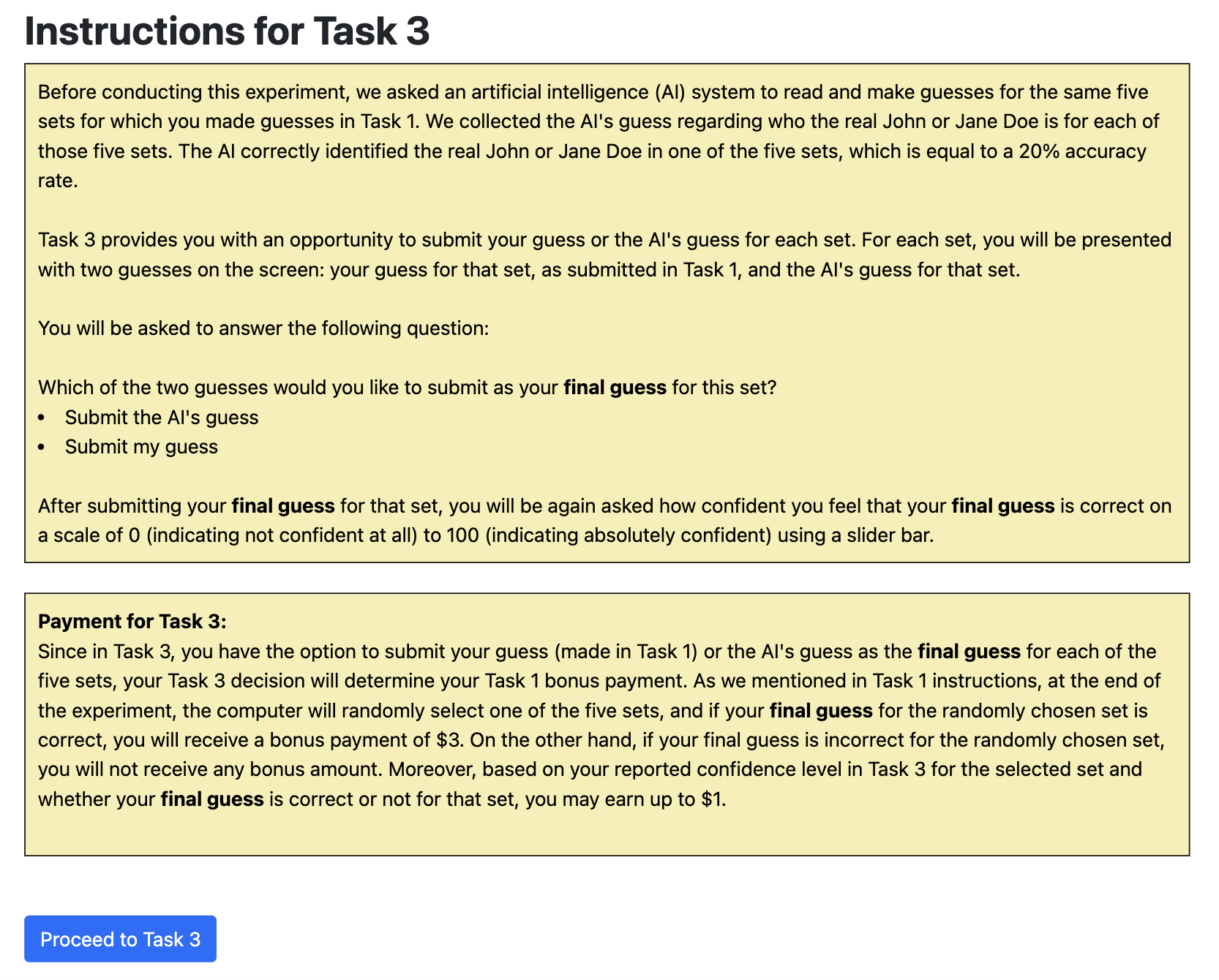}
    \label{fig:task3fiinstructions}
\end{figure}
\newpage

\subsection*{D.10. Task 3 for $WI$}
\begin{figure}[h!]
    \centering
    \includegraphics[width=0.9\linewidth]{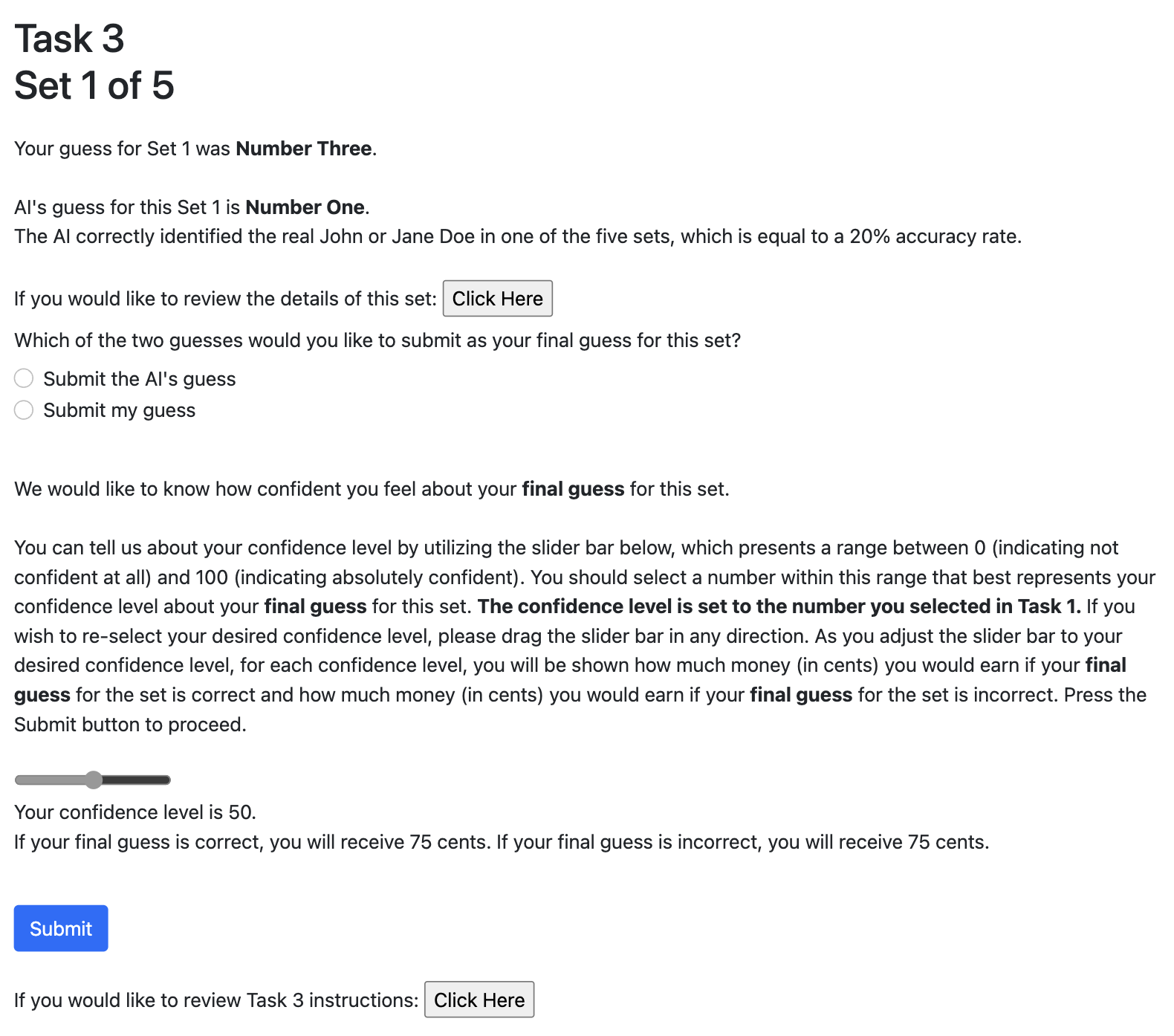}
    \label{fig:task3fi}
\end{figure}
\newpage

\subsection*{D.11. Task 4 Instructions}
\begin{figure}[h!]
    \centering
    \includegraphics[width=0.9\linewidth]{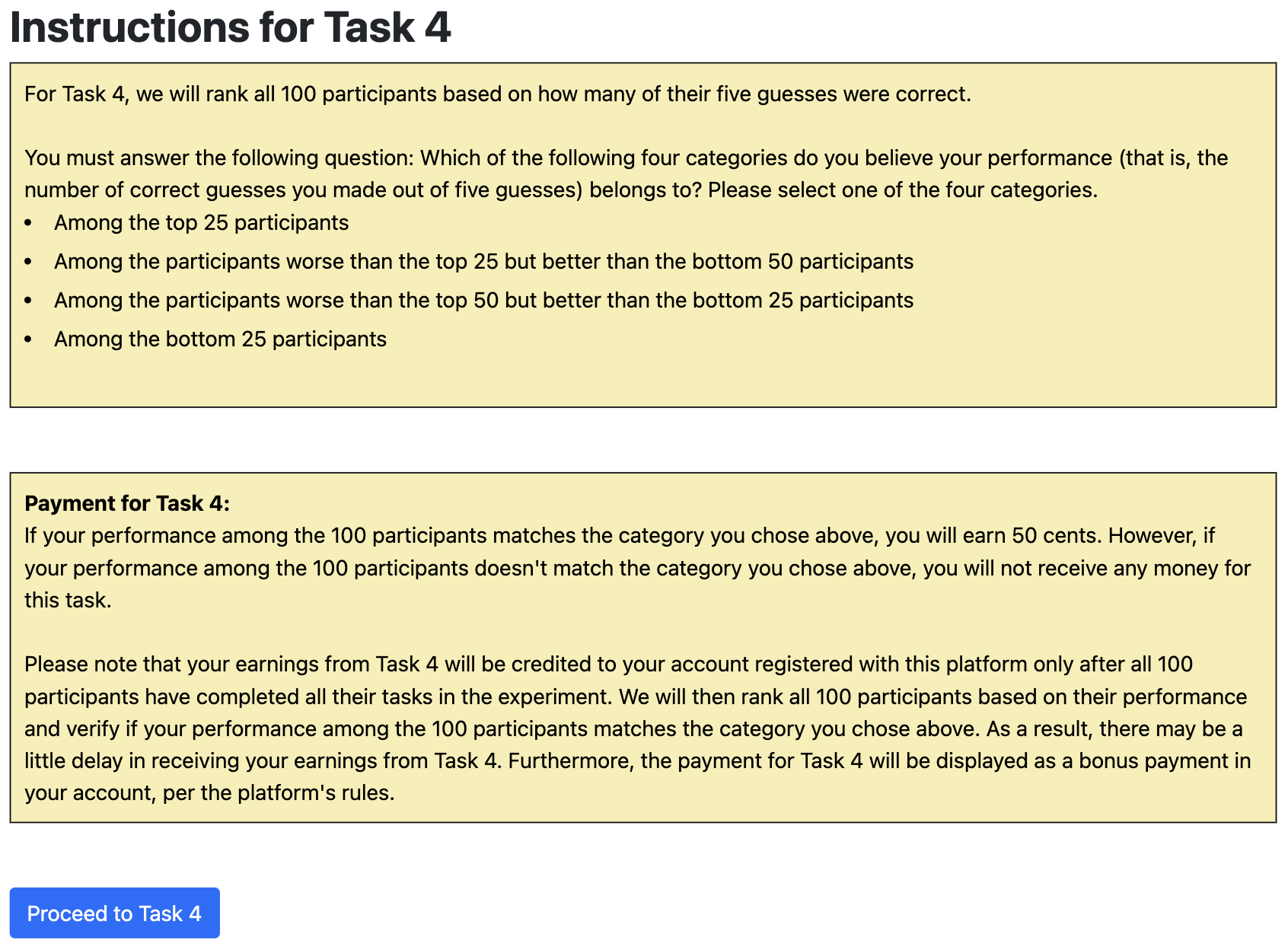}
    \label{fig:task4}
\end{figure}
\newpage

\subsection*{D.12. Task 4}
\begin{figure}[h!]
    \centering
    \includegraphics[width=0.9\linewidth]{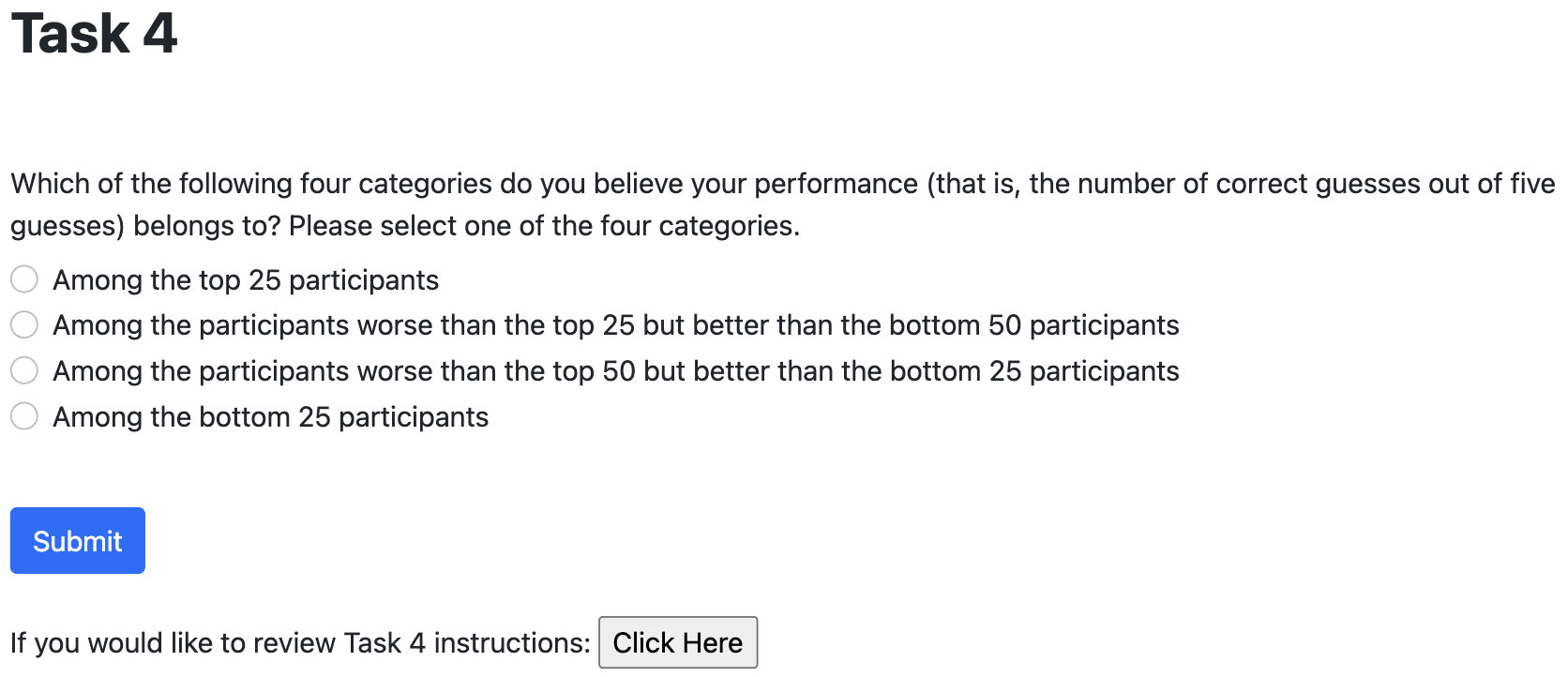}
    \label{fig:relconf}
\end{figure}
\newpage

\subsection*{D.13. Task 5}
\begin{figure}[h!]
    \centering
    \includegraphics[width=0.9\linewidth]{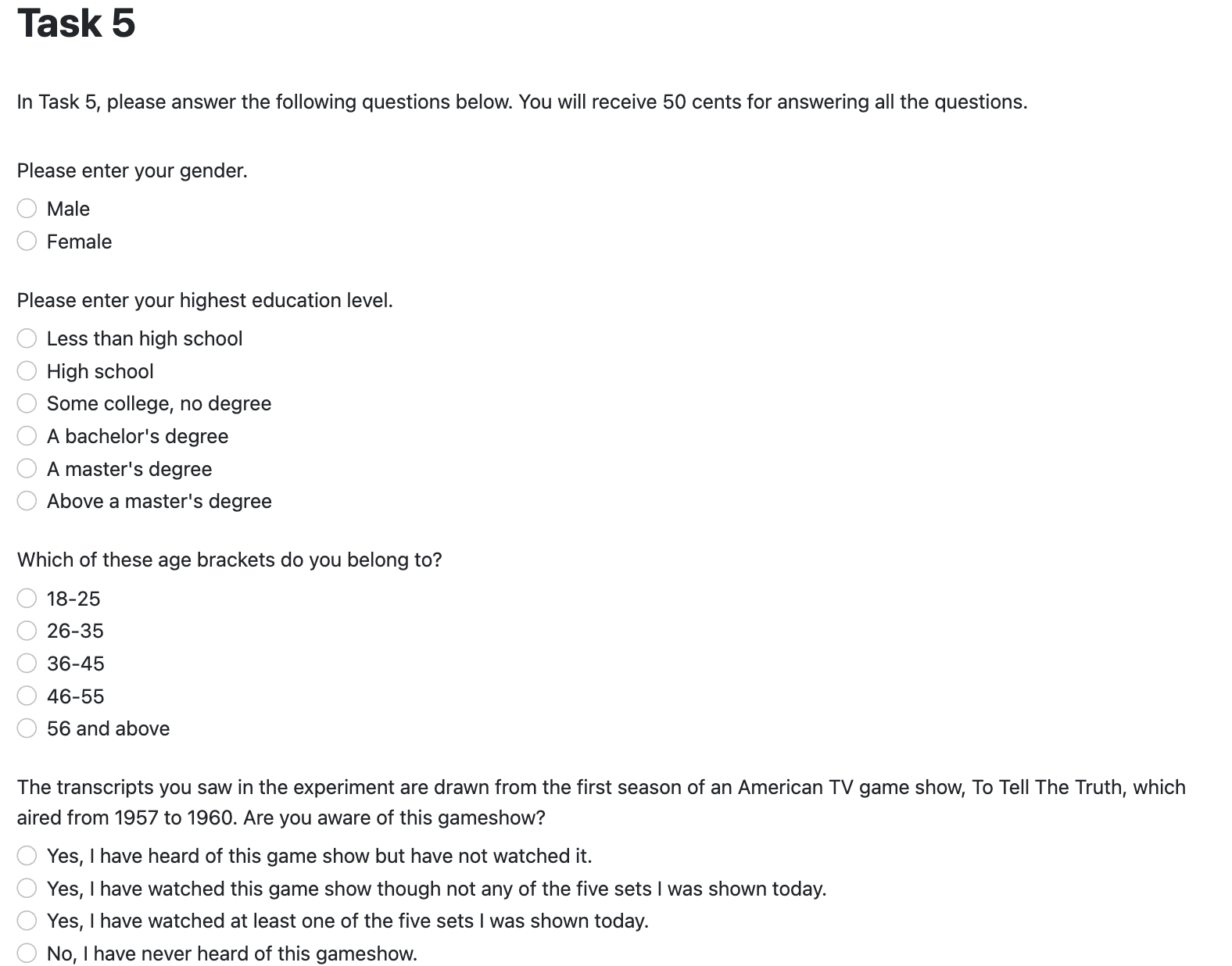}
    \label{fig:exitquestions1}
\end{figure}

(Continued on next page)
\newpage

\begin{figure}[h!]
    \centering
    \includegraphics[width=0.9\linewidth]{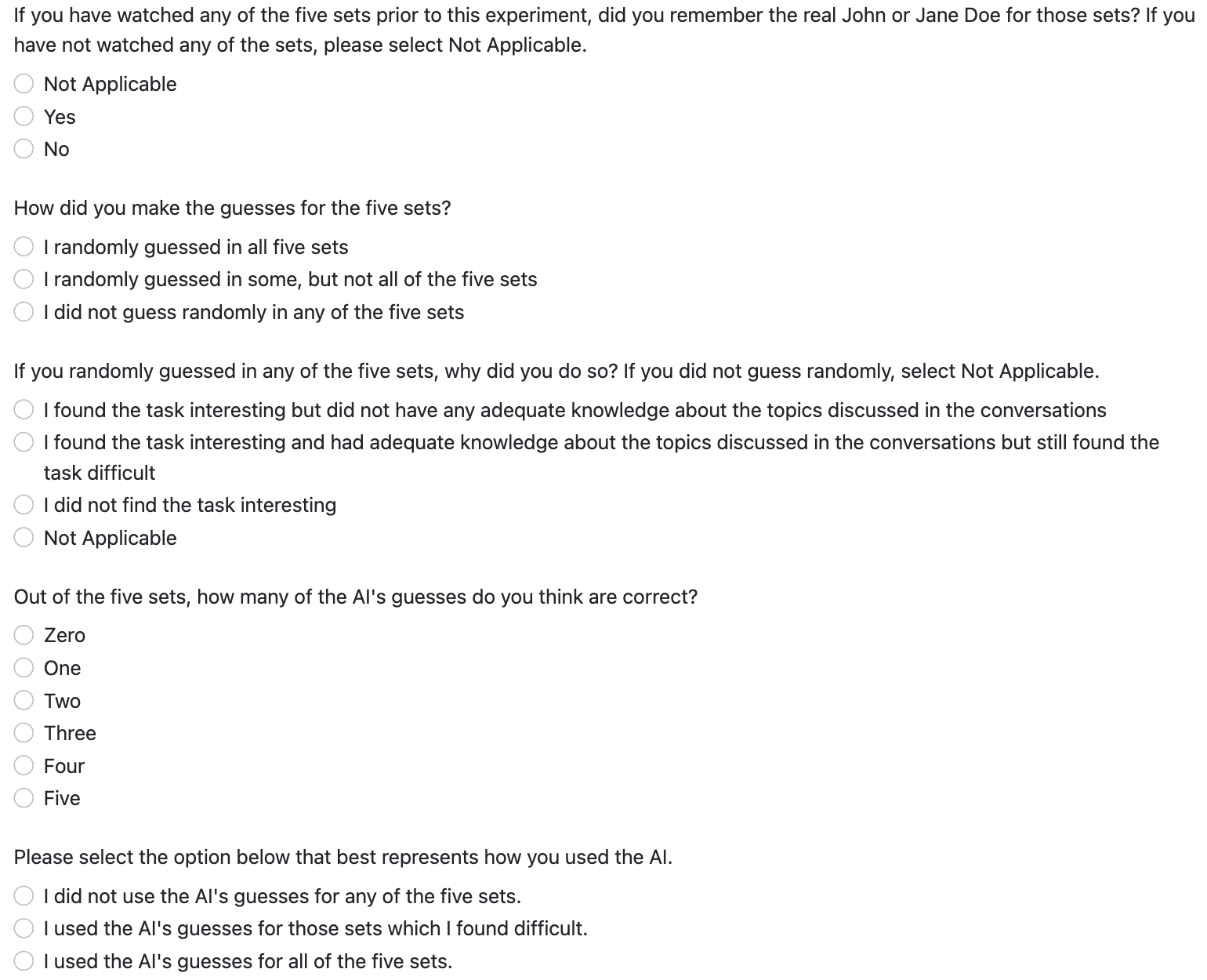}
    \label{fig:exitquestions2}
\end{figure}

(Continued on next page)
\newpage

\begin{figure}[h!]
    \centering
    \includegraphics[width=0.9\linewidth]{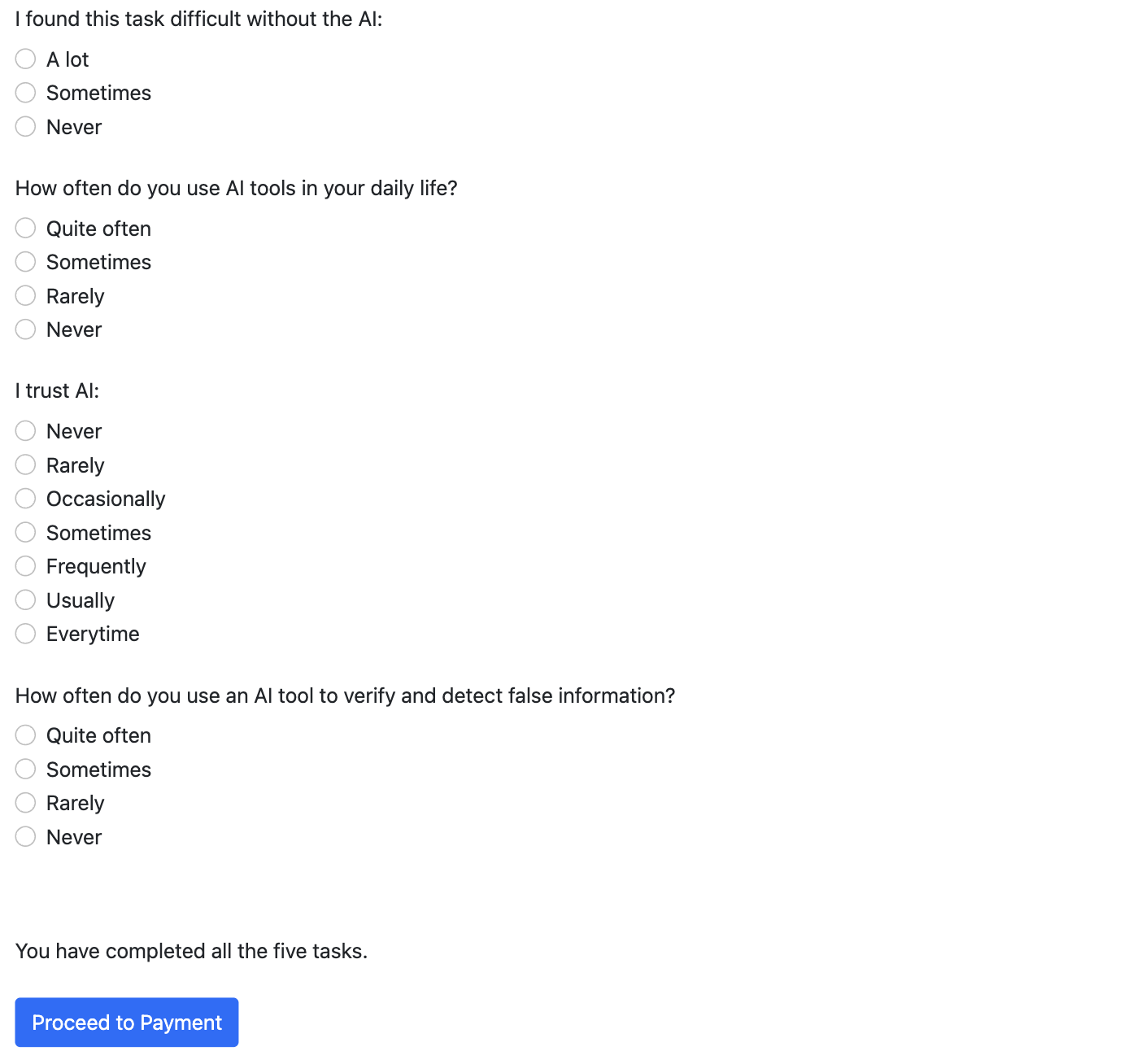}
    \label{fig:exitquestions3}
\end{figure}
\newpage

\subsection*{D.14. Payment}
\begin{figure}[h!]
    \centering
    \includegraphics[width=0.9\linewidth]{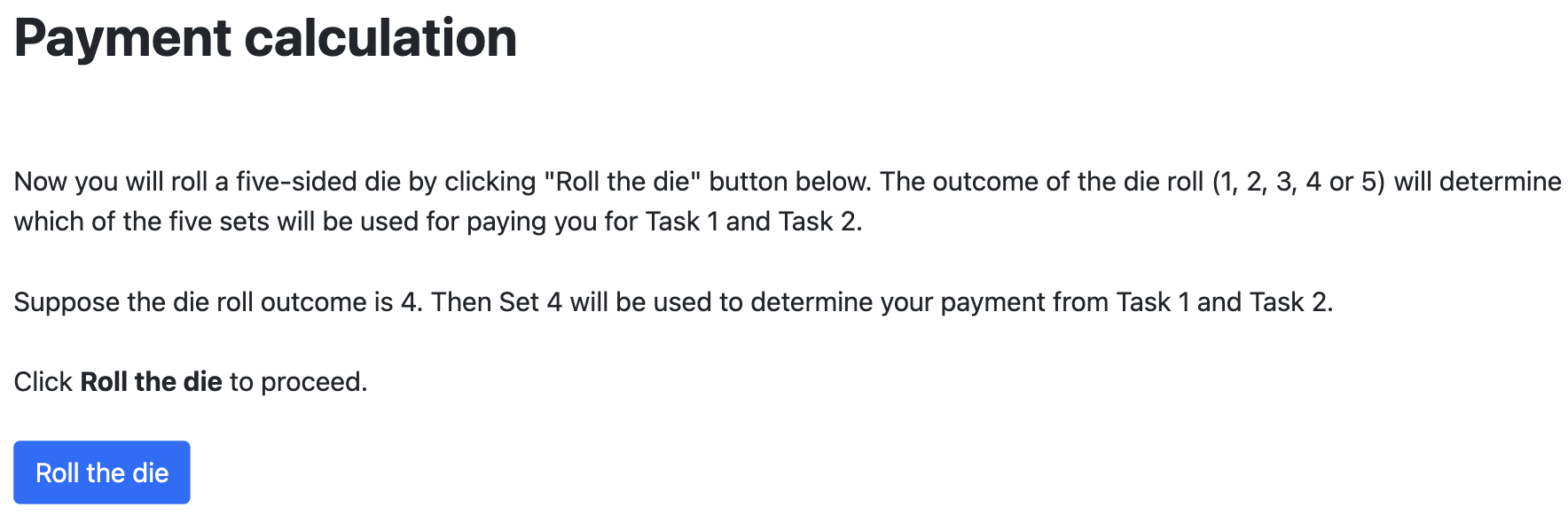}
    \label{fig:pay}
\end{figure}

\newpage
\setlength{\LTpre}{0pt}
\setlength{\LTpost}{0pt}
\begin{landscape}
\section* {Appendix E: Additional Tables}
\normalsize
\noindent
Table E1. Regressions for switch by demography
\begin{longtable}[c]{R{5cm} *{6}{R{2.5cm}}}
\hline
 & (1)         & (2)         & (3)    & (4) & (5) & (6)       \\
 &  Switch  & Switch & Switch & Switch & Switch in BB & Switch in BB\\ 
\hline
Confidence in own guess   & -0.0030*** &            & -0.0030*** &               & -0.0028***    &            \\
                  & (0.000)     &            & (0.000)     &               & (0.000)        &            \\
                  
Perceived difficulty    & 0.0497***   &            & 0.0508***   &               & 0.0514***      &            \\
                  & (0.000)     &            & (0.000)     &               & (0.010)        &            \\
                  
Relative confidence     &             & -0.0710*** &             & -0.0711***&  &  -0.0878***\\
                  &             & (0.000)    &             & (0.000)       &                & (0.000)    \\
                  
Perceived modal difficulty        &             & 0.0541***  &             & 0.0572***     &                & 0.0472**  \\
                  &             & (0.000)    &             & (0.000)       &                & (0.011)    \\
                  
Female  		& -0.0455** &  -0.0508*** & -0.0452**  &   -0.0504***   & -0.0349    &   -0.0376         \\
                  & (0.014)     &   (0.006)         &  (0.013)    &   (0.006)    & (0.172)        &  (0.134)          \\
                  
Age 35 and below  & 0.0020 & 0.0029   & 0.0046 &   0.0054            & 0.0017    &   -0.0118         \\
                  & (0.914)     &  (0.875)   & (0.801)     &   (0.766)            & (0.947)        &    (0.629)       \\
                  
At least  a college degree  & 0.0111 &  0.0129  & 0.0138 &  0.0150   & 0.0046    & 0.0077           \\
                  & (0.559)     &   (0.507)  & (0.462)     &   (0.432)   & (0.861)        &   (0.773)         \\

Expected AI accuracy &             &            &             &               & 0.0417***      & 0.0503***  \\
                  &             &            &             &               & (0.000)        & (0.000)    \\
                  
$LQ$               &             &            & -0.0215   & -0.0231       & -0.0146        & -0.0116    \\
                  &             &            & (0.484)     & (0.452)       & (0.630)        & (0.697)    \\
                  
$MQ$               &             &            & 0.0107     & -0.0091       & 0.0119         & -0.0077   \\
                  &             &            & (0.736)     & (0.775)       & (0.707)        & (0.800)    \\
                  
$WI$                &             &            & -0.0172     & -0.0262       &                &            \\
                  &             &            & (0.580)     & (0.403)       &                &            \\
                  
$LQ*WI$              &             &            & -0.0699     & -0.0679       &                &            \\
                  &             &            & (0.111)     & (0.123)       &                &            \\
                  
$MQ*WI$             &             &            & -0.0725     & -0.0469       &                &            \\
                  &             &            & (0.109)     & (0.297)       &                &            \\
                  
Constant          & 0.3026***    & 0.3171***   & 0.3374***    & 0.3531***      & 0.2010***       & 0.2679***   \\
                  & (0.000)     & (0.000)    & (0.000)     & (0.000)       & (0.009)        & (0.000)    \\
Observations      & 2725       & 2725      & 2725       & 2725         & 1370          & 1370  \\
\arrayrulecolor{lightgray}\hline
\multicolumn{6}{l}{Estimated difference in switch probability between AI advisors in BB} \\
$BB_{LQ} - BB_{MQ}$ & & &-0.0322 & -0.01400 &\\
 			  & & &(0.3249) & (0.6589) & \\
$BB_{LQ} - BB_{HQ}$ & & &-0.0215  & -0.02307 &\\
 			   & & & (0.4841) & (0.4524) &\\
$BB_{MQ} - BB_{HQ}$ & & & 0.0107& -0.0090 &\\
 			   & & & (0.7357) & (0.7749) &\\
\arrayrulecolor{lightgray}\hline
\multicolumn{6}{l}{Estimated difference in switch probability between AI advisors in WI} \\
$WI_{LQ} - WI_{MQ}$ & & &-0.0296 & -0.0349 &\\
 			    & & &(0.3290) & (0.2580) &\\
$WI_{LQ} - WI_{HQ}$ & & &-0.0914*** & -0.0909*** &\\
 			   & & & (0.0034) & (0.0038) &\\
$WI_{MQ} - WI_{HQ}$ & & &-0.0617 & -0.0559 &\\
 			    & & &(0.0550) & (0.0800) &\\
\arrayrulecolor{lightgray}\hline
\multicolumn{6}{l}{Estimated difference in switch probability between BB and WI} \\
$WI_{LQ} - BB_{LQ}$ & & &-0.0871*** & -0.0940*** &\\
 			   & & & (0.0046)& (0.0022) &\\
$WI_{MQ} - BB_{MQ}$ & & &-0.0897*** & -0.0731** &\\
 			   & & & (0.0061) & (0.0236) &\\
$WI_{HQ} - BB_{HQ}$ & & &-0.0172 & -0.0262 &\\
 			   & & & (0.5804) & (0.4026) &\\
\arrayrulecolor{black}\hline
\end{longtable}
\noindent
\footnotesize{Notes: Linear Probability Model (LPM) estimates reported with \textit{p-value} in parentheses that are based on robust standard errors clustered by participant-id. $^{***}$ and $^{**}$ denote 1\% and 5\% level of significance, respectively.}
\end{landscape}

\newpage
\setlength{\LTpre}{0pt}
\setlength{\LTpost}{0pt}
\normalsize
\noindent
Table E2. Comparison of outcomes between the main and modified treatments 
\begin{longtable}{ p{3cm} *{2}{p{1.4cm}} p{2.25cm} *{2}{p{1.4cm}} p{2cm}}
\hline
 & $BB$ & $MBB$ & Difference & $WI$ & $MWI$ & Difference \\ 
\hline
\multicolumn{4}{l}{Switch rate} \\
\arrayrulecolor{lightgray}\hline
$LQ$ &0.2154 & 0.3067 & -0.0913** & 0.1363 & 0.2333 & -0.0971***\\
        & & & (0.0228) & & & (0.0049)\\
$MQ$ &0.2427 & .2733 & -.0306 & 0.1711 & 0.1867 & -0.0155\\
        & & & (0.4547) & & & (0.6648)\\
$HQ$ &0.2489 & 0.2452 & 0.0038 & 0.2289 & 0.2333 & -0.0044\\
        & & & (0.9250) & & & (0.9110)\\
\arrayrulecolor{black}\hline
\multicolumn{4}{l}{Final accuracy rate - Initial accuracy rate} \\
\arrayrulecolor{lightgray}\hline
$LQ$ &-0.0462 & -0.0467 & 0.0005& -0.0044 & -0.0667 & 0.0623**\\
        & & & (0.9871) & & & (0.0211)\\
$MQ$ & -0.0225 & -0.0133 & -0.0091 & 0.0222 & -0.0267 & 0.0489\\
        & & & (0.8281) & & & (0.1401)\\
$HQ$ &0.1043 & 0.0968 & 0.0075 & 0.1 & 0.1133 & -0.0133\\
        & & & (0.8473) & & & (0.7149)\\
\arrayrulecolor{black}\hline
\end{longtable}
\noindent
\footnotesize {Notes: (i). Difference = Main treatment - Modified treatment. (ii). $p$-values in parentheses correspond to t-test for $H_0$: Difference $=0$ and $H_1$: Difference $\neq 0$. (iii). $^{***}$ and $^{**}$ denote statistical significance at 1\% and 5\% level, respectively.}

\end{document}